\documentclass[10pt,twocolumn,letterpaper]{article}

\usepackage{cvpr}              %

\usepackage{booktabs}   %
\usepackage[table,dvipsnames]{xcolor}
\definecolor{oursrow}{RGB}{228,245,249}
\newcommand{\makecell}[1]{\begin{tabular}{@{}c@{}}#1\end{tabular}}
\usepackage{multirow}
\newcommand{\hdashline}{\midrule}
\usepackage{pifont}     %
\newcommand{\cmark}{\textcolor{ForestGreen}{\ding{51}}}  %
\newcommand{\xmark}{\textcolor{BrickRed}{\ding{55}}}     %

\usepackage{etoc}     %
\usepackage{xspace}
\newcommand{\benchmarkname}{TAPVid-MV\xspace}  %

\newcommand{\dmetric}[1]{\ensuremath{\delta_{\mathrm{#1}}^{3D}}\xspace}

\newcommand{\dthresh}[2]{\ensuremath{\delta_{<#1,\mathrm{#2}}^{3D}}\xspace}

\makeatletter
\renewcommand\paragraph{\@startsection{paragraph}{4}{\z@}{0.7ex plus 0.2ex minus 0.1ex}{-1em}{\normalfont\normalsize\bfseries}}
\makeatother

\definecolor{cvprblue}{rgb}{0.21,0.49,0.74}
\usepackage[pagebackref,breaklinks,colorlinks,citecolor=cvprblue]{hyperref}

\def\paperID{178} %
\def\confName{3DV\xspace}
\def\confYear{2027\xspace}

\title{\benchmarkname: A Benchmark for Tracking Any Point in 3D Across Multiple Views}

\author{%
Skanda Koppula\textsuperscript{1,2*} \quad
Frano Raji\v{c}\textsuperscript{3*} \quad
Abdullah Faiz Ur Rahman\textsuperscript{2} \quad
Yi Yang\textsuperscript{1}\\
Ignacio Rocco\textsuperscript{1} \quad
Jeet Thakwani\textsuperscript{2} \quad
Rishabh Kabra\textsuperscript{1,2} \quad
Andrew Zisserman\textsuperscript{1,4}\\
Joao Carreira\textsuperscript{1} \quad
Siyu Tang\textsuperscript{3} \quad
Carl Doersch\textsuperscript{1} \quad
Gabriel Brostow\textsuperscript{2}\\[3pt]
\small
\textsuperscript{1}Google DeepMind \quad
\textsuperscript{2}University College London \quad
\textsuperscript{3}ETH Z\"urich \quad
\textsuperscript{4}University of Oxford\\
{\small\textsuperscript{*}Equal contribution}
}

\begin{document}
\twocolumn[{
\renewcommand\twocolumn[1][]{#1}
\maketitle
\begin{center}
    \vspace{-0.85cm}
    \captionsetup{type=figure}
    \includegraphics[width=\linewidth]{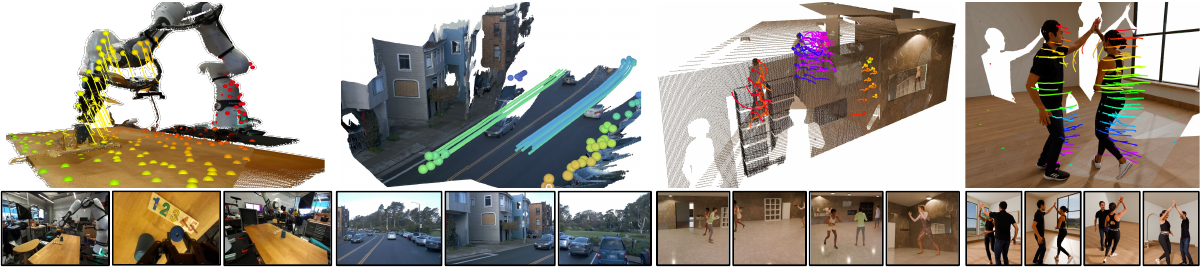}
    \vspace{-0.75cm}
    \captionof{figure}{%
      \textbf{\benchmarkname} is the first benchmark for long-term 3D point tracking across multiple synchronized views under camera motion and changing camera layouts. From left to right, the examples span robotics, driving, indoor activities, and human interaction.
    }
    \label{fig:teaser}
\end{center}
}]
\maketitle

\etocsettocdepth.toc{none}

\begin{abstract}
\vspace{-0.5cm}
Multi-camera systems are increasingly practical for robotics, AR/VR, and autonomous driving because complementary views reduce depth ambiguity and preserve visibility under occlusion. Existing point-tracking benchmarks, however, focus on a single video or static multi-camera rigs. None test long-term 3D point tracking across several synchronized views under camera motion. We introduce \textbf{\benchmarkname} (Tracking Any Point in Video across Multiple Views), the first benchmark for this setting. It contains a curated set of 238 sequences, 1,004 calibrated camera streams, and 74,147 point tracks across seven subsets spanning indoor and outdoor domains, from robotics and human activity to driving and synthetic procedural scenes. We obtain these trajectories using dataset-specific auxiliary modalities: sensor depth, LiDAR, SLAM and SfM points, human meshes, posed object meshes, and simulation. Every sequence and trajectory is visually verified by human annotators. Across more than 30 baselines, no method comes close to solving the task. Surprisingly, existing multi-view point trackers do not consistently outperform monocular point trackers. By evaluating reconstruction and point tracking on the same datasets, \benchmarkname helps distinguish errors in recovered geometry from errors in point correspondence. Through this joint analysis, we identify geometry recovery as a major bottleneck for accurate 3D point tracking. Our released annotations also support monocular 2D and 3D point tracking, future-trajectory prediction, and 4D reconstruction. Project page: \href{https://tapvidmv.github.io/}{\color{cvprblue}tapvidmv.github.io}.
\end{abstract}
\vspace{-12pt}
\section{Introduction}
\label{sec:intro}

What can many views reveal that one cannot? Multi-camera systems are becoming increasingly practical for robotics, autonomous driving, AR/VR, and human capture. Even two synchronized views reduce monocular depth ambiguity, while additional views can keep a tracked point observable through occlusion and extend coverage beyond any single camera. These signals should make long-term 3D point tracking substantially more reliable across diverse scenes and camera configurations. Yet existing point-tracking benchmarks~\cite{doersch2022tapvid,vecerik2024robotap,balasingam2024drivetrack,koppula2024tapvid3d,rajic2025mvtracker,koo2025mvtap,galoaa2025lookaround} evaluate a single video or static multi-camera rigs. No benchmark covers long-term 3D point tracking across several synchronized views under camera motion and changing camera layouts.

Evaluating this capability requires separating point correspondence from geometry. Modern systems obtain 3D trajectories by lifting 2D tracks using estimated depth and cameras~\cite{koppula2024tapvid3d}, tracking directly in a supplied reconstruction~\cite{zhang2025tapip3d,rajic2025mvtracker}, or jointly estimating geometry and trajectories~\cite{xiao2025spatialtrackerv2,karhade2025any4d,zhang2026efficiently,jiang2026omnix}. A final 3D tracking error may therefore originate in correspondence, depth, camera calibration, or cross-view registration. Existing point-tracking benchmarks report trajectory accuracy without measuring reconstruction quality on the same data. They therefore cannot reveal whether progress comes from the point tracker or the geometry. Comparing systems without controlling or measuring their reconstruction conflates the two.

We introduce \textbf{\benchmarkname} (\emph{Tracking Any Point in Video across Multiple Views}), the first benchmark for 3D tracking in this setting. It contains a curated set of 238 sequences, 1,004 calibrated camera streams, and 74,147 ground-truth 3D trajectories across seven subsets. Those span indoor and outdoor domains, from robotics and human activity to driving and synthetic procedural scenes. Every subset contains camera motion, including egocentric views and independently moving cameras. Constructing such labels is difficult because humans cannot annotate metric 3D point trajectories from 2D video alone. Recovering them requires consistent associations across views and time, together with accurate depth or precisely calibrated and synchronized cameras. We obtain the trajectories using dataset-specific auxiliary modalities (sensor depth, LiDAR, SLAM and SfM points, human meshes, and posed object meshes) or simulation, and we manually validate all sequences. Given the synchronized RGB streams and a query that identifies one physical point by its pixel location in one camera at one frame (but no ground-truth depth maps or camera poses), a model must recover a single 3D trajectory for that point through time, shared across all views. We score this trajectory in the query camera's coordinate frame, in each remaining camera's frame, and in a common world frame, while measuring reconstruction quality on the same data.

Our evaluation of more than 30 baselines reveals three findings. First, no method comes close to solving the task. Second, existing multi-view point trackers do not consistently outperform monocular trackers. Given the same reconstruction, the monocular TAPIP3D~\cite{zhang2025tapip3d} leads aggregate query-view and world-space accuracy and performs comparably to MVTracker~\cite{rajic2025mvtracker} in non-query views (21.5 versus 21.7). Third, current joint models remain constrained by their own reconstructions. The geometry they recover is itself a major source of tracking error. For SpatialTrackerV2~\cite{xiao2025spatialtrackerv2}, keeping its weights fixed while replacing its native reconstruction with the shared VGGT-$\Omega$~\cite{wang2026vggtomega} reconstruction more than doubles aggregate query-view accuracy from 10.8 to 23.9. Measuring reconstruction and tracking on the same data allows \benchmarkname to identify this geometry bottleneck separately from errors in correspondence.

We identify four priorities for future work. (1) Reconstruction models must reliably recover dynamic scenes from a small number of moving cameras, which none currently achieves. (2) Future evaluations should report 3D point tracking metrics alongside reconstruction metrics on the same sequences, while architecture comparisons should control for learned geometry priors and training data so gains from either are not mistaken for improvements in point tracking. (3) New point tracking architectures should exploit additional views and cross-view information more effectively. (4) Accurate geometry does not remove the familiar challenges of point tracking, since trackers running on the most accurate reconstruction still drift along surfaces and swap identities when objects touch, making these failures key targets for model development. Alongside the benchmark data, baseline implementations, and evaluation code, we release our procedural Perpetua data generator and metric 3D trajectories for 5,371 robotic episodes, expanding the multi-view training data available for reconstruction and tracking. We invite the community to use these resources to drive progress in multi-view 3D point tracking and related tasks, including monocular 2D and 3D point tracking, future-trajectory prediction, and 4D reconstruction.

\section{Related Work}
\label{sec:related}

\begin{table*}[!t]
\centering
\setlength{\tabcolsep}{1.8pt}
\renewcommand{\arraystretch}{1.0}
\caption{Positioning of our multi-view TAP-3D task relative to related 3D tasks, highlighting which properties each supports. We design for long-term, pixel-level 3D tracking of any rigid or deformable object, given multiple RGB views of the scene.}
\label{tab:related_task_comparison}
\vspace{-0.27cm}
\resizebox{\textwidth}{!}{%
\begin{tabular}{llcccccc}
\toprule
\textbf{Task}
&
\textbf{Example benchmark}
&
\makecell{\textbf{Long-term}\\\textbf{tracking}}
&
\makecell{\textbf{3D}\\\textbf{output}}
&
\makecell{\textbf{Pixel-level}\\\textbf{motion}}
&
\makecell{\textbf{Non-rigid}\\\textbf{surfaces}}
&
\makecell{\textbf{No prior}\\\textbf{3D model}}
&
\makecell{\textbf{Multi-}\\\textbf{view}}
\\
\midrule

Monocular depth
& NYU Depth v2~\cite{silberman2012nyudepth}
& \xmark & \cmark & \xmark & \cmark & \cmark & \xmark \\

Scene flow
& FlyingThings3D~\cite{mayer2016flyingthings}
& \xmark & \cmark & \cmark & \cmark & \cmark & \cmark \\

Scene reconstruction
& ScanNet~\cite{dai2017scannet}, Sintel~\cite{butler2012sintel}
& \xmark & \cmark & \xmark & \cmark & \cmark & \cmark \\

3D object box tracking
& KITTI Tracking~\cite{geiger2012kitti}, Waymo Open~\cite{sun2020waymo}
& \cmark & \cmark & \xmark & \xmark & \cmark & \cmark \\

3D keypoint tracking
& Human3.6M~\cite{ionescu2014human36m}
& \cmark & \cmark & \xmark & \xmark & \xmark & \cmark \\

Human motion reconstruction
& EgoBody~\cite{zhang2022egobody}, RICH~\cite{huang2022rich}, EMDB~\cite{kaufmann2023emdb}
& \cmark & \cmark & \xmark & \cmark & \xmark & \xmark \\

2D point tracking (TAP)
& TAPVid-DAVIS~\cite{doersch2022tapvid}
& \cmark & \xmark & \cmark & \cmark & \cmark & \xmark \\

3D point tracking (TAP-3D)
& TAPVid-3D~\cite{koppula2024tapvid3d}, WorldTrack~\cite{feng2025st4rtrack}
& \cmark & \cmark & \cmark & \cmark & \cmark & \xmark \\

\rowcolor{oursrow}
\textbf{Multi-view TAP-3D}
& \textbf{\benchmarkname} (ours)
& \cmark & \cmark & \cmark & \cmark & \cmark & \cmark \\

\bottomrule
\end{tabular}%
}
\vspace{-0.35cm}
\end{table*}

\begin{table}[!t]
\centering
\setlength{\tabcolsep}{1.8pt}
\renewcommand{\arraystretch}{1.0}
\caption{Comparison of the data and camera settings covered by TAP benchmarks. \benchmarkname is the only benchmark with real videos, more than two views, moving cameras, and cameras that move independently rather than as a fixed rig.}
\label{tab:related_work_comparison}
\vspace{-0.27cm}
\resizebox{\columnwidth}{!}{%
\begin{tabular}{lccccc}
\toprule
\textbf{Benchmark}
&
\makecell{\textbf{Real}\\\textbf{videos}}
&
\makecell{\textbf{3D}\\\textbf{labels}}
&
\makecell{\textbf{$>2$}\\\textbf{views}}
&
\makecell{\textbf{Moving}\\\textbf{Objects/}\\\textbf{Cameras}}
&
\makecell{\textbf{Independently}\\\textbf{Moving Cameras}}
\\
\midrule

\addlinespace[0.15em]
\multicolumn{6}{l}{\textbf{Synthetic benchmarks}}\\
\midrule
Kubric~\cite{greff2022kubric}
& \xmark & \cmark & \xmark & \cmark / \cmark & \xmark \\

PointOdyssey~\cite{zheng2023pointodyssey}
& \xmark & \cmark & \xmark & \cmark / \cmark & \xmark \\

SynthVerse~\cite{zhao2026synthverse}
& \xmark & \cmark & \xmark & \cmark / \cmark & \xmark \\

Syn4D~\cite{jiang2026syn4d}
& \xmark & \cmark & \cmark & \cmark / \cmark & \cmark \\

\midrule
\addlinespace[0.15em]
\multicolumn{6}{l}{\textbf{2D-only benchmarks}}\\
\midrule
TAPVid-DAVIS~\cite{doersch2022tapvid}
& \cmark & \xmark & \xmark & \cmark / \cmark & \xmark \\

EgoPoints~\cite{egopoints}
& \cmark & \xmark & \xmark & \cmark / \cmark & \xmark \\

RoboTAP~\cite{vecerik2024robotap}
& \cmark & \xmark & \xmark & \cmark / \cmark & \xmark \\

\midrule
\addlinespace[0.15em]
\multicolumn{6}{l}{\textbf{Single-view 3D benchmarks}}\\
\midrule
TAPVid-3D~\cite{koppula2024tapvid3d}
& \cmark & \cmark & \xmark & \cmark / \cmark & \xmark \\

DriveTrack~\cite{balasingam2024drivetrack}
& \cmark & \cmark & \xmark & \cmark / \cmark & \xmark \\

\midrule
\addlinespace[0.15em]
\multicolumn{6}{l}{\textbf{Multi-view benchmarks}}\\
\midrule
DexYCB~\cite{chao2021dexycb}
& \cmark & \cmark & \cmark & \cmark / \xmark & \xmark \\

Panoptic Studio~\cite{joo2015panoptic}
& \cmark & \cmark & \cmark & \cmark / \xmark & \xmark \\

Stereo4D~\cite{jin2025stereo4d}
& \cmark & \cmark & \xmark & \cmark / \cmark & \xmark \\

PACE~\cite{you2024pace}
& \cmark & \cmark & \cmark & \xmark / \cmark & \xmark \\

\rowcolor{oursrow}
\textbf{\benchmarkname} (ours)
& \cmark & \cmark & \cmark & \cmark / \cmark & \cmark \\

\bottomrule
\end{tabular}%
}
\vspace{-0.35cm}
\end{table}

Tracking Any Point (TAP) is the task of predicting the pixel trajectory of a queried surface point through a monocular video~\cite{doersch2022tapvid}. A broad family of 2D point trackers addresses this task~\cite{harley2022particle,zheng2023pointodyssey,li2024taptrv3,doersch2023tapir,karaev2024cotracker,doersch2024bootstap,lemoing2024dense,cho2024locotrack,karaev2024cotracker3,zhang2026megaflow,jung2026tapnextpp,aydemir2025trackon,harley2025alltracker,lai2026cowtracker}. TAP-3D~\cite{koppula2024tapvid3d} extends this formulation to metric 3D trajectories in monocular video. Our \emph{multi-view TAP-3D} task requires long-term, pixel-level 3D trajectories across multiple views in a common world frame. \Cref{tab:related_task_comparison} contrasts it with TAP, TAP-3D, and other 3D tasks.

Recent 3D trackers combine point tracking with estimated depth and cameras~\cite{xiao2024spatialtracker,wang2024scenetracker,ngo2025delta,xiao2025spatialtrackerv2,zhang2025tapip3d}, while joint 4D models estimate geometry and trajectories together~\cite{badki2026l4p,feng2025st4rtrack,karhade2025any4d,xiao2025spatialtrackerv2,sucar2026vdpm,luo2026fourrc,zhang2026efficiently,miao2026trajvg}. Because metric 3D tracking depends on both correspondence and recovered geometry, a trajectory score alone cannot reveal whether an error arises from incorrect correspondence, inaccurate geometry, or both. Moreover, monocular evaluations cannot test whether a query in one camera can be tracked across synchronized views. Multi-view methods broaden point tracking beyond monocular video through several formulations. MV-TAP~\cite{koo2025mvtap} tracks across views in 2D, but requires a query in every view. MVTracker~\cite{rajic2025mvtracker} and LAPA~\cite{galoaa2025lookaround} recover 3D trajectories from multiple cameras, but their evaluations assume known calibration and static cameras. OmniX~\cite{jiang2026omnix} jointly estimates multi-view geometry and 3D trajectories, but reports quantitative multi-view trajectory results only on its synthetic benchmark.

\Cref{tab:related_work_comparison} summarizes the remaining gap across existing point tracking benchmarks. Real-world multi-view benchmarks use either static cameras or a moving rigid camera rig. None combines more than two views with independently moving cameras. \benchmarkname closes this gap and evaluates reconstruction and tracking on the same data.

\section{Multi-View TAP-3D}
\label{sec:task}

We provide a short description of our multi-view TAP-3D task. A model is given time-synchronized RGB streams of a dynamic scene, captured by multiple cameras that may move throughout the sequence, and a set of query points. Each query identifies a physical surface point through one visible pixel in \emph{exactly one} view at one frame, such as a point selected on a tablecloth during robotic manipulation. Unlike MV-TAP~\cite{koo2025mvtap}, no corresponding query is provided in other views, because assuming that the same physical point has already been identified in every camera is unrealistic in applications.

For each query point, the model must predict its 3D trajectory across all frames in a single world coordinate frame shared across all views, together with its visibility in every view and frame. Since ground-truth depth and camera poses are not provided as input, the model defines this world frame. We use the model's estimate of the first camera at the first frame as the reference coordinate system and evaluate its predictions up to a single rescaling factor.

\begin{figure*}[t]
  \centering
  \includegraphics[width=\textwidth]{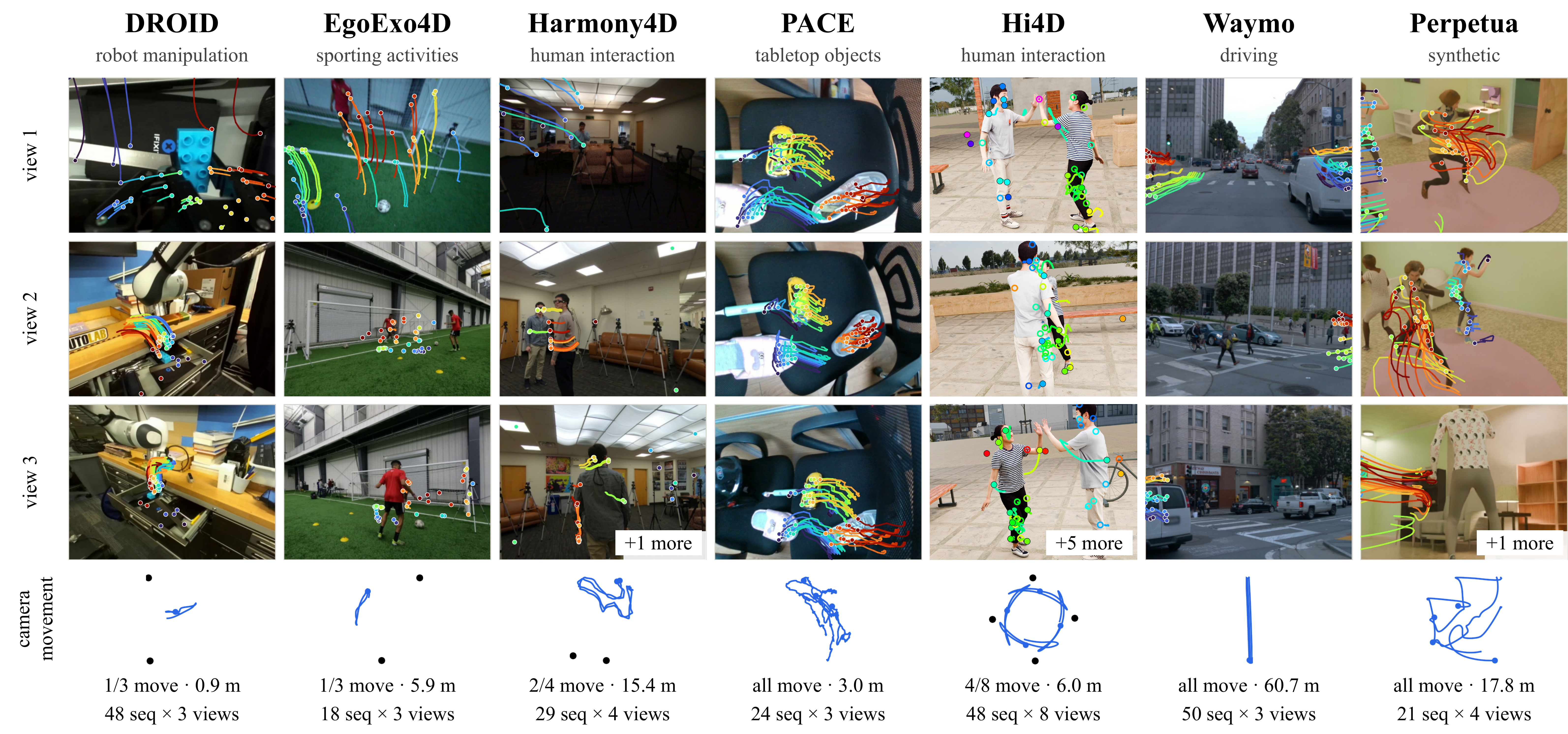}
  \vspace{-0.7cm}
  \caption{%
    \textbf{The \benchmarkname subsets: 238 sequences and 1,004 calibrated camera views across seven domains.} Each row shows a different camera observing the \emph{same} instant, with the same 3D tracks projected into every row. Glyphs beneath each subset show the cameras' locations from above, over the depicted sequence.
  }
  \vspace{-0.2cm}
  \label{fig:mvtap_splits}
\end{figure*}

\paragraph{Metrics.}
We adapt the TAP-3D metrics~\cite{koppula2024tapvid3d} to the multi-view setting: average Jaccard $\mathrm{AJ}^{3D}$, average 3D location accuracy $\delta^{3D}$, and occlusion accuracy $\mathrm{OA}$. Because GT geometry is not provided as input, predictions are rescaled before scoring. We organize the evaluation into three regimes:
\begin{enumerate}
    \item \textbf{Query View (Q):} We evaluate predicted trajectories in the camera frame of the view where each query was provided. At $V=1$, this reduces to standard TAP-3D, isolating single-view tracking capabilities and making it directly comparable to the monocular literature.
    \item \textbf{Non-Query Views (N):} We score the predictions in the camera frames of the \emph{other} views, probing recovery beyond the query view. Because visibility differs across cameras, the Q and N location metrics generally average over different sets of visible point-time observations.
    \item \textbf{World Space (W):} While Q and N evaluate each camera in its own frame, we additionally score all queries in a single, globally consistent world frame. This tests whether a model's multi-view predictions form globally consistent 3D trajectories, directly penalizing cross-view misalignment and pose drift.
\end{enumerate}
For average 3D location accuracy, we denote the regimes by \dmetric{Q}, \dmetric{N}, and \dmetric{W}. Full definitions of query sampling, rescaling, and adaptive thresholds are given in \cref{sec:supp_metrics}.

\section{The \benchmarkname Benchmark}
\label{sec:tapvidmv}

We build \benchmarkname{} using datasets from domains in which multi-camera capture is common: DROID, Ego-Exo4D, Harmony4D, PACE, Hi4D, Waymo, and Perpetua. The source datasets provide different forms of supervision, so we construct metric 3D tracks for each subset from its most reliable available signal: robot kinematics and stereo, human meshes, posed object meshes, LiDAR, or exact synthetic geometry. We apply automatic filtering and manually inspect the tracks projected into every view. The benchmark comprises 238 sequences and 1,004 views: five subsets contain real-world recordings, Hi4D is rendered from real capture assets, and Perpetua is fully synthetic. Each scene has three to eight cameras. Every subset contains camera motion, and three include egocentric views. \Cref{fig:mvtap_splits,tab:subset_preview} summarize the domains, cameras, depth sources, and sizes. Visualizations for every single sequence for external verification are released in \cref{sec:per_sequence_visuals}.

\vspace{-0.1cm}
\subsection{DROID}
\label{sec:tapvidmv:droid}
\vspace{-0.15cm}
DROID~\cite{khazatsky2024droid} provides synchronized stereo video, a moving wrist camera, and 7-DOF robot kinematics. We refine S2M2 stereo depth~\cite{s2m2} on the specular gripper, optimize camera extrinsics against the robot model and cross-view scene consistency, then track static scene points under a static-scene prior and robot-surface points by forward kinematics. We process 5,371 episodes for training and pre-training and curate 48 for evaluation. Each candidate track is reviewed in 3D and 72.6\% are kept, leaving about 39 static and 48 robot tracks per sequence (\cref{sec:supp_droid}).

\begin{table*}[tb]
\centering
\setlength{\tabcolsep}{2.7pt}
\renewcommand{\arraystretch}{0.98}
\caption{\textbf{\benchmarkname seven subsets.} Tracks/scene is the mean number of tracks per scene. Dyn. is the fraction of tracks whose world-space position moves by more than 5\,cm within the sequence. $^*$Hi4D renders real-world captured scans inside synthetic environments.}
\label{tab:subset_preview}
\vspace{-0.27cm}
\footnotesize
\begin{tabular}{@{}llcc@{\hspace{0.45em}}cccc@{\hspace{0.45em}}ccc@{\hspace{0.45em}}ccc@{}}
\toprule
\textbf{Subset} & \textbf{Domain}
& \multicolumn{2}{c}{\textbf{Source}}
& \multicolumn{4}{c}{\textbf{Cameras}}
& \multicolumn{3}{c}{\textbf{Depth}}
& \multicolumn{3}{c}{\textbf{Size}}
\\
\cmidrule(lr){3-4}\cmidrule(lr){5-8}\cmidrule(lr){9-11}\cmidrule(lr){12-14}
& & Real & Synth.
& Total & Static & Dyn. & Ego
& GT & LiDAR & Pseudo
& Scenes & Tracks/scene & Dyn.
\\
\midrule

DROID~\cite{khazatsky2024droid}
& Robotics
& \cmark & -
& 3 & 2 & 1 & 1
& \xmark & \xmark & \cmark
& 48 & 87 & 45.5\% \\

EgoExo4D~\cite{grauman2024egoexo4d}
& Soccer
& \cmark & -
& 3 & 2 & 1 & 1
& \xmark & \xmark & \xmark
& 18 & 497 & 39.5\% \\

Harmony4D~\cite{khirodkar2024harmony4d}
& Sports, dance
& \cmark & -
& 4 & 2 & 2 & 2
& \xmark & \xmark & \xmark
& 29 & 107 & 41.4\% \\

PACE~\cite{you2024pace}
& Tabletop objects
& \cmark & -
& 3 & 0 & 3 & 0
& \cmark & \xmark & \xmark
& 24 & 76 & 31.0\% \\

Hi4D~\cite{yin2023hi4d}
& Humans
& - & ~~\cmark$^*$
& 8 & 4 & 4 & 0
& \cmark & - & -
& 48 & 218 & 49.6\% \\

Waymo~\cite{sun2020waymo}
& Driving
& \cmark & -
& 3 & 0 & 3 & 0
& \xmark & \cmark & \cmark
& 50 & 483 & 46.1\% \\

Perpetua
& Synthetic
& - & \cmark
& 4 & 0 & 4 & 0
& \cmark & - & -
& 21 & 1,024 & 50.0\% \\

\bottomrule
\end{tabular}
\vspace{-0.1cm}
\end{table*}

\vspace{-0.1cm}
\subsection{Ego-Exo4D}
\label{sec:tapvidmv:egoexo4d}
\vspace{-0.15cm}
From Ego-Exo4D~\cite{grauman2024ego}, we select soccer sequences in which an interacting participant is observed by the head-mounted camera and at least two static cameras. Dynamic tracks come from THFM~\cite{wang2026thfm} human fits fused across the static views; static tracks come from the provided SLAM point cloud, filtered by VGGT-$\Omega$~\cite{wang2026vggtomega} depth. Occlusion by other players and the ball is resolved with SAM~3~\cite{sam3} masks, and we retain 500 tracks jointly visible from the three views for at least 50 frames (\cref{sec:supp_datasets_egoexo4d}).

\vspace{-0.1cm}
\subsection{Harmony4D}
\label{sec:tapvidmv:harmony4d}
\vspace{-0.15cm}
We extend Harmony4D~\cite{khirodkar2024harmony4d}, as used by MV-TAP~\cite{koo2025mvtap}, with both egocentric cameras, static scene tracks, and verification of the human tracks: each of the 29 sequences keeps two egocentric and two static exocentric views. Candidate dynamic tracks are vertices of the provided SMPL fits, with visibility from mesh ray casting; we reject tracks that persistently disagree with a cycle-consistent TAPNext++~\cite{jung2026tapnextpp} image-tracking consensus. Static tracks come from COLMAP points filtered by VGGT-$\Omega$ depth consistency, followed by manual inspection (\cref{sec:supp_datasets_harmony4d}).

\vspace{-0.1cm}
\subsection{PACE}
\label{sec:tapvidmv:pace}
\vspace{-0.15cm}
PACE~\cite{you2024pace} provides scanned objects, per-frame 6-DoF poses, and synchronized RGB-D from a hand-carried three-camera rig in cluttered rearrangement scenes: 238 household objects from 43 categories that touch, overlap, and are moved during the take. We initialize tracks from sensor depth, propagate them with the annotated object poses, and recover camera motion using a reference object assumed to remain stationary. We estimate visibility from depth, apply depth and segmentation checks to tracks, and visually review sequences to exclude poor-quality ones. After filtering, visual review, and balancing static and dynamic tracks, we retain 24 of the 69 candidate sequences, with an average of 76 tracks per sequence (\cref{sec:supp_datasets_pace}).

\vspace{-0.1cm}
\subsection{Hi4D}
\label{sec:tapvidmv:fhi4d}
\vspace{-0.15cm}
Hi4D~\cite{yin2023hi4d} provides real two-person captures with per-frame SMPL fits and textured scans, and calibrated cameras. We propagate points through the fitted bodies, transfer them to clothing by intersecting the scan mesh, and review the candidates in 3D. This yields 5,228 verified human tracks across 48 sequences. We place the textured scans in sequence-specific synthetic environments in Blender and render using Cycles from four fixed viewpoints derived from the released calibration and four simulated moving cameras. \Cref{sec:supp_datasets_hi4d} gives the full construction and labeling details, and \cref{sec:hi4d_static_only} evaluates the original static captures.

\vspace{-0.1cm}
\subsection{Waymo}
\label{sec:tapvidmv:waymo}
\vspace{-0.15cm}
For Waymo~\cite{sun2020waymo}, we extend the DriveTrack-style pipeline~\cite{balasingam2024drivetrack,koppula2024tapvid3d} from one camera to three synchronized adjacent front cameras. LiDAR points inside annotated vehicle boxes are propagated using vehicle rigidity and box poses, then composed with the ego-vehicle pose into a shared world frame. We track all annotated vehicles jointly, and both vehicle and background tracks inherit metric LiDAR range; all three cameras ride on the moving ego-vehicle. Visibility is checked against the closest LiDAR return, and we sample 500 tracks per sequence that are visible in at least two views (\cref{sec:supp_datasets_waymo}).

\vspace{-0.1cm}
\subsection{Perpetua}
\label{sec:tapvidmv:perpetua}
\vspace{-0.15cm}
We develop Perpetua, a procedural generator for dynamic indoor multi-view video. Infinigen~\cite{infinigen2024indoors} creates furnished rooms. In each room, four actor routes are jointly optimized for clearance, actor separation, and smooth motion. Kimodo~\cite{Kimodo2026} generates route-conditioned full-body motion from natural-language action descriptions for four dressed SMPL-X actors~\cite{SMPL-X:2019,tesch2025bedlam2}. Four moving cameras are optimized for collision avoidance, actor framing, and viewpoint diversity. Since every scene component is known, Perpetua provides exact 3D tracks and per-view visibility (\cref{sec:supp_datasets_perpetua}).

\begin{table*}[!t]
\centering
\setlength{\tabcolsep}{3.5pt}
\caption{\textbf{Select TAPVid-MV baseline results.} World-space 3D tracking accuracy \dmetric{W} per subset for the strongest method of some families of \cref{sec:models}, with Overall \dmetric{W}, \dmetric{N}, and \dmetric{Q} averaged over the seven subsets. All point trackers receive the same shared VGGT-$\Omega$~\cite{wang2026vggtomega} reconstruction; the monocular joint methods predict no shared world frame and appear only in \cref{tab:main_results}. Gray cells mark a method trained on that subset's source data, and the gray reference row uses ground-truth depth where available. Best and second-best \dmetric{W} are bold and underlined; full results and reconstruction quality are in \cref{tab:main_results,tab:reconstruction_quality}.}
\label{tab:headline_results}
\vspace{-0.27cm}
\footnotesize
\resizebox{\linewidth}{!}{%
\begin{tabular}{llcrrrcrrrrrrr}
\toprule
 & & & \multicolumn{3}{c}{\textbf{Overall}} & & \multicolumn{7}{c}{\textbf{World-space 3D tracking accuracy \dmetric{W} per subset}}\\
\cmidrule(lr){4-6}\cmidrule(lr){8-14}
Method & Family & & \dmetric{W} & \dmetric{N} & \dmetric{Q} & & \textbf{DROID} & \textbf{EgoExo4D} & \textbf{Harmony4D} & \textbf{PACE} & \textbf{Hi4D} & \textbf{Waymo} & \textbf{Perpetua}\\
\midrule
OmniX~\cite{jiang2026omnix} & Multi-view joint &  & \textcolor{gray}{10.1} & \textcolor{gray}{9.3} & \textcolor{gray}{13.7} &  & 2.4 & 12.1 & 3.8 & 37.2 & 1.2 & \textcolor{gray}{6.6} & 7.4\\
MV-TAP~\cite{koo2025mvtap} & Multi-view 2D tracker &  & 20.0 & 18.7 & 23.5 &  & 5.8 & 47.7 & 18.9 & 40.7 & 17.0 & 2.1 & 7.8\\
MVTracker~\cite{rajic2025mvtracker} & Multi-view 3D tracker &  & \underline{22.8} & \textbf{21.7} & 24.7 &  & 5.2 & 48.8 & 23.2 & 43.7 & 19.1 & 3.4 & \textbf{16.1}\\
MVTracker (finetuned)~\cite{rajic2025mvtracker} & Multi-view 3D tracker &  & \textcolor{gray}{24.4} & \textcolor{gray}{23.3} & \textcolor{gray}{26.1} &  & \textbf{6.0} & \underline{50.7} & \textbf{24.7} & \textbf{45.0} & \textbf{20.8} & \textbf{3.7} & \textcolor{gray}{20.2}\\
CoWTracker~\cite{lai2026cowtracker} & Monocular 2D tracker &  & 21.5 & 18.6 & \underline{27.2} &  & \underline{5.9} & 46.5 & 21.9 & \underline{44.6} & 16.7 & 2.6 & 12.4\\
TAPIP3D~\cite{zhang2025tapip3d} & Monocular 3D tracker &  & \textbf{23.4} & \underline{21.5} & \textbf{27.3} &  & 5.8 & \textbf{51.1} & \underline{23.5} & 44.1 & \underline{20.5} & \underline{3.5} & \underline{15.5}\\
\midrule
\textcolor{gray}{TAPIP3D~\cite{zhang2025tapip3d} \emph{w/ GT depth}} & \textcolor{gray}{Monocular 3D tracker} &  & \textcolor{gray}{--} & \textcolor{gray}{--} & \textcolor{gray}{--} &  & \textcolor{gray}{39.3} & \textcolor{gray}{--} & \textcolor{gray}{--} & \textcolor{gray}{52.9} & \textcolor{gray}{53.8} & \textcolor{gray}{--} & \textcolor{gray}{43.2}\\
\bottomrule
\end{tabular}%
}
\end{table*}

\section{Baselines and Results}
\label{sec:models}

\begin{figure*}[t]
  \centering
  \includegraphics[width=\textwidth]{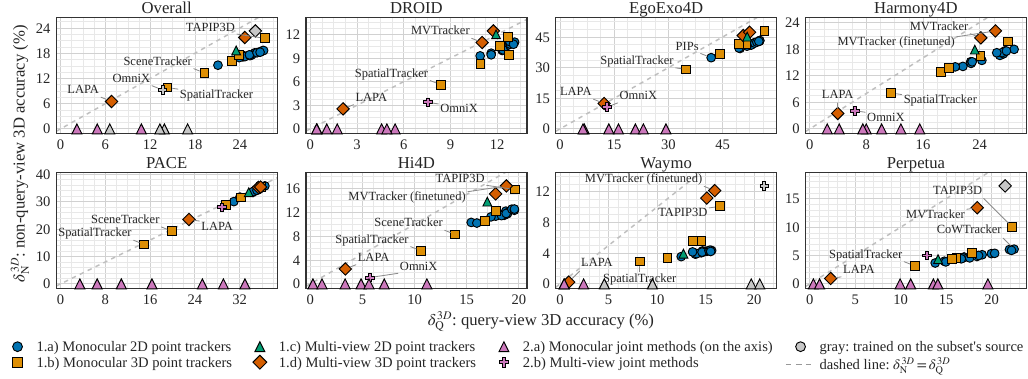}
  \vspace{-0.8cm}
  \caption{\textbf{Cross-view tracking collapses: query-view against non-query-view 3D accuracy.} Each marker is a method of \cref{tab:main_results} (point trackers on the shared VGGT-$\Omega$ reconstruction; the monocular joint methods have no \dmetric{N} and sit on the horizontal axis). The dashed line marks \dmetric{N} = \dmetric{Q}, and the vertical distance below it is what a method loses in the views it was not queried in. Methods sit below the line in nearly every case: tracking in the query view, the closest setting to monocular benchmarks, is far ahead of tracking the same points in the other views. The drop is largest where the viewpoints differ most (Perpetua, Waymo, Hi4D) and smallest on PACE, whose three rig-mounted cameras see nearly the same surfaces, so both metrics score nearly the same locations. The lifted monocular 2D point trackers form one tight cluster per subset, since the shared reconstruction determines most of their accuracy. The multi-view 3D point tracker MVTracker reaches the highest \dmetric{N} overall, but only marginally above the monocular TAPIP3D.}
  \label{fig:qn_scatter}
  \vspace{-0.5cm}
\end{figure*}

\paragraph{Method taxonomy.}
We adapt a broad set of existing models to \benchmarkname{} and place them in a taxonomy with two groups and six families, named as in the legend of \cref{fig:qn_scatter}. All of them track points. The groups differ in where the geometry comes from. Group~1 contains point trackers that receive a reconstruction as input: \textbf{monocular 2D point trackers} (1.a; 15 models), whose pixel trajectories we lift to 3D with the depth maps and cameras of that reconstruction, \textbf{monocular 3D point trackers} (1.b; 5 models), which consume its depth maps directly, the \textbf{multi-view 2D point tracker} MV-TAP (1.c), whose per-view pixel trajectories we lift by triangulation and by backprojection, and the \textbf{multi-view 3D point tracker} MVTracker (1.d). Every model in group~1 receives the same reconstruction, predicted from all views by the feed-forward model VGGT-$\Omega$~\cite{wang2026vggtomega}, so that differences between them come from the tracker alone. Group~2 contains joint methods, which estimate the geometry and the trajectories in one model without any reconstruction as input: \textbf{monocular joint methods} (2.a; 7 models) process each view separately and place its tracks in that view's own coordinate frame, whereas the \textbf{multi-view joint method} OmniX (2.b) processes all views and places the tracks in one shared frame. The numbering (1.a., 1.b., etc.) index the appendix table \ref{tab:non_query_view_thresholds}. \Cref{sec:baseline_impl} details each adaptation and \cref{tab:training_data} the training data.

\begin{figure*}[t]
  \centering
  \includegraphics[width=\textwidth]{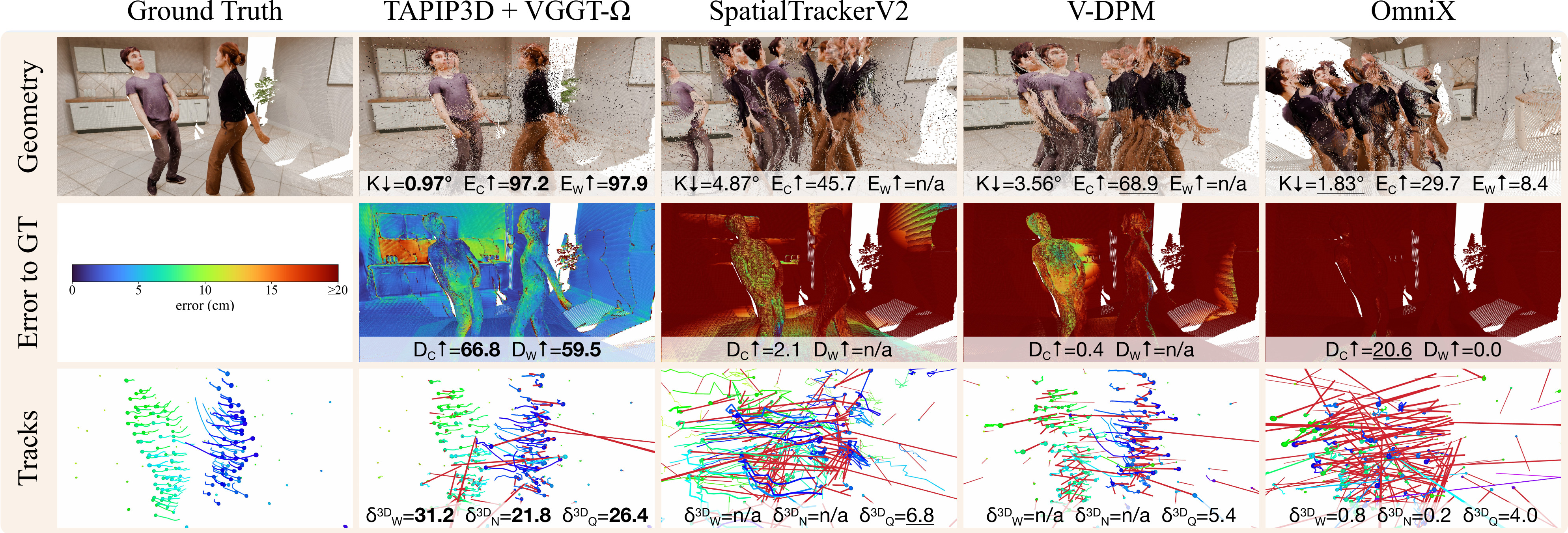}
  \vspace{-0.6cm}
  \caption{\textbf{Qualitative comparison on Hi4D.} Ground truth next to four methods; \cref{fig:qualitative_droid_pace,fig:qualitative_full} show DROID, PACE, and more models. The \emph{Error to GT} row colors the ground-truth geometry by its distance to the model's reconstruction, and the \emph{Tracks} row connects each predicted 3D point to its ground-truth position with a red line, so longer lines mean larger errors. Reconstruction errors propagate into the tracks: the joint methods estimate their own geometry, with duplicated and misaligned bodies, and their tracks err visibly more, while TAPIP3D on the VGGT-$\Omega$ reconstruction stays closest to ground truth.}
  \vspace{-0.4cm}
  \label{fig:qualitative}
\end{figure*}

\paragraph{Evaluation protocol.}
Every baseline runs under the same protocol and metrics (\cref{sec:task}), on videos with 512 pixels on the longer side. Each query is a pixel location in a single view (the \emph{query view}) at a single timestep. Each point tracker runs twice, once forward in time and once on the time-reversed video, and we merge the two halves at the query timestep. See \cref{sec:baseline_impl} for the exceptions and the per-method alignment details. The monocular joint methods reconstruct each view in its own frame, so they can be scored only in the query view. Predictions are recovered up to scale, so we rescale each by one factor per sequence, the ratio of the median GT to the median predicted distance of the query points from the origin of the metric's frame. \Cref{fig:qualitative} shows each prediction on the geometry it was made in, aligned to the GT by a single Sim(3) fit (one fit per view for the monocular joint methods). The appendix reports full per-method results and reconstruction quality (\cref{tab:main_results,tab:reconstruction_quality}), results with GT and other depth sources (\cref{tab:gt_depth_results,tab:depth_source_results}), and per-threshold, world-space, and pixel-space metrics (\cref{tab:query_view_thresholds,tab:non_query_view_thresholds,tab:world_metrics,tab:recognition_metrics}).

\begin{figure*}[t]
  \centering
  \includegraphics[width=\textwidth]{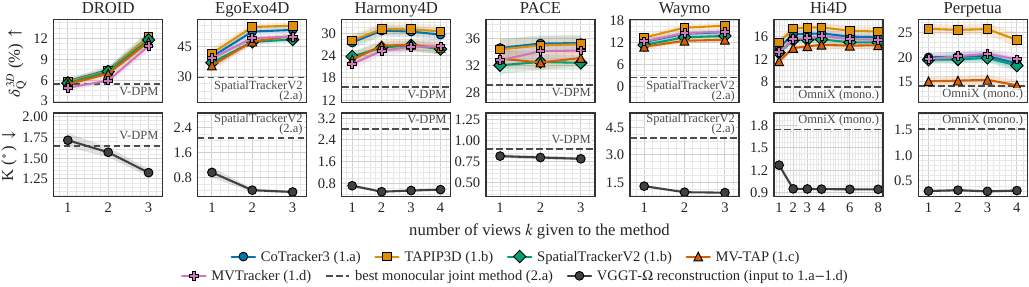}
  \vspace{-0.55cm}
  \caption{\textbf{More views improve tracking, even in the query view.} We vary the number of cameras $k$ that each method receives. Each evaluated scene is run $V$ times, once with each of its $V$ cameras as the query view plus the $k-1$ cameras nearest to it. The top row reports query-view tracking accuracy \dmetric{Q}, and the bottom row the intrinsics error K of the shared VGGT-$\Omega$ reconstruction. The query-view task is identical at every $k$, and the monocular trackers see the added cameras only through the reconstruction, yet \dmetric{Q} rises wherever the added views improve the reconstruction (DROID, EgoExo4D, PACE, Waymo, Hi4D), especially on DROID. Harmony4D and Perpetua exhibit mixed results for view scaling. Bands show $\pm 1$ standard error over scenes.}
  \label{fig:views_scaling}
\end{figure*}

\subsection{Results}
\label{sec:results}

\Cref{tab:headline_results} gives an overview of the results. The first headline is that the best overall system is TAPIP3D~\cite{zhang2025tapip3d} on the shared VGGT-$\Omega$~\cite{wang2026vggtomega} reconstruction,  reaching \dmetric{W} of 23.4 averaged over the seven subsets. The second is that scores are poor for two subsets with pressing real-world applications, so no method reaches above 6.6 on driving (Waymo), nor 6.0 on robot manipulation (DROID). These are not weak baselines, as the same trackers top existing monocular 2D and 3D benchmarks~\cite{doersch2022tapvid,koppula2024tapvid3d}. Accordingly, in contrast to world space, they do well on our query-view metric \dmetric{Q}. Our community has built good foundations: strong monocular point trackers, accurate feed-forward multi-view reconstruction, and synchronized multi-view training data. But current integrations are disappointing. We walk through the families in the order a reader might expect them to solve the task. \cref{fig:qualitative,fig:qualitative_droid_pace} show the challenges.

\paragraph{Joint multi-view models fail at their own geometry.}
The most direct solution is a multi-view joint method. OmniX, the only released one, reaches \dmetric{W} of 10.1 overall and 1.2 on Hi4D. The failure is in its geometry, not its tracking head: on dynamic multi-view footage its cross-view registration collapses, at E\textsubscript{W} of 21.0 on Hi4D against 95.6 for VGGT-$\Omega$ (\cref{tab:reconstruction_quality}). \cref{fig:qualitative} shows the duplicated, misaligned human surfaces. Part of the gap stems from training on 16-image samples \vs generalizing to our longer clips (\cref{sec:baseline_impl}). Frustratingly, multi-view input barely changes OmniX over its monocular variant (13.7 against 13.3 \dmetric{Q}).

\paragraph{Purpose-built multi-view trackers barely use the views.}
The next most natural approach is to use a dedicated multi-view tracker. MVTracker on the shared VGGT-$\Omega$ reconstruction leads on non-query-view accuracy, but not by much: \dmetric{N} of 22.2 against 22.0 for the strictly monocular TAPIP3D, while trailing it on \dmetric{Q} (25.6 against 28.5). MVTracker is trained only on multi-view Kubric and overfits to its scene scales and layouts, despite scene normalization. MV-TAP trails the best lifted monocular 2D trackers on every metric (\dmetric{W} of 20.9 against 22.4), and its multi-view machinery pays off only with ground-truth depth (where it leads on DROID) or oracle multi-view queries (\cref{tab:gt_depth_results,tab:main_results}). Cross-view fusion, the one capability this family adds, yields almost nothing on real moving-camera footage.

\paragraph{Monocular point trackers win, and their ceiling is the reconstruction.}
What remains is the pragmatic pipeline: a monocular tracker in 2D or camera-space 3D, placed in the world by an off-the-shelf reconstruction. This family wins nearly every subset, yet the geometry, not the tracker, sets its accuracy. On one reconstruction, twenty trackers spanning four years land within 3.1 points on DROID (2.8 to 5.9 \dmetric{W}), while changing the geometry moves them far more. Ground-truth depth would lift TAPIP3D from 5.8 to 39.3 on DROID and from 20.5 to 53.8 on Hi4D (\cref{tab:gt_depth_results}). The SpatialTrackerV2 comparison in \cref{fig:recon_scatter} shows how different geometry doubles its accuracy. The monocular joint methods mark the lower end of the same axis, with the least accurate geometry (\cref{fig:recon_scatter}). Even this ceiling is low (39.3 on DROID with perfect depth), so better reconstruction alone cannot close the gap.

\paragraph{Added views raise the ceiling, even in the query view.}
More views offers more information about  a scene and should improve reconstruction. In \cref{fig:views_scaling}, we start by reconstructing from a single camera, and then increase the number of views $k$ each method receives, measuring whether reconstruction in that camera view \dmetric{Q} improves. TAPIP3D rises from 5.7 to 7.3 to 12.2 \dmetric{Q} on DROID with one, two, and three views. The gain is gated on the geometry improving, and Harmony4D and Perpetua, where tracking failures dominate, show mixed scaling. Many of our baselines use additional views by fusing single-view tracking results via geometry. This is a modest use of a second camera. A model that also fuses evidence across views should gain far more, but as we show next, none does.

\paragraph{Cross-view correspondence is the unsolved core.}
Innovations are needed, even with perfect geometry, to get markers in \cref{fig:qn_scatter}, closer to the diagonal by improving accuracy in the \emph{not} queried views. The best lifted monocular 2D tracker, AllTracker, drops from 28.1 (\dmetric{Q}) to 18.8 (\dmetric{N}) overall, falling most where viewpoints differ most, from 25.9 to 6.3 on Perpetua and 15.3 to 4.3 on Waymo. This is not a reconstruction artifact. With ground-truth depth on Perpetua it still falls from 63.4 \dmetric{Q} to 19.2 \dmetric{N}, while the multi-view MVTracker manages 44.2 \dmetric{N}, so better cross-view fusion can preserve accuracy across cameras.  There is a real need for methods that both track well and fuse views. World space adds a harder rung: where global registration fails, \dmetric{W} falls below even \dmetric{N}, as for TAPIP3D on Waymo (\dmetric{Q}/\dmetric{N}/\dmetric{W} of 16.5/10.1/3.5).

\paragraph{Recurring failure patterns.}
We observe five recurring behaviors in the qualitative comparisons (\cref{fig:failure_modes}). First, a lifted 2D track reads its depth at the predicted pixel, so a drift of one pixel across an object boundary moves the 3D point by the full depth difference between the two surfaces, and under occlusion the only depth available belongs to the occluder. Second, point trackers often drift along the tracked surface over time and jump to a neighboring surface when two surfaces come into contact. For example, TAPIP3D on VGGT-$\Omega$ jumps from one leg to the other in Hi4D and EgoExo4D, without recovering afterwards. Third, errors in the reconstruction propagate into the predicted tracks, such as jitter, the tilt of walls and floors in Perpetua, and humans fused into the background in Harmony4D.  Fourth, out-of-distribution visuals easily break the priors from the training data. For example, on the original Hi4D captures the nearly symmetric green-screen studio breaks cross-view registration despite eight highly redundant views (\cref{sec:hi4d_static_only}).  Fifth, trackers fail once points are no longer visible in the query view, for example in Waymo where cameras point in widely differing directions.

\paragraph{What \benchmarkname{} reveals.}
These new challenges are finally apparent. Monocular benchmarks score only the query view~\cite{koppula2024tapvid3d}, exactly where every method looked healthiest. Existing multi-view evaluations provide fixed, known extrinsics from static cameras~\cite{rajic2025mvtracker,koo2025mvtap,galoaa2025lookaround}, removing the registration and cross-view problems that dominate here. \benchmarkname{} separates the failure modes by construction: the \dmetric{Q}/\dmetric{N}/\dmetric{W} triplet localizes an error to monocular tracking, cross-view correspondence, or global registration. Pairing every tracking score with reconstruction metrics on the same sequences (\cref{fig:recon_scatter}) separates the tracker from the geometry it stands on. The result is a diagnosis previous evaluations could not produce: while monocular tracking is strong, reconstruction of dynamic multi-view scenes puts a ceiling on today's best pipeline, and cross-view correspondence is where current architectures, modular and joint alike, falter.

\subsection{Implications for Future Work}
\label{sec:discussion}
We recommend reporting reconstruction quality alongside every 3D tracking result, measured on the same sequences, because a tracking score confounds the tracker with the geometry beneath it (\cref{fig:recon_scatter}). We also recommend controlling for training data when a new architecture is proposed: every method we evaluate is trained on a different mix of reconstruction and point tracking supervision (\cref{tab:training_data}), so an architectural gain cannot be separated from a data gain.

For 2D and 3D tracking models, we find that long-horizon temporal drift and identity swaps during multi-object contact remains largely unsolved even when using the most accurate reconstruction (\cref{fig:failure_modes}). This is important for applications that follow a point through multiple object interaction.

Training data remains a bottleneck for both 4D reconstruction and point tracking. Synchronized multi-view dynamic supervision is becoming available, synthetically in Syn4D~\cite{jiang2026syn4d}, multi-view Kubric~\cite{rajic2025mvtracker,koo2025mvtap}, and the OmniX data engine~\cite{jiang2026omnix}, and in real LiDAR-annotated driving footage~\cite{sun2020waymo,balasingam2024drivetrack}. Training on them jointly would show how far current architectures can be taken, and the transfer we observe from PointOdyssey training to Perpetua (\cref{sec:additional_results}) suggests such supervision generalizes across synthetic domains. Fine-tuning MVTracker on a synthetic training mix containing Perpetua improves all three aggregate metrics and performance on every real subset (\cref{tab:main_results}), demonstrating that synthetic multi-view supervision transfers to real scenes. We add two such sources: the procedural Perpetua generator and our DROID pipeline, which annotates 5,371 episodes with metric 3D trajectories (\cref{sec:supp_droid}).

\section{Conclusions}
We introduce \benchmarkname, a curated benchmark for multi-view 3D point tracking. We show how \benchmarkname{} helps in diagnosing deficiencies in the current approaches to 3D TAP. The benchmark can be used for multiple 3D tasks, including posing, reconstruction, and far more. We discuss limitations of \benchmarkname{} in \cref{sec:limitations}. We release all artifacts, so baselines, data, data generation tools, evaluation protocols, and invite the community to use these to advance model capabilities.

\section*{Acknowledgements}
This work was supported under project ID \#169 as part of the Swiss AI Initiative, through a grant from the ETH Domain and computational resources provided by the Swiss National Supercomputing Centre (CSCS) under the Alps infrastructure. It was also made possible by a generous GPU grant from Modal (\url{https://modal.com/}). We thank Goker Erdogan, David Fleet, Mehdi Sajjadi, and Ross Murphy for their advice, insights, and feedback throughout the course of the work. We also thank Juan José Zarate for discussions and assistance concerning Hi4D.

{
    \small
    \bibliographystyle{ieeenat_fullname}
    \bibliography{main}
}
\clearpage
\hypersetup{pageanchor=false}
\providecommand{\theHpage}{\arabic{page}}
\renewcommand{\theHpage}{supp.\arabic{page}}
\maketitlesupplementary
\appendix
\setcounter{section}{0}
\setcounter{table}{0}
\setcounter{figure}{0}
\setcounter{equation}{0}
\renewcommand{\thetable}{\thesection.\arabic{table}}
\renewcommand{\thefigure}{\thesection.\arabic{figure}}
\renewcommand{\theequation}{\thesection.\arabic{equation}}
\makeatletter
\@addtoreset{table}{section}
\@addtoreset{figure}{section}
\@addtoreset{equation}{section}
\makeatother

\etocsettocdepth.toc{section}
{\small
  \etocsetnexttocdepth{section}
  \etocsettocstyle{\section*{Appendix Contents}\vspace{-0.15cm}}{}
  \etocsetstyle{section}
    {}{}
    {\noindent\etocnumber\hspace{1em}\etocname\dotfill\etocpage\par}
    {}
  \tableofcontents
}
\vspace{0.15cm}

\section{Additional Results}
\label{sec:additional_results}
\begin{figure*}[t]
  \centering
  \includegraphics[width=\textwidth]{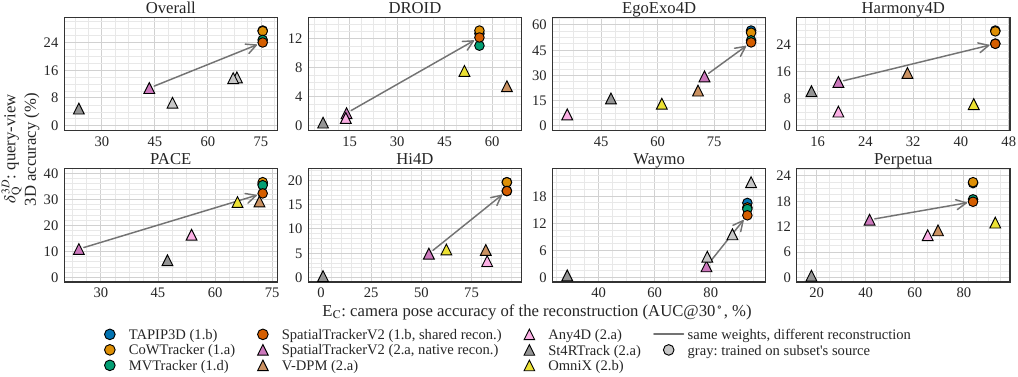}
  \vspace{-0.55cm}
  \caption{\textbf{Tracking accuracy is set by reconstruction accuracy.} Query-view accuracy \dmetric{Q} against the camera pose accuracy E\textsubscript{C} of the reconstruction each system tracks in (\cref{tab:reconstruction_quality}). Circles are point trackers on the shared VGGT-$\Omega$ reconstruction, so they share one $x$ per panel; triangles are joint methods on their own geometry. Accuracy rises with reconstruction quality, and trackers on one reconstruction differ far less than reconstructions do: the gray arrow hands the same SpatialTrackerV2 weights the VGGT-$\Omega$ geometry instead of their own and more than doubles \dmetric{Q} (10.8 to 23.9). Good geometry is necessary but not sufficient: on PACE, V-DPM matches VGGT-$\Omega$ in pose accuracy yet tracks below the point trackers running on it.}
  \label{fig:recon_scatter}
\end{figure*}

\begin{figure*}[t]
  \centering
  \includegraphics[width=\textwidth]{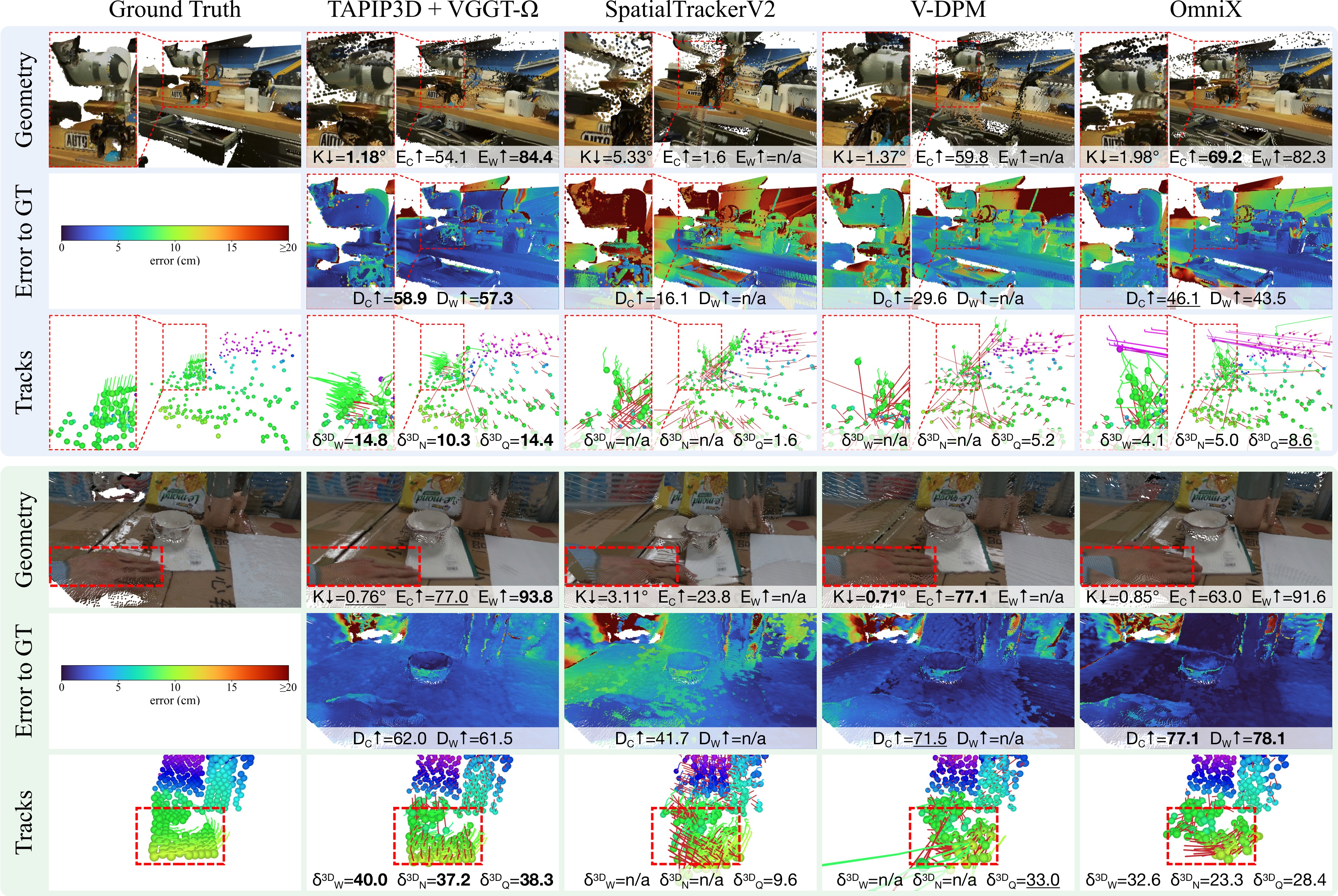}
  \vspace{-0.6cm}
  \caption{\textbf{Qualitative comparison on DROID (rows 1--3) and PACE (rows 4--6).} The companion of \cref{fig:qualitative}, with the same rows and conventions. Reconstruction errors propagate into the tracks, especially on DROID, where the joint methods misplace the gripper and its tracks, while all methods recover the rigid PACE tabletop and differ mainly on the moving hand.}
  \label{fig:qualitative_droid_pace}
\end{figure*}

\begin{figure*}[p]
  \centering
  \includegraphics[width=\textwidth,height=0.86\textheight,keepaspectratio]{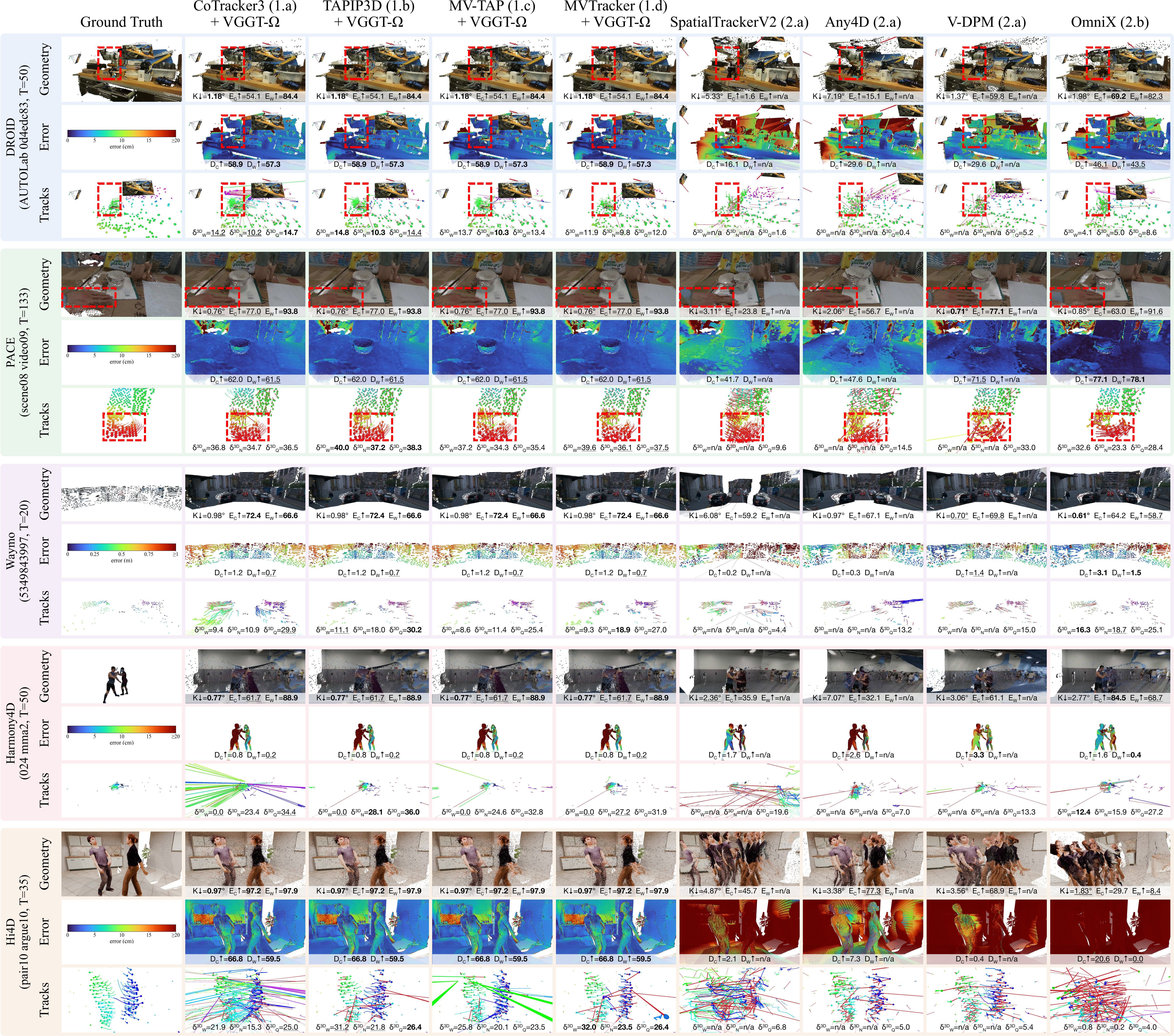}
  \caption{\textbf{Qualitative comparison with additional models.} The figure extends \cref{fig:qualitative,fig:qualitative_droid_pace} with additional models of \cref{tab:main_results}. The rows and colors follow the same conventions, and this figure additionally shows each method's estimated cameras, with red lines to their ground-truth poses, the group of every method, and the sequence and timestep of every subset. The color bar spans 1\,m on Waymo and 20\,cm elsewhere. CoTracker3, TAPIP3D, MV-TAP, and MVTracker all receive the same VGGT-$\Omega$ reconstruction, so their Geometry and Error to GT rows and reconstruction scores are identical. The lifted 2D trackers additionally show the depth jumps typical of lifting at depth edges and under occlusion. Harmony4D lacks ground-truth depth dense enough to constrain a point-cloud fit, so predictions there are aligned by matching the estimated camera centers to the ground-truth ones instead.}
  \label{fig:qualitative_full}
\end{figure*}

We provide the full baseline comparison behind the headline results of \cref{tab:headline_results}, along with additional breakdowns. \Cref{tab:main_results} reports all baselines on every subset, \cref{tab:reconstruction_quality} evaluates their reconstructions, \cref{tab:gt_depth_results} replaces estimated geometry with ground-truth depth, and \cref{tab:depth_source_results} varies the reconstruction source. \Cref{tab:query_view_thresholds,tab:non_query_view_thresholds,tab:world_metrics} provide per-threshold query-view, non-query-view, and world-space results, while \cref{tab:recognition_metrics} reports pixel-space tracking and visibility. \Cref{fig:recon_scatter} plots tracking accuracy against the quality of the reconstruction each system tracks in, summarizing \cref{tab:reconstruction_quality}, \cref{fig:qualitative_droid_pace} shows the DROID and PACE examples of the qualitative comparison of \cref{fig:qualitative}, and \cref{fig:qualitative_full} extends both with additional models.

\renewcommand{\dblfloatpagefraction}{0.7}
\renewcommand{\floatpagefraction}{0.7}

\begin{figure*}[p]
  \centering
  \includegraphics[width=\textwidth]{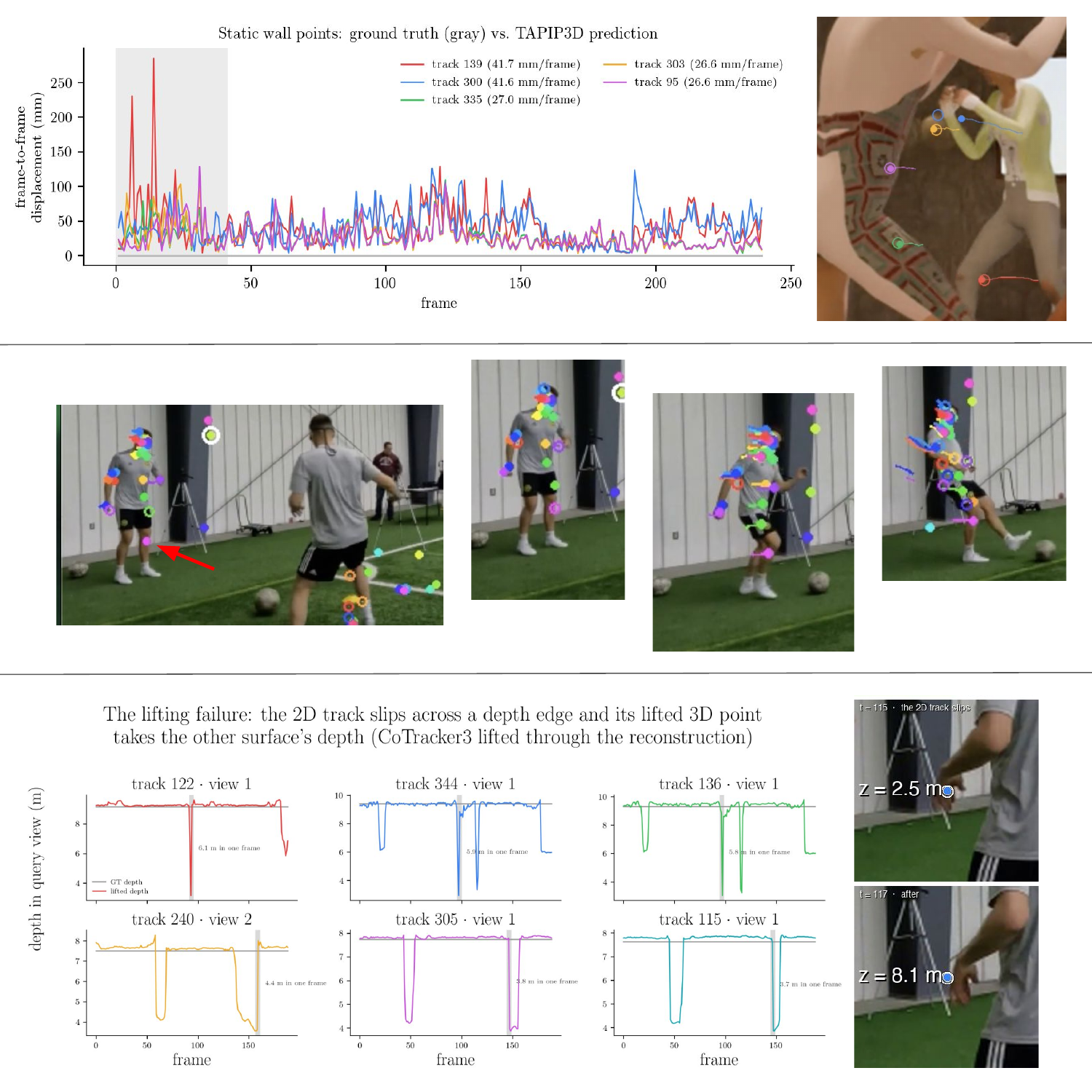}
  \vspace{-0.4cm}
  \caption{\textbf{Failure modes shared across methods.} \emph{Top:} points on a static wall in Perpetua inherit the jitter of the reconstruction. Their ground-truth frame-to-frame displacement is zero (gray), while the same points predicted by TAPIP3D move by 27 to 42\,mm per frame throughout the sequence. These five sample points are visualized in a freeze frame on the right, with the ground-truth location indicated by the open circle and the predicted location by the filled-in circle and its trajectory. \emph{Middle:} four frames of an EgoExo4D sequence in temporal order. The red arrow marks a track on one leg of the near player: as the legs cross and touch, it switches from leg to leg through the rest of the video. \emph{Bottom:} a lifted 2D track slips across a depth edge, so its 3D point takes the depth of the surface behind it, here jumping from 2.5 to 8.1\,m and back. The depth of six such tracks is plotted against time.}
  \label{fig:failure_modes}
\end{figure*}

\paragraph{PointOdyssey training transfers to Perpetua.}
The three recent monocular 2D point trackers trained on PointOdyssey (AllTracker, TAPNext++, and CoWTracker, see \cref{tab:training_data}) are the most accurate 2D point trackers on Perpetua. They reach \dmetric{Q} of 21.9 to 22.5 against 20.3 for the best 2D point tracker not trained on PointOdyssey (MegaFlow). OmniX, also trained on PointOdyssey, ties for the highest per-view camera accuracy and has the best per-view depth accuracy on Perpetua (E\textsubscript{C} of 92.8 and D\textsubscript{C} of 28.0 in \cref{tab:reconstruction_quality}). This indicates transfer from PointOdyssey to Perpetua despite the two differing in environments, motions, and actors.

\paragraph{Evaluations used by recent work.}
Recent joint 4D models compare against weaker modular pipelines than the ones we evaluate. Any4D, D4RT, and 4RC lift CoTracker3 through VGGT or MapAnything and compare against that~\cite{karhade2025any4d,zhang2026efficiently,luo2026fourrc}, so their modular baseline receives a current reconstruction but only a 2D point tracker, and none of the three evaluates a 3D point tracker on it. OmniX and V-DPM report no modular baseline at all and compare only against other joint models~\cite{jiang2026omnix,sucar2026vdpm}. Any4D, D4RT, 4RC, and OmniX score reconstruction on different datasets from the ones they track on, so neither factor can be attributed. V-DPM scores reconstruction and tracking on one set of four datasets, but merges them into a single point-map error and trains on those four datasets~\cite{sucar2026vdpm}. None of these comparisons controls for training data either, and the mixtures differ widely. \Cref{tab:training_data} records the training data of every baseline we run, and \cref{tab:main_results} marks in gray the cells where a method was trained on that subset's source data.

\begin{table*}[!t]
\centering
\setlength{\tabcolsep}{1.8pt}
\caption{\textbf{Full TAPVid-MV tracking results with off-the-shelf or self-estimated multi-view depth maps.} The table expands the headline comparison of \cref{tab:headline_results} to every baseline. Higher is better for all displayed metrics. \dmetric{W} reports world-space 3D tracking accuracy, and \dmetric{N} and \dmetric{Q} report camera-space 3D tracking accuracy in the non-query views and in the query view. \dmetric{W} and \dmetric{N} require multi-view consistency, while \dmetric{Q} is closest to monocular 3D tracking. 3D tracking metrics use query-median scaling, and the quality of the cameras and depth each method relies on is reported in \cref{tab:reconstruction_quality}. The Overall column averages each metric over the seven datasets. Best and second-best entries among non-gray results are bold and underlined. Gray cells mark a method trained on that subset's source data. The gray MV-TAP rows use privileged oracle multi-timestep (\(^{\dagger}\), MT) and multi-view (\(^{\ddagger}\), MV) ground-truth queries. Baseline implementation details are given in \cref{sec:baseline_impl}. The last column reports the median wall-clock runtime per multi-view frame (one timestep of all 8 Hi4D views) on one NVIDIA GH200 GPU, excluding model loading. For trackers conditioned on the shared reconstruction, it excludes the shared VGGT-$\Omega$ reconstruction, which itself takes about 1.9\,s per frame, while methods that estimate their own geometry include that estimation. AllTracker, DOT, MegaFlow, CoWTracker, and St4RTrack cannot jointly process queries originating at different timesteps and are rerun for every distinct query timestep (a median of 69 times per Hi4D sequence). Their reported runtime includes all such passes. V-DPM is slow for a different reason: it decodes every target frame conditioned on all video frames.}
\label{tab:main_results}
\vspace{-0.27cm}
\resizebox{\linewidth}{!}{%
\begin{tabular}{lcrrr@{\hspace{6pt}}crrrcrrrcrrrcrrrcrrrcrrrcrrrcr}
\toprule
 &  & \multicolumn{3}{c}{\textbf{Overall}} &  & \multicolumn{3}{c}{\textbf{DROID}} &  & \multicolumn{3}{c}{\textbf{EgoExo4D}} &  & \multicolumn{3}{c}{\textbf{Harmony4D}} &  & \multicolumn{3}{c}{\textbf{PACE}} &  & \multicolumn{3}{c}{\textbf{Hi4D}} &  & \multicolumn{3}{c}{\textbf{Waymo}} &  & \multicolumn{3}{c}{\textbf{Perpetua}} &  & \multicolumn{1}{c}{Time}\\
\cmidrule(lr){3-5}\cmidrule(lr){7-9}\cmidrule(lr){11-13}\cmidrule(lr){15-17}\cmidrule(lr){19-21}\cmidrule(lr){23-25}\cmidrule(lr){27-29}\cmidrule(lr){31-33}\cmidrule(lr){35-35}
Method &  & \dmetric{W} & \dmetric{N} & \dmetric{Q} &  & \dmetric{W} & \dmetric{N} & \dmetric{Q} &  & \dmetric{W} & \dmetric{N} & \dmetric{Q} &  & \dmetric{W} & \dmetric{N} & \dmetric{Q} &  & \dmetric{W} & \dmetric{N} & \dmetric{Q} &  & \dmetric{W} & \dmetric{N} & \dmetric{Q} &  & \dmetric{W} & \dmetric{N} & \dmetric{Q} &  & \dmetric{W} & \dmetric{N} & \dmetric{Q} &  & (s/frame)\\
\midrule
\addlinespace[0.15em]
\multicolumn{35}{l}{1.a) \textbf{Monocular 2D point trackers} \emph{(lifted with VGGT-$\Omega$~\cite{wang2026vggtomega})}}\\
\midrule
PIPs~\cite{harley2022particle} &  & 17.0 & 15.1 & 21.1 &  & 4.8 & 9.3 & 10.9 &  & 36.9 & 34.9 & 41.8 &  & 16.9 & 14.0 & 20.6 &  & 36.6 & 30.0 & 31.0 &  & 13.5 & 10.3 & 15.4 &  & 2.2 & 3.9 & 13.9 &  & 7.7 & 3.7 & 13.8 &  & 1.06\\
PIPs++~\cite{zheng2023pointodyssey} &  & 18.4 & 16.3 & 23.0 &  & 5.1 & 9.6 & 11.6 &  & 42.5 & 39.4 & 49.8 &  & 18.4 & 15.5 & 24.3 &  & 38.5 & 31.6 & 32.1 &  & 13.8 & 10.2 & 16.0 &  & 2.0 & 3.5 & 12.5 &  & 8.3 & 4.0 & 14.8 &  & 1.05\\
TAPTRv3~\cite{li2024taptrv3} &  & 19.3 & 16.9 & 23.9 &  & 5.4 & 10.0 & 12.4 &  & 43.7 & 40.5 & 51.8 &  & 18.5 & 15.1 & 22.9 &  & 41.7 & 33.7 & 34.4 &  & 15.1 & 11.2 & 17.3 &  & 2.3 & 4.2 & 13.7 &  & 8.3 & 3.9 & 15.0 &  & 0.81\\
CoTracker2~\cite{karaev2024cotracker} &  & 19.4 & 17.1 & 24.4 &  & 5.5 & 10.4 & 12.3 &  & 44.5 & 41.2 & 52.5 &  & 16.4 & 14.1 & 21.6 &  & 42.0 & 34.0 & 34.7 &  & 15.4 & 11.4 & 17.8 &  & 2.5 & 4.2 & 15.0 &  & 9.5 & 4.6 & 16.9 &  & 0.51\\
TAPIR~\cite{doersch2023tapir} &  & 19.5 & 17.2 & 24.7 &  & 5.0 & 9.4 & 11.6 &  & 42.2 & 39.6 & 49.5 &  & 20.3 & 16.6 & 26.9 &  & 41.7 & 34.6 & 35.2 &  & 15.7 & 11.7 & 18.7 &  & 2.3 & 3.9 & 14.1 &  & 9.2 & 4.4 & 16.7 &  & 0.03\\
CoTracker1~\cite{karaev2024cotracker} &  & 19.7 & 17.4 & 24.6 &  & 5.5 & 10.4 & 12.5 &  & 44.4 & 41.3 & 52.1 &  & 17.8 & 15.1 & 22.7 &  & 42.2 & 34.1 & 34.8 &  & 15.7 & 11.7 & 18.0 &  & 2.4 & 4.3 & 14.7 &  & 9.6 & 4.7 & 17.0 &  & 0.29\\
BootsTAPIR~\cite{doersch2024bootstap} &  & 20.0 & 17.4 & 25.3 &  & 5.3 & 10.1 & 12.4 &  & 44.0 & 41.0 & 51.5 &  & 20.8 & 17.0 & 27.4 &  & 41.6 & 33.4 & 34.4 &  & 15.6 & 11.5 & 18.3 &  & 2.5 & 4.2 & 15.1 &  & 10.1 & 4.8 & 18.4 &  & 0.04\\
DOT~\cite{lemoing2024dense} &  & 20.0 & 17.6 & 25.3 &  & 5.8 & 11.0 & 12.8 &  & 45.6 & 42.0 & 53.9 &  & 17.3 & 14.8 & 22.8 &  & 42.8 & 34.4 & 35.2 &  & 16.0 & 12.0 & 18.9 &  & 2.4 & 4.2 & 14.8 &  & 10.3 & 5.0 & 18.6 &  & 12.18\\
LocoTrack~\cite{cho2024locotrack} &  & 20.2 & 17.6 & 25.6 &  & 5.3 & 10.0 & 12.3 &  & 44.1 & 41.0 & 52.9 &  & 20.6 & 16.8 & 27.2 &  & 43.5 & 35.0 & 35.8 &  & 15.7 & 11.5 & 18.6 &  & 2.4 & 4.0 & 14.6 &  & 9.7 & 4.6 & 17.6 &  & 0.04\\
CoTracker3~\cite{karaev2024cotracker3} &  & 20.6 & 18.1 & 26.0 &  & 5.7 & 10.7 & 12.7 &  & 45.4 & 42.2 & 53.8 &  & 21.2 & 17.4 & 27.5 &  & 42.8 & 34.7 & 35.3 &  & 16.3 & 12.2 & 18.7 &  & 2.4 & 4.2 & 14.8 &  & 10.6 & 5.1 & 18.9 &  & 0.19\\
MegaFlow~\cite{zhang2026megaflow} &  & 20.9 & 18.1 & 26.3 &  & 5.7 & 10.8 & 12.8 &  & 45.8 & 42.5 & 54.5 &  & 21.2 & 17.2 & 27.2 &  & 43.7 & 34.9 & 35.7 &  & 16.0 & 11.8 & 18.8 &  & 2.5 & 4.1 & 14.6 &  & 11.1 & 5.4 & 20.3 &  & 37.15\\
TAPNext++~\cite{jung2026tapnextpp} &  & 21.0 & 18.2 & 26.7 &  & 5.7 & 10.8 & \underline{13.0} &  & 46.1 & 42.9 & 55.0 &  & 21.3 & 17.2 & 26.4 &  & 42.5 & 34.1 & 35.1 &  & 16.8 & 12.6 & \textbf{19.6} &  & 2.6 & 4.2 & 15.5 &  & 12.2 & 5.9 & \underline{22.2} &  & 0.45\\
Track-On-R~\cite{aydemir2025trackon} &  & 21.1 & 18.3 & 26.6 &  & 5.8 & 10.9 & \underline{13.0} &  & 46.2 & 42.9 & 55.2 &  & 21.8 & 17.6 & 27.6 &  & 44.0 & 35.2 & \underline{36.0} &  & 16.1 & 11.8 & 18.9 &  & 2.6 & 4.3 & 15.6 &  & 11.1 & 5.3 & 20.0 &  & 0.99\\
AllTracker~\cite{harley2025alltracker} &  & 21.4 & 18.6 & 27.1 &  & 5.8 & 11.1 & 12.8 &  & 46.5 & 43.2 & \underline{55.3} &  & 22.4 & 17.9 & \textbf{28.9} &  & 43.9 & 35.1 & 35.9 &  & 16.7 & 12.5 & \underline{19.3} &  & 2.5 & 4.3 & 15.3 &  & 12.2 & 6.1 & 21.9 &  & 2.10\\
CoWTracker~\cite{lai2026cowtracker} &  & 21.5 & 18.6 & \underline{27.2} &  & \underline{5.9} & 11.0 & \textbf{13.1} &  & 46.5 & 43.3 & \underline{55.3} &  & 21.9 & 17.6 & 27.8 &  & \underline{44.6} & \textbf{35.7} & \textbf{36.5} &  & 16.7 & 12.4 & \textbf{19.6} &  & 2.6 & 4.4 & 15.5 &  & 12.4 & 6.1 & \textbf{22.5} &  & 16.36\\
\midrule
\addlinespace[0.15em]
\multicolumn{35}{l}{1.b) \textbf{Monocular 3D point trackers} \emph{(with VGGT-$\Omega$~\cite{wang2026vggtomega} depth input)}}\\
\midrule
SpatialTracker~\cite{xiao2024spatialtracker} &  & 10.7 & 9.8 & 14.3 &  & 2.8 & 5.6 & 8.4 &  & 31.7 & 29.1 & 34.9 &  & 8.9 & 8.1 & 11.5 &  & 16.5 & 14.4 & 15.0 &  & 7.8 & 5.5 & 10.6 &  & 1.5 & 2.9 & 8.2 &  & 5.7 & 3.2 & 11.5 &  & 0.40\\
SceneTracker~\cite{wang2024scenetracker} &  & 14.7 & 13.3 & 19.2 &  & 4.2 & 8.3 & 10.9 &  & 39.4 & 36.6 & 44.2 &  & 15.4 & 12.9 & 18.6 &  & 22.6 & 19.2 & 19.9 &  & 11.1 & 8.3 & 13.9 &  & 2.0 & 3.4 & 11.1 &  & 8.3 & 4.5 & 16.1 &  & 0.15\\
DELTA~\cite{ngo2025delta} &  & 17.8 & 16.2 & 22.9 &  & 4.8 & 9.3 & 12.7 &  & 44.7 & 41.3 & 51.4 &  & 15.3 & 13.7 & 19.6 &  & 35.0 & 28.7 & 29.5 &  & 13.6 & 10.6 & 16.8 &  & 2.8 & 5.6 & 14.6 &  & 8.5 & 4.4 & 15.6 &  & 0.33\\
SpatialTrackerV2~\cite{xiao2025spatialtrackerv2} &  & 19.5 & 17.6 & 23.9 &  & 5.5 & 10.5 & 12.2 &  & 44.3 & 41.5 & 49.5 &  & 19.6 & 16.4 & 24.1 &  & 38.2 & 31.5 & 32.2 &  & 16.1 & 12.3 & 17.8 &  & 2.5 & 5.6 & 13.7 &  & 10.3 & 5.5 & 17.8 &  & 0.51\\
TAPIP3D~\cite{zhang2025tapip3d} &  & \textbf{23.4} & \underline{21.5} & \textbf{27.3} &  & 5.8 & 11.6 & 12.7 &  & \textbf{51.1} & \textbf{48.0} & \textbf{56.6} &  & \underline{23.5} & 19.7 & \underline{28.0} &  & 44.1 & \underline{35.5} & 35.8 &  & \underline{20.5} & \underline{15.8} & \textbf{19.6} &  & \underline{3.5} & 10.1 & \textbf{16.5} &  & \underline{15.5} & \underline{10.0} & \underline{22.2} &  & 1.56\\
\midrule
\addlinespace[0.15em]
\multicolumn{35}{l}{1.c) \textbf{Multi-view 2D point trackers} \emph{(lifted with VGGT-$\Omega$~\cite{wang2026vggtomega})}}\\
\midrule
MV-TAP~\cite{koo2025mvtap} &  & 20.0 & 18.7 & 23.5 &  & 5.8 & \underline{12.0} & 11.9 &  & 47.7 & 45.2 & 51.8 &  & 18.9 & 17.9 & 23.3 &  & 40.7 & 33.4 & 33.6 &  & 17.0 & 13.8 & 17.0 &  & 2.1 & 3.9 & 12.7 &  & 7.8 & 4.4 & 14.1 &  & 6.24\\
\textcolor{gray}{MV-TAP (MV)$^\ddagger$~\cite{koo2025mvtap}} &  & \textcolor{gray}{23.9} & \textcolor{gray}{24.2} & \textcolor{gray}{26.1} &  & \textcolor{gray}{8.2} & \textcolor{gray}{17.5} & \textcolor{gray}{14.1} &  & \textcolor{gray}{59.4} & \textcolor{gray}{62.1} & \textcolor{gray}{61.0} &  & \textcolor{gray}{23.1} & \textcolor{gray}{23.9} & \textcolor{gray}{25.9} &  & \textcolor{gray}{43.7} & \textcolor{gray}{36.0} & \textcolor{gray}{36.3} &  & \textcolor{gray}{22.9} & \textcolor{gray}{21.7} & \textcolor{gray}{23.0} &  & \textcolor{gray}{1.3} & \textcolor{gray}{2.5} & \textcolor{gray}{7.2} &  & \textcolor{gray}{8.5} & \textcolor{gray}{6.0} & \textcolor{gray}{15.3} &  & \textcolor{gray}{3.47}\\
\textcolor{gray}{MV-TAP (MT)$^\dagger$~\cite{koo2025mvtap}} &  & \textcolor{gray}{24.6} & \textcolor{gray}{26.0} & \textcolor{gray}{26.0} &  & \textcolor{gray}{8.2} & \textcolor{gray}{17.7} & \textcolor{gray}{14.1} &  & \textcolor{gray}{60.4} & \textcolor{gray}{63.5} & \textcolor{gray}{62.1} &  & \textcolor{gray}{22.9} & \textcolor{gray}{24.4} & \textcolor{gray}{24.9} &  & \textcolor{gray}{43.9} & \textcolor{gray}{36.1} & \textcolor{gray}{36.5} &  & \textcolor{gray}{24.0} & \textcolor{gray}{22.9} & \textcolor{gray}{22.9} &  & \textcolor{gray}{1.7} & \textcolor{gray}{4.6} & \textcolor{gray}{6.0} &  & \textcolor{gray}{11.2} & \textcolor{gray}{12.8} & \textcolor{gray}{15.3} &  & \textcolor{gray}{0.25}\\
\midrule
\addlinespace[0.15em]
\multicolumn{35}{l}{1.d) \textbf{Multi-view 3D point trackers} \emph{(with VGGT-$\Omega$~\cite{wang2026vggtomega} depth input)}}\\
\midrule
LAPA~\cite{galoaa2025lookaround} &  & 6.1 & 6.5 & 6.8 &  & 0.6 & 2.5 & 2.1 &  & 13.0 & 12.4 & 12.2 &  & 3.3 & 3.5 & 4.0 &  & 22.0 & 23.5 & 23.0 &  & 2.3 & 2.5 & 3.4 &  & 0.1 & 0.2 & 0.9 &  & 1.2 & 1.0 & 2.3 &  & 5.88\\
MVTracker~\cite{rajic2025mvtracker} &  & \underline{22.8} & \textbf{21.7} & 24.7 &  & 5.2 & 11.0 & 11.0 &  & 48.8 & 45.7 & 50.7 &  & 23.2 & \underline{20.5} & 24.2 &  & 43.7 & 35.2 & 35.4 &  & 19.1 & 15.1 & 17.8 &  & 3.4 & \underline{11.2} & 15.1 &  & \textbf{16.1} & \textbf{13.4} & 18.4 &  & 0.08\\
MVTracker (finetuned)~\cite{rajic2025mvtracker} &  & \textcolor{gray}{24.4} & \textcolor{gray}{23.3} & \textcolor{gray}{26.1} &  & \textbf{6.0} & \textbf{12.4} & 11.8 &  & \underline{50.7} & \underline{47.3} & 52.5 &  & \textbf{24.7} & \textbf{22.1} & 26.2 &  & \textbf{45.0} & 35.3 & 35.8 &  & \textbf{20.8} & \textbf{16.4} & 18.8 &  & \textbf{3.7} & \textbf{12.1} & \underline{15.9} &  & \textcolor{gray}{20.2} & \textcolor{gray}{17.3} & \textcolor{gray}{21.5} &  & 0.09\\
\midrule
\addlinespace[0.15em]
\multicolumn{35}{l}{2.a) \textbf{Monocular 4D reconstruction and point tracking}}\\
\midrule
L4P~\cite{badki2026l4p} &  & -- & -- & 2.2 &  & -- & -- & 0.4 &  & -- & -- & 6.4 &  & -- & -- & 2.5 &  & -- & -- & 3.4 &  & -- & -- & 1.1 &  & -- & -- & 0.4 &  & -- & -- & 1.1 &  & 0.22\\
St4RTrack~\cite{feng2025st4rtrack} &  & -- & -- & 4.9 &  & -- & -- & 0.4 &  & -- & -- & 16.2 &  & -- & -- & 10.2 &  & -- & -- & 6.6 &  & -- & -- & 0.3 &  & -- & -- & 0.4 &  & -- & -- & 0.4 &  & 6.92\\
Any4D~\cite{karhade2025any4d} &  & -- & -- & \textcolor{gray}{6.6} &  & -- & -- & 1.1 &  & -- & -- & 6.7 &  & -- & -- & 4.2 &  & -- & -- & 16.3 &  & -- & -- & 3.3 &  & \textcolor{gray}{--} & \textcolor{gray}{--} & \textcolor{gray}{4.6} &  & -- & -- & 9.9 &  & 6.01\\
SpatialTrackerV2~\cite{xiao2025spatialtrackerv2} &  & -- & -- & 10.8 &  & -- & -- & 1.7 &  & -- & -- & 29.3 &  & -- & -- & 12.9 &  & -- & -- & 10.9 &  & -- & -- & 4.9 &  & -- & -- & 2.4 &  & -- & -- & 13.6 &  & 2.05\\
OmniX~\cite{jiang2026omnix} (monocular) &  & -- & -- & \textcolor{gray}{13.3} &  & -- & -- & 4.9 &  & -- & -- & 13.5 &  & -- & -- & 7.5 &  & -- & -- & 25.3 &  & -- & -- & 7.1 &  & \textcolor{gray}{--} & \textcolor{gray}{--} & \textcolor{gray}{20.6} &  & -- & -- & 14.1 &  & 4.43\\
V-DPM~\cite{sucar2026vdpm} &  & -- & -- & \textcolor{gray}{13.9} &  & -- & -- & 5.4 &  & -- & -- & 20.9 &  & -- & -- & 15.5 &  & -- & -- & 29.1 &  & -- & -- & 5.6 &  & \textcolor{gray}{--} & \textcolor{gray}{--} & \textcolor{gray}{9.5} &  & -- & -- & 11.1 &  & 88.78\\
4RC~\cite{luo2026fourrc} &  & -- & -- & \textcolor{gray}{17.0} &  & -- & -- & 4.6 &  & -- & -- & 23.1 &  & -- & -- & 8.0 &  & -- & -- & 33.0 &  & -- & -- & 11.2 &  & \textcolor{gray}{--} & \textcolor{gray}{--} & \textcolor{gray}{19.7} &  & -- & -- & 19.6 &  & 2.81\\
\midrule
\addlinespace[0.15em]
\multicolumn{35}{l}{2.b) \textbf{Multi-view 4D reconstruction and point tracking}}\\
\midrule
OmniX~\cite{jiang2026omnix} &  & \textcolor{gray}{10.1} & \textcolor{gray}{9.3} & \textcolor{gray}{13.7} &  & 2.4 & 3.4 & 7.5 &  & 12.1 & 10.7 & 13.0 &  & 3.8 & 4.1 & 6.4 &  & 37.2 & 27.9 & 28.9 &  & 1.2 & 1.1 & 5.7 &  & \textcolor{gray}{6.6} & \textcolor{gray}{12.7} & \textcolor{gray}{21.0} &  & 7.4 & 5.0 & 12.9 &  & 5.01\\
\bottomrule
\end{tabular}%
}
\end{table*}

\begin{table*}[!t]
\centering
\setlength{\tabcolsep}{2.4pt}
\caption{\textbf{Camera and depth quality.} \Cref{fig:recon_scatter} summarizes this table. The point trackers in \cref{tab:main_results} (sections 1.a--1.d) receive the shared VGGT-$\Omega$ reconstruction as input. The 4D methods (sections 2.a--2.b) produce a reconstruction as part of their output, and each of their rows reports the quality of that reconstruction. K reports the intrinsics error as the mean angle, in degrees, between the pixel viewing rays implied by the estimated and ground-truth intrinsics. Subscript C metrics treat each camera's video independently, in that video's own coordinate frame and scale, so they are defined for every method: E\textsubscript{C} is relative pose AUC@30\textdegree{} over same-video pairs, excluding static cameras, whose pairs have no baseline, and D\textsubscript{C} is AUC@10\,cm, the mean fraction of ground-truth-valid pixels whose predicted 3D points lie within thresholds from 1 to 10\,cm after aligning each video separately with a Sim(3); a pixel the method leaves unpredicted counts as a miss. Methods that predict point maps are scored on those points, and the rest are unprojected from depth and intrinsics. Subscript W metrics instead require all views in one shared coordinate frame and therefore measure multi-view registration: E\textsubscript{W} uses cross-view pose pairs and D\textsubscript{W} the same AUC after a single shared Sim(3) alignment. Monocular reconstructions have no shared frame and show --. D columns are omitted for datasets without ground-truth depth. Best and second-best entries per subset and metric are bold and underlined. Reconstruction quality directly limits point tracking quality: a tracker localizes points in the 3D geometry it is given, so errors in cameras and depth propagate into every 3D track. Where both are measurable, the native reconstructions of the 4D methods are substantially less accurate than the shared VGGT-$\Omega$ reconstruction, for example E\textsubscript{C} of 13.0 versus 43.7 for SpatialTrackerV2 on Hi4D, and this accounts for much of their tracking gap in \cref{tab:main_results}. Gray cells belong to methods trained on that subset's source data (\cref{tab:training_data}) and are excluded from ranking. The second block lists the feed-forward sources of the depth-source study (\cref{tab:depth_source_results}), each evaluated with all frames of the sequence or one forward per timestep on the $(t_0, t)$ pair, and the first row is VGGT-$\Omega$ in the configuration used by the trackers of \cref{tab:main_results}: all frames, except on DROID, where the $t_0{+}t$ variant is used.}
\label{tab:reconstruction_quality}
\vspace{-0.27cm}
\resizebox{\linewidth}{!}{%
\begin{tabular}{lcccccccccccccccccccccccccccccccccc}
\toprule
 &  & \multicolumn{3}{c}{\textbf{DROID}} &  & \multicolumn{3}{c}{\textbf{EgoExo4D}} &  & \multicolumn{3}{c}{\textbf{Harmony4D}} &  & \multicolumn{5}{c}{\textbf{PACE}} &  & \multicolumn{5}{c}{\textbf{Hi4D}} &  & \multicolumn{3}{c}{\textbf{Waymo}} &  & \multicolumn{5}{c}{\textbf{Perpetua}}\\
\cmidrule(lr){3-5}\cmidrule(lr){7-9}\cmidrule(lr){11-13}\cmidrule(lr){15-19}\cmidrule(lr){21-25}\cmidrule(lr){27-29}\cmidrule(lr){31-35}
Reconstruction &  & K\,\(\downarrow\) & E\textsubscript{C}\,\(\uparrow\) & E\textsubscript{W}\,\(\uparrow\) &  & K\,\(\downarrow\) & E\textsubscript{C}\,\(\uparrow\) & E\textsubscript{W}\,\(\uparrow\) &  & K\,\(\downarrow\) & E\textsubscript{C}\,\(\uparrow\) & E\textsubscript{W}\,\(\uparrow\) &  & K\,\(\downarrow\) & E\textsubscript{C}\,\(\uparrow\) & E\textsubscript{W}\,\(\uparrow\) & D\textsubscript{C}\,\(\uparrow\) & D\textsubscript{W}\,\(\uparrow\) &  & K\,\(\downarrow\) & E\textsubscript{C}\,\(\uparrow\) & E\textsubscript{W}\,\(\uparrow\) & D\textsubscript{C}\,\(\uparrow\) & D\textsubscript{W}\,\(\uparrow\) &  & K\,\(\downarrow\) & E\textsubscript{C}\,\(\uparrow\) & E\textsubscript{W}\,\(\uparrow\) &  & K\,\(\downarrow\) & E\textsubscript{C}\,\(\uparrow\) & E\textsubscript{W}\,\(\uparrow\) & D\textsubscript{C}\,\(\uparrow\) & D\textsubscript{W}\,\(\uparrow\)\\
\midrule
\multicolumn{35}{l}{\emph{Reconstructions used in \cref{tab:main_results}}}\\
\midrule
VGGT-$\Omega$~\cite{wang2026vggtomega} (as served) &  & \textbf{1.39} & 56.0 & \textbf{85.9} &  & \textbf{0.33} & \textbf{84.8} & \textbf{98.7} &  & \textbf{0.84} & \textbf{45.7} & \textbf{87.7} &  & 0.78 & \underline{72.5} & \textbf{89.4} & 74.3 & 70.5 &  & 0.95 & \underline{92.6} & \textbf{95.6} & \textbf{45.0} & \underline{40.7} &  & 0.92 & \underline{93.1} & 86.6 &  & 0.41 & 83.7 & 86.1 & 21.8 & \underline{19.7}\\
L4P~\cite{badki2026l4p} &  & 10.44 & 10.4 & -- &  & 7.92 & 40.3 & -- &  & 9.45 & 20.5 & -- &  & 4.11 & 36.2 & -- & 43.2 & -- &  & 5.83 & 74.1 & -- & 1.2 & -- &  & 4.85 & 81.4 & -- &  & 4.57 & 60.9 & -- & 0.3 & --\\
St4RTrack~\cite{feng2025st4rtrack} &  & 9.00 & 6.6 & -- &  & 3.60 & 47.6 & -- &  & 4.29 & 15.0 & -- &  & 2.63 & 47.5 & -- & 43.5 & -- &  & 6.52 & 1.0 & -- & 0.0 & -- &  & 2.85 & 28.8 & -- &  & 3.66 & 17.9 & -- & 0.0 & --\\
Any4D~\cite{karhade2025any4d} &  & 6.98 & 13.8 & -- &  & 8.17 & 36.0 & -- &  & 8.52 & 19.5 & -- &  & 2.14 & 53.8 & -- & 55.9 & -- &  & 3.89 & 82.8 & -- & 6.2 & -- &  & \textcolor{gray}{1.31} & \textcolor{gray}{78.8} & \textcolor{gray}{--} &  & 0.43 & 65.3 & -- & 2.3 & --\\
SpatialTrackerV2~\cite{xiao2025spatialtrackerv2} &  & 5.71 & 14.0 & -- &  & 2.06 & 72.4 & -- &  & 2.99 & 19.5 & -- &  & 3.12 & 24.2 & -- & 33.6 & -- &  & 3.41 & 53.7 & -- & 3.1 & -- &  & 3.89 & 78.5 & -- &  & 0.47 & 41.6 & -- & 0.2 & --\\
V-DPM~\cite{sucar2026vdpm} &  & \underline{1.65} & \textbf{64.6} & -- &  & 3.84 & 70.7 & -- &  & 2.81 & 31.0 & -- &  & 0.89 & 71.7 & -- & 74.2 & -- &  & 3.75 & 82.1 & -- & 0.9 & -- &  & \textcolor{gray}{0.91} & \textcolor{gray}{87.7} & \textcolor{gray}{--} &  & 1.08 & 69.5 & -- & 2.9 & --\\
4RC~\cite{luo2026fourrc} &  & 2.68 & 51.1 & -- &  & 3.42 & \underline{81.5} & -- &  & 4.73 & 32.0 & -- &  & 0.77 & \textbf{75.0} & -- & 71.5 & -- &  & 2.24 & \textbf{96.1} & -- & 23.0 & -- &  & \textcolor{gray}{0.60} & \textcolor{gray}{92.5} & \textcolor{gray}{--} &  & 0.46 & \textbf{92.8} & -- & 16.5 & --\\
OmniX~\cite{jiang2026omnix} &  & 1.85 & 51.2 & 54.7 &  & 2.72 & 61.1 & 88.3 &  & 2.90 & 42.1 & 77.3 &  & 0.85 & 65.9 & 86.6 & \textbf{77.5} & \textbf{73.6} &  & 1.50 & 62.5 & 21.0 & 24.8 & 0.5 &  & \textcolor{gray}{0.75} & \textcolor{gray}{94.4} & \textcolor{gray}{88.4} &  & 1.50 & \textbf{92.8} & \underline{86.9} & \textbf{28.0} & 18.4\\
\midrule
\multicolumn{35}{l}{\emph{Feed-forward sources of \cref{tab:depth_source_results}}}\\
\midrule
VGGT-$\Omega$~\cite{wang2026vggtomega} (all frames) &  & 1.71 & \underline{63.5} & \underline{83.4} &  & \textbf{0.33} & \textbf{84.8} & \textbf{98.7} &  & \textbf{0.84} & \textbf{45.7} & \textbf{87.7} &  & 0.78 & \underline{72.5} & \textbf{89.4} & 74.3 & 70.5 &  & 0.95 & \underline{92.6} & \textbf{95.6} & \textbf{45.0} & \underline{40.7} &  & 0.92 & \underline{93.1} & 86.6 &  & 0.41 & 83.7 & 86.1 & 21.8 & \underline{19.7}\\
VGGT-$\Omega$~\cite{wang2026vggtomega} ($t_0{+}t$) &  & \textbf{1.39} & 56.0 & \textbf{85.9} &  & \underline{0.41} & 76.1 & \underline{98.1} &  & \underline{1.01} & 38.9 & 83.6 &  & 0.79 & 66.8 & 87.3 & 73.9 & 69.8 &  & 0.97 & 88.0 & \underline{95.0} & 43.6 & 38.9 &  & 1.03 & 85.7 & 81.1 &  & 0.63 & 70.5 & 76.0 & 11.7 & 9.7\\
VGGT~\cite{wang2026vggt} (all frames) &  & 5.02 & 42.5 & 45.5 &  & 2.37 & 49.0 & 81.9 &  & 4.55 & 17.7 & 45.5 &  & 3.32 & 60.5 & 82.0 & 42.0 & 39.8 &  & 2.88 & 73.1 & 77.7 & 3.9 & 1.7 &  & 2.83 & 88.8 & 78.4 &  & 0.56 & 56.8 & 56.3 & 4.0 & 3.6\\
VGGT~\cite{wang2026vggt} ($t_0{+}t$) &  & 4.71 & 34.9 & 53.5 &  & 1.89 & 44.7 & 84.5 &  & 4.11 & 18.6 & 51.5 &  & 3.37 & 54.4 & 79.0 & 41.9 & 39.2 &  & 2.65 & 67.0 & 73.1 & 3.3 & 1.1 &  & 2.84 & 85.2 & 74.7 &  & 0.58 & 49.6 & 51.8 & 2.5 & 1.8\\
Pi3~\cite{wang2025pi3} (all frames) &  & 2.50 & 38.4 & 73.6 &  & 0.94 & 80.7 & 96.9 &  & 1.50 & \underline{42.2} & \underline{86.0} &  & \underline{0.75} & 71.1 & \underline{89.0} & \underline{76.6} & \textbf{73.6} &  & \underline{0.94} & 89.9 & 94.2 & 36.7 & 32.9 &  & 0.88 & 92.1 & 80.5 &  & 0.74 & 82.7 & 83.8 & 14.8 & 13.5\\
Pi3~\cite{wang2025pi3} ($t_0{+}t$) &  & 2.46 & 29.1 & 69.7 &  & 1.14 & 65.8 & 91.5 &  & 1.70 & 36.8 & 80.6 &  & 0.81 & 64.1 & 86.9 & 75.7 & \underline{73.1} &  & \underline{0.94} & 82.2 & 91.5 & 32.1 & 29.1 &  & 1.44 & 82.6 & 73.5 &  & 0.95 & 65.8 & 71.8 & 7.8 & 6.4\\
Pi3X~\cite{wang2025pi3} (all frames) &  & 2.39 & 38.2 & 69.9 &  & 0.66 & 80.7 & 97.7 &  & 1.76 & 38.6 & 81.2 &  & \textbf{0.74} & 71.8 & \textbf{89.4} & 76.4 & 73.0 &  & \textbf{0.90} & 91.5 & 94.2 & \underline{44.2} & \textbf{41.1} &  & \underline{0.71} & \textbf{94.8} & \textbf{90.2} &  & \textbf{0.28} & \underline{86.4} & \textbf{87.9} & \underline{23.2} & \textbf{21.2}\\
Pi3X~\cite{wang2025pi3} ($t_0{+}t$) &  & 2.26 & 31.4 & 67.5 &  & 0.61 & 69.0 & 96.0 &  & 2.19 & 33.8 & 78.0 &  & 0.78 & 65.7 & 87.4 & 75.5 & 72.1 &  & \textbf{0.90} & 84.1 & 92.4 & 38.2 & 37.2 &  & \textbf{0.61} & 89.2 & \underline{86.8} &  & \underline{0.37} & 70.4 & 76.1 & 12.4 & 10.9\\
DUSt3R~\cite{wang2024dust3r} (all frames) &  & 4.45 & 4.2 & 24.4 &  & 2.08 & 27.3 & 81.2 &  & 3.48 & 18.5 & 48.4 &  & 2.03 & 33.4 & 63.9 & 56.6 & 55.2 &  & 1.40 & 71.1 & 83.9 & 11.9 & 8.9 &  & 6.49 & 56.7 & 31.6 &  & 2.69 & 5.7 & 10.4 & 0.0 & 0.0\\
\bottomrule
\end{tabular}%
}
\end{table*}

\begin{table*}[!t]
\centering
\setlength{\tabcolsep}{2.2pt}
\caption{\textbf{Pixel-space and visibility metrics.} \(\mathrm{AJ}^{2D}\), \(\delta^{2D}\), and \(\mathrm{OA}\) report the TAP-Vid pixel-space average Jaccard, location accuracy, and occlusion-classification accuracy, each in the query view (Q) and in the non-query views (N). The N columns require cross-view tracking and are undefined (--) for monocular methods. A method that predicts one view-independent visibility is scored on it in every view; Jaccard and occlusion metrics are undefined for methods without a visibility head. Best and second-best entries are bold and underlined. Gray cells belong to methods trained on that subset's source data (\cref{tab:training_data}) and are excluded from ranking.}
\label{tab:recognition_metrics}
\vspace{-0.27cm}
\resizebox{\linewidth}{!}{%
\begin{tabular}{lccccccccccccccccccccccccccccccccccccccccccccccccc}
\toprule
 &  & \multicolumn{6}{c}{\textbf{DROID}} &  & \multicolumn{6}{c}{\textbf{EgoExo4D}} &  & \multicolumn{6}{c}{\textbf{Harmony4D}} &  & \multicolumn{6}{c}{\textbf{PACE}} &  & \multicolumn{6}{c}{\textbf{Hi4D}} &  & \multicolumn{6}{c}{\textbf{Waymo}} &  & \multicolumn{6}{c}{\textbf{Perpetua}}\\
\cmidrule(lr){3-8}\cmidrule(lr){10-15}\cmidrule(lr){17-22}\cmidrule(lr){24-29}\cmidrule(lr){31-36}\cmidrule(lr){38-43}\cmidrule(lr){45-50}
Method &  & \(\mathrm{AJ}^{2D}_{\mathrm{Q}}\) & \(\mathrm{AJ}^{2D}_{\mathrm{N}}\) & \(\delta^{2D}_{\mathrm{Q}}\) & \(\delta^{2D}_{\mathrm{N}}\) & \(\mathrm{OA}_{\mathrm{Q}}\) & \(\mathrm{OA}_{\mathrm{N}}\) &  & \(\mathrm{AJ}^{2D}_{\mathrm{Q}}\) & \(\mathrm{AJ}^{2D}_{\mathrm{N}}\) & \(\delta^{2D}_{\mathrm{Q}}\) & \(\delta^{2D}_{\mathrm{N}}\) & \(\mathrm{OA}_{\mathrm{Q}}\) & \(\mathrm{OA}_{\mathrm{N}}\) &  & \(\mathrm{AJ}^{2D}_{\mathrm{Q}}\) & \(\mathrm{AJ}^{2D}_{\mathrm{N}}\) & \(\delta^{2D}_{\mathrm{Q}}\) & \(\delta^{2D}_{\mathrm{N}}\) & \(\mathrm{OA}_{\mathrm{Q}}\) & \(\mathrm{OA}_{\mathrm{N}}\) &  & \(\mathrm{AJ}^{2D}_{\mathrm{Q}}\) & \(\mathrm{AJ}^{2D}_{\mathrm{N}}\) & \(\delta^{2D}_{\mathrm{Q}}\) & \(\delta^{2D}_{\mathrm{N}}\) & \(\mathrm{OA}_{\mathrm{Q}}\) & \(\mathrm{OA}_{\mathrm{N}}\) &  & \(\mathrm{AJ}^{2D}_{\mathrm{Q}}\) & \(\mathrm{AJ}^{2D}_{\mathrm{N}}\) & \(\delta^{2D}_{\mathrm{Q}}\) & \(\delta^{2D}_{\mathrm{N}}\) & \(\mathrm{OA}_{\mathrm{Q}}\) & \(\mathrm{OA}_{\mathrm{N}}\) &  & \(\mathrm{AJ}^{2D}_{\mathrm{Q}}\) & \(\mathrm{AJ}^{2D}_{\mathrm{N}}\) & \(\delta^{2D}_{\mathrm{Q}}\) & \(\delta^{2D}_{\mathrm{N}}\) & \(\mathrm{OA}_{\mathrm{Q}}\) & \(\mathrm{OA}_{\mathrm{N}}\) &  & \(\mathrm{AJ}^{2D}_{\mathrm{Q}}\) & \(\mathrm{AJ}^{2D}_{\mathrm{N}}\) & \(\delta^{2D}_{\mathrm{Q}}\) & \(\delta^{2D}_{\mathrm{N}}\) & \(\mathrm{OA}_{\mathrm{Q}}\) & \(\mathrm{OA}_{\mathrm{N}}\)\\
\midrule
\addlinespace[0.15em]
\multicolumn{50}{l}{1.a) \textbf{Monocular 2D point trackers} \emph{(lifted with VGGT-$\Omega$~\cite{wang2026vggtomega})}}\\
\midrule
PIPs~\cite{harley2022particle} &  & 41.0 & 10.8 & 60.0 & 26.9 & 80.9 & 43.7 &  & 48.8 & -- & 65.2 & 45.7 & 85.6 & -- &  & 29.5 & 8.6 & 50.6 & 26.2 & 67.4 & 37.5 &  & 53.6 & 46.4 & 64.1 & 57.3 & 96.7 & 94.2 &  & 35.5 & 6.2 & 60.9 & 28.2 & 70.3 & 33.0 &  & 31.6 & -- & 64.4 & 18.7 & 71.3 & -- &  & 12.4 & 1.1 & 39.9 & 8.0 & 50.0 & 31.8\\
PIPs++~\cite{zheng2023pointodyssey} &  & -- & -- & 68.3 & 28.9 & -- & -- &  & -- & -- & 79.4 & 52.1 & -- & -- &  & -- & -- & 61.9 & 29.2 & -- & -- &  & -- & -- & 68.0 & 59.6 & -- & -- &  & -- & -- & 63.0 & 27.7 & -- & -- &  & -- & -- & 58.2 & 21.0 & -- & -- &  & -- & -- & 43.1 & 8.7 & -- & --\\
TAPTRv3~\cite{li2024taptrv3} &  & 62.6 & 12.7 & 76.3 & 30.2 & 89.5 & 45.0 &  & 71.3 & -- & 83.5 & 53.4 & 88.2 & -- &  & 43.8 & 12.2 & 59.4 & 28.9 & 77.6 & 51.0 &  & 65.6 & 54.0 & 75.2 & 64.3 & 95.0 & 91.7 &  & 51.6 & 8.2 & 71.7 & 31.0 & 83.4 & 46.5 &  & 52.7 & -- & 67.3 & 20.3 & 94.8 & -- &  & 30.9 & 2.4 & 45.8 & 8.4 & 83.6 & 71.2\\
CoTracker2~\cite{karaev2024cotracker} &  & 61.2 & 13.8 & 75.7 & 31.2 & 84.6 & 51.1 &  & 67.7 & -- & 84.4 & 54.5 & 81.7 & -- &  & 41.2 & 11.4 & 55.4 & 27.0 & 72.8 & 53.8 &  & 66.7 & 54.6 & 75.4 & 64.7 & 94.1 & 90.9 &  & 50.7 & 8.0 & 73.5 & 31.6 & 78.6 & 50.0 &  & 57.4 & -- & 72.4 & 34.3 & 95.4 & -- &  & 28.9 & 2.3 & 50.0 & 10.1 & 81.8 & 60.3\\
TAPIR~\cite{doersch2023tapir} &  & 55.5 & 12.0 & 67.7 & 28.6 & 83.7 & 44.1 &  & 65.6 & -- & 78.3 & 51.5 & 88.4 & -- &  & 49.6 & 12.4 & 68.5 & 31.0 & 82.2 & 46.3 &  & 62.8 & 52.4 & 72.7 & 63.1 & 92.9 & 89.7 &  & 56.9 & 8.5 & 75.2 & 31.4 & 86.5 & 47.6 &  & 51.9 & -- & 66.4 & 18.9 & 92.1 & -- &  & 31.0 & 2.5 & 47.8 & 9.5 & 82.5 & 69.7\\
CoTracker1~\cite{karaev2024cotracker} &  & 62.3 & 14.0 & 76.6 & 31.2 & 86.2 & \underline{51.5} &  & 67.5 & -- & 83.9 & 54.6 & 82.0 & -- &  & 40.6 & 10.7 & 57.8 & 28.9 & 70.6 & 53.1 &  & 67.6 & 55.1 & 76.5 & 65.3 & 92.6 & 89.3 &  & 52.3 & 8.2 & 74.4 & 32.6 & 80.0 & 49.9 &  & 54.0 & -- & 71.1 & 34.1 & 94.1 & -- &  & 27.9 & 2.2 & 50.5 & 10.3 & 81.2 & 58.7\\
BootsTAPIR~\cite{doersch2024bootstap} &  & 63.5 & 12.8 & 75.1 & 30.1 & 88.1 & 44.8 &  & 71.2 & -- & 83.5 & 54.0 & 90.1 & -- &  & 53.6 & 12.9 & 72.0 & 31.9 & \underline{83.7} & 48.0 &  & 67.5 & 55.5 & 76.3 & 65.5 & 94.8 & 91.5 &  & 62.2 & 8.7 & 79.3 & 32.1 & 88.1 & 49.0 &  & 60.1 & -- & 73.7 & 20.4 & 94.9 & -- &  & 34.4 & 2.5 & 55.0 & 10.5 & 82.5 & 72.0\\
DOT~\cite{lemoing2024dense} &  & 69.7 & 14.1 & 81.5 & 32.6 & 91.2 & 47.6 &  & 75.5 & -- & 87.1 & 55.9 & 90.0 & -- &  & 46.9 & 12.5 & 59.5 & 28.5 & 77.8 & 52.6 &  & 69.3 & 55.7 & 78.4 & 66.4 & 93.6 & 89.9 &  & 63.6 & 8.7 & 81.3 & 33.4 & 89.0 & 48.2 &  & 53.3 & -- & 72.7 & 34.2 & 91.8 & -- &  & 37.6 & 2.8 & 56.8 & 11.2 & 85.8 & 67.0\\
LocoTrack~\cite{cho2024locotrack} &  & 63.6 & 12.2 & 76.2 & 30.3 & 83.6 & 44.5 &  & 75.1 & -- & 86.1 & 54.2 & 86.0 & -- &  & 54.4 & 13.1 & 73.0 & 32.0 & 81.0 & 49.7 &  & 69.8 & 56.3 & 79.1 & 66.9 & 92.1 & 88.9 &  & 65.5 & 8.8 & 81.3 & 32.2 & 87.4 & 49.9 &  & 58.0 & -- & 71.6 & 19.8 & 92.7 & -- &  & 35.3 & 2.6 & 53.6 & 10.0 & 81.9 & \underline{76.8}\\
CoTracker3~\cite{karaev2024cotracker3} &  & 67.6 & 13.5 & 79.9 & 32.0 & 90.5 & 44.6 &  & 74.8 & -- & 86.6 & 56.0 & 90.2 & -- &  & \underline{54.9} & 13.3 & 73.0 & 32.7 & 83.2 & 47.9 &  & 68.8 & 55.7 & 77.4 & 65.9 & 95.7 & 92.3 &  & 61.0 & 9.1 & 78.3 & 33.5 & 87.3 & 48.6 &  & 57.9 & -- & 72.1 & 27.0 & 95.2 & -- &  & 39.0 & 2.8 & 56.7 & 11.2 & 85.5 & 68.5\\
MegaFlow~\cite{zhang2026megaflow} &  & -- & -- & 82.3 & 32.4 & -- & -- &  & -- & -- & 89.4 & 56.8 & -- & -- &  & -- & -- & 73.7 & 32.5 & -- & -- &  & -- & -- & 80.4 & 67.9 & -- & -- &  & -- & -- & 83.0 & 33.4 & -- & -- &  & -- & -- & 72.7 & 28.9 & -- & -- &  & -- & -- & 64.3 & 12.1 & -- & --\\
TAPNext++~\cite{jung2026tapnextpp} &  & \underline{72.0} & 14.0 & \underline{83.5} & 32.2 & \textbf{93.5} & 46.1 &  & 77.8 & -- & 89.6 & 56.5 & \textbf{92.2} & -- &  & 51.2 & 13.3 & 69.0 & 32.2 & 83.1 & 48.8 &  & 70.0 & 56.4 & 78.8 & 66.6 & 94.5 & 90.8 &  & \textbf{69.8} & 9.2 & \underline{85.6} & 35.0 & \textbf{92.0} & 48.0 &  & \textbf{64.7} & -- & \textbf{77.3} & 20.5 & \textbf{97.1} & -- &  & \underline{49.3} & 3.4 & 68.3 & 12.9 & \textbf{90.4} & 67.1\\
Track-On-R~\cite{aydemir2025trackon} &  & 71.7 & 14.1 & 82.9 & 32.6 & \underline{93.2} & 46.2 &  & \underline{78.3} & -- & 89.9 & 56.7 & 91.6 & -- &  & 53.9 & 13.8 & 73.4 & 33.1 & 82.9 & 48.9 &  & \underline{72.2} & \underline{57.8} & \underline{80.7} & 68.1 & 95.1 & 91.2 &  & 65.4 & 8.9 & 82.4 & 33.0 & 89.7 & 48.4 &  & \underline{63.1} & -- & 76.6 & 21.0 & 96.3 & -- &  & 42.0 & 3.0 & 61.5 & 11.7 & 87.7 & 68.7\\
AllTracker~\cite{harley2025alltracker} &  & 70.9 & 14.0 & 82.7 & 32.6 & 92.0 & 44.5 &  & \textbf{79.2} & -- & \textbf{90.5} & 57.7 & \underline{92.1} & -- &  & \textbf{60.4} & \underline{14.1} & \textbf{77.4} & 34.0 & \textbf{86.0} & 47.8 &  & 71.4 & 57.6 & 79.8 & 67.6 & 95.7 & 92.1 &  & 67.3 & 9.3 & 83.4 & 34.9 & \underline{90.1} & 48.2 &  & \underline{63.1} & -- & 75.6 & 34.3 & 96.4 & -- &  & \textbf{49.4} & 3.3 & \underline{69.1} & 13.7 & \underline{88.2} & 66.5\\
CoWTracker~\cite{lai2026cowtracker} &  & \textbf{73.5} & 14.1 & \textbf{85.2} & 32.9 & \underline{93.2} & 44.6 &  & 78.0 & -- & \underline{90.3} & 57.8 & \underline{92.1} & -- &  & 52.8 & 12.7 & \underline{74.8} & 33.3 & 81.6 & 43.4 &  & \textbf{74.9} & \textbf{59.7} & \textbf{82.5} & \textbf{69.6} & \textbf{96.9} & 93.4 &  & \underline{67.8} & 9.1 & \textbf{87.3} & 35.1 & 88.7 & 44.8 &  & 60.8 & -- & \underline{77.2} & 35.0 & 94.6 & -- &  & 43.9 & 3.0 & \textbf{70.6} & 13.8 & 85.4 & 58.6\\
\midrule
\addlinespace[0.15em]
\multicolumn{50}{l}{1.b) \textbf{Monocular 3D point trackers} \emph{(with VGGT-$\Omega$~\cite{wang2026vggtomega} depth input)}}\\
\midrule
SpatialTracker~\cite{xiao2024spatialtracker} &  & 57.4 & 6.8 & 72.0 & 16.6 & 86.1 & 44.3 &  & 72.0 & -- & 84.6 & 41.6 & 87.0 & -- &  & 38.0 & 5.8 & 54.7 & 17.8 & 74.1 & 49.1 &  & 62.7 & 31.0 & 73.6 & 41.3 & 93.0 & 89.9 &  & 47.9 & 3.6 & 69.4 & 15.2 & 80.3 & 44.9 &  & 54.1 & -- & 71.8 & 31.7 & 93.2 & -- &  & 29.3 & 1.6 & 52.3 & 8.2 & 81.4 & 60.4\\
SceneTracker~\cite{wang2024scenetracker} &  & 40.7 & 8.7 & 61.6 & 24.3 & 75.8 & 42.3 &  & 60.8 & -- & 77.6 & 49.2 & 83.6 & -- &  & 30.6 & 7.5 & 53.4 & 25.6 & 59.8 & 30.4 &  & 54.8 & 31.4 & 66.0 & 42.9 & \underline{96.8} & \textbf{95.5} &  & 32.0 & 4.1 & 57.7 & 21.7 & 59.8 & 21.3 &  & 21.3 & -- & 56.0 & 22.7 & 37.5 & -- &  & 11.2 & 1.1 & 44.4 & 9.6 & 29.2 & 11.7\\
DELTA~\cite{ngo2025delta} &  & 59.2 & 11.5 & 71.9 & 26.9 & 83.0 & 47.9 &  & 74.3 & -- & 85.6 & 54.8 & 86.0 & -- &  & 45.9 & 11.7 & 57.2 & 26.8 & 76.1 & 53.5 &  & 70.7 & 52.6 & 78.8 & 62.5 & 94.6 & 90.9 &  & 56.9 & 7.5 & 74.8 & 29.2 & 84.2 & 49.0 &  & 61.1 & -- & 75.1 & 37.3 & 96.1 & -- &  & 31.6 & 2.3 & 51.2 & 10.0 & 83.3 & 64.1\\
SpatialTrackerV2~\cite{xiao2025spatialtrackerv2} &  & 42.2 & 10.7 & 63.9 & 31.5 & 70.5 & 49.9 &  & 55.6 & -- & 79.3 & 54.3 & 72.9 & -- &  & 37.8 & 10.0 & 59.4 & 30.9 & 71.8 & \underline{54.9} &  & 50.5 & 41.7 & 70.3 & 60.8 & 73.0 & 70.8 &  & 41.8 & 6.5 & 68.2 & 32.9 & 73.3 & \underline{57.8} &  & 36.1 & -- & 64.4 & 27.8 & 83.2 & -- &  & 23.1 & 2.2 & 46.8 & 11.7 & 79.6 & 75.2\\
TAPIP3D~\cite{zhang2025tapip3d} &  & 54.2 & 13.5 & 68.9 & 34.3 & 82.4 & 47.0 &  & 77.3 & -- & \underline{90.3} & \textbf{63.2} & 89.7 & -- &  & 48.3 & 13.8 & 69.3 & 37.1 & 79.9 & 49.8 &  & 71.1 & 57.5 & 80.0 & \underline{68.5} & 95.6 & 92.0 &  & 60.4 & 9.6 & 79.9 & \underline{43.9} & 85.9 & 50.7 &  & 61.6 & -- & 76.3 & 44.7 & 95.1 & -- &  & 37.5 & 3.2 & 62.1 & \underline{21.3} & 85.5 & 65.5\\
\midrule
\addlinespace[0.15em]
\multicolumn{50}{l}{1.c) \textbf{Multi-view 2D point trackers} \emph{(lifted with VGGT-$\Omega$~\cite{wang2026vggtomega})}}\\
\midrule
MV-TAP~\cite{koo2025mvtap} &  & 53.4 & \textbf{16.3} & 67.9 & \underline{35.3} & 81.7 & \underline{51.5} &  & 66.7 & -- & 82.5 & 60.5 & 82.3 & -- &  & 44.8 & \textbf{16.1} & 60.5 & 34.3 & 74.9 & 51.2 &  & 66.2 & 56.9 & 75.5 & 66.4 & 95.1 & 93.9 &  & 44.8 & \textbf{10.5} & 67.6 & 39.8 & 74.6 & 42.4 &  & 51.1 & -- & 62.1 & 21.2 & 90.8 & -- &  & 32.2 & \textbf{3.6} & 41.2 & 9.5 & 81.2 & 73.1\\
\textcolor{gray}{MV-TAP (MV)$^\ddagger$~\cite{koo2025mvtap}} &  & \textcolor{gray}{47.7} & \textcolor{gray}{29.2} & \textcolor{gray}{62.4} & \textcolor{gray}{58.2} & \textcolor{gray}{80.1} & \textcolor{gray}{53.1} &  & \textcolor{gray}{64.9} & \textcolor{gray}{--} & \textcolor{gray}{81.4} & \textcolor{gray}{79.2} & \textcolor{gray}{82.5} & \textcolor{gray}{--} &  & \textcolor{gray}{42.7} & \textcolor{gray}{22.8} & \textcolor{gray}{60.4} & \textcolor{gray}{49.9} & \textcolor{gray}{74.2} & \textcolor{gray}{51.5} &  & \textcolor{gray}{65.0} & \textcolor{gray}{64.5} & \textcolor{gray}{74.3} & \textcolor{gray}{74.5} & \textcolor{gray}{95.4} & \textcolor{gray}{94.1} &  & \textcolor{gray}{45.1} & \textcolor{gray}{14.9} & \textcolor{gray}{69.1} & \textcolor{gray}{56.5} & \textcolor{gray}{74.3} & \textcolor{gray}{42.1} &  & \textcolor{gray}{50.6} & \textcolor{gray}{--} & \textcolor{gray}{61.9} & \textcolor{gray}{24.3} & \textcolor{gray}{90.5} & \textcolor{gray}{--} &  & \textcolor{gray}{32.4} & \textcolor{gray}{5.5} & \textcolor{gray}{41.4} & \textcolor{gray}{14.3} & \textcolor{gray}{81.4} & \textcolor{gray}{73.5}\\
\textcolor{gray}{MV-TAP (MT)$^\dagger$~\cite{koo2025mvtap}} &  & \textcolor{gray}{48.3} & \textcolor{gray}{45.3} & \textcolor{gray}{59.5} & \textcolor{gray}{61.0} & \textcolor{gray}{83.1} & \textcolor{gray}{92.6} &  & \textcolor{gray}{66.6} & \textcolor{gray}{64.6} & \textcolor{gray}{81.9} & \textcolor{gray}{81.4} & \textcolor{gray}{85.2} & \textcolor{gray}{85.2} &  & \textcolor{gray}{44.2} & \textcolor{gray}{34.5} & \textcolor{gray}{57.9} & \textcolor{gray}{53.2} & \textcolor{gray}{77.6} & \textcolor{gray}{85.2} &  & \textcolor{gray}{64.7} & \textcolor{gray}{64.4} & \textcolor{gray}{74.4} & \textcolor{gray}{74.7} & \textcolor{gray}{93.8} & \textcolor{gray}{92.8} &  & \textcolor{gray}{48.9} & \textcolor{gray}{37.7} & \textcolor{gray}{67.2} & \textcolor{gray}{61.8} & \textcolor{gray}{81.9} & \textcolor{gray}{88.7} &  & \textcolor{gray}{51.5} & \textcolor{gray}{50.0} & \textcolor{gray}{61.3} & \textcolor{gray}{59.3} & \textcolor{gray}{91.8} & \textcolor{gray}{97.3} &  & \textcolor{gray}{32.0} & \textcolor{gray}{25.4} & \textcolor{gray}{39.7} & \textcolor{gray}{34.1} & \textcolor{gray}{82.8} & \textcolor{gray}{91.7}\\
\midrule
\addlinespace[0.15em]
\multicolumn{50}{l}{1.d) \textbf{Multi-view 3D point trackers} \emph{(with VGGT-$\Omega$~\cite{wang2026vggtomega} depth input)}}\\
\midrule
LAPA~\cite{galoaa2025lookaround} &  & 4.8 & 4.7 & 9.0 & 10.5 & 55.9 & \textbf{75.5} &  & 9.6 & \underline{9.4} & 16.6 & 16.6 & 52.6 & \underline{56.2} &  & 5.2 & 3.6 & 9.7 & 7.7 & 55.8 & \textbf{72.8} &  & 33.5 & 32.2 & 39.6 & 38.6 & 93.4 & 93.1 &  & 4.0 & 1.9 & 8.1 & 5.4 & 58.6 & \textbf{72.6} &  & 1.1 & \underline{0.9} & 5.1 & 3.6 & 70.3 & \textbf{86.9} &  & 2.7 & 1.0 & 6.1 & 2.5 & 75.6 & \textbf{84.6}\\
MVTracker~\cite{rajic2025mvtracker} &  & 28.1 & 12.9 & 43.1 & 32.1 & 76.2 & 42.8 &  & 57.9 & -- & 76.3 & 60.6 & 83.7 & -- &  & 27.3 & 12.4 & 51.2 & \underline{37.7} & 60.3 & 30.6 &  & 60.4 & 54.1 & 70.9 & 64.5 & 96.7 & \underline{95.4} &  & 32.9 & 8.8 & 61.6 & 42.1 & 63.5 & 25.8 &  & 29.4 & -- & 63.9 & \underline{46.8} & 61.3 & -- &  & 11.8 & \underline{3.5} & 42.4 & \textbf{27.6} & 37.3 & 20.2\\
MVTracker (finetuned)~\cite{rajic2025mvtracker} &  & 31.9 & \underline{14.8} & 47.4 & \textbf{35.8} & 77.3 & 43.9 &  & 62.3 & \textbf{42.4} & 80.8 & \underline{63.1} & 83.8 & \textbf{75.1} &  & 30.9 & 11.3 & 56.8 & \textbf{40.8} & 61.7 & 30.8 &  & 63.4 & 56.0 & 73.6 & 66.4 & 96.7 & \underline{95.4} &  & 36.5 & \underline{9.9} & 66.1 & \textbf{46.2} & 65.4 & 28.0 &  & 32.9 & \textbf{10.8} & 68.7 & \textbf{50.7} & 64.6 & \underline{41.5} &  & \textcolor{gray}{15.6} & \textcolor{gray}{5.0} & \textcolor{gray}{50.6} & \textcolor{gray}{36.0} & \textcolor{gray}{45.2} & \textcolor{gray}{28.2}\\
\midrule
\addlinespace[0.15em]
\multicolumn{50}{l}{2.a) \textbf{Monocular 4D reconstruction and point tracking}}\\
\midrule
L4P~\cite{badki2026l4p} &  & 59.4 & -- & 73.3 & -- & 92.1 & -- &  & 64.4 & -- & 78.1 & -- & 91.8 & -- &  & 34.6 & -- & 51.7 & -- & 78.8 & -- &  & 59.7 & -- & 69.5 & -- & 95.9 & -- &  & 54.9 & -- & 74.4 & -- & 87.8 & -- &  & 56.0 & -- & 69.5 & -- & \underline{96.9} & -- &  & 27.2 & -- & 52.3 & -- & 79.8 & --\\
St4RTrack~\cite{feng2025st4rtrack} &  & -- & -- & 44.1 & -- & -- & -- &  & -- & -- & 59.0 & -- & -- & -- &  & -- & -- & 41.9 & -- & -- & -- &  & -- & -- & 41.7 & -- & -- & -- &  & -- & -- & 23.6 & -- & -- & -- &  & -- & -- & 14.1 & -- & -- & -- &  & -- & -- & 12.6 & -- & -- & --\\
Any4D~\cite{karhade2025any4d} &  & -- & -- & 39.2 & -- & -- & -- &  & -- & -- & 53.1 & -- & -- & -- &  & -- & -- & 37.8 & -- & -- & -- &  & -- & -- & 52.4 & -- & -- & -- &  & -- & -- & 47.7 & -- & -- & -- &  & \textcolor{gray}{--} & \textcolor{gray}{--} & \textcolor{gray}{31.7} & \textcolor{gray}{--} & \textcolor{gray}{--} & \textcolor{gray}{--} &  & -- & -- & 28.7 & -- & -- & --\\
SpatialTrackerV2~\cite{xiao2025spatialtrackerv2} &  & 65.4 & -- & 78.3 & -- & 88.9 & -- &  & 75.0 & -- & 87.3 & -- & 91.7 & -- &  & 41.9 & -- & 55.4 & -- & 70.9 & -- &  & 67.1 & -- & 76.1 & -- & 93.3 & -- &  & 54.8 & -- & 73.5 & -- & 86.0 & -- &  & 53.9 & -- & 73.5 & -- & 92.6 & -- &  & 27.1 & -- & 51.9 & -- & 77.5 & --\\
OmniX~\cite{jiang2026omnix} (monocular) &  & -- & -- & 41.5 & -- & -- & -- &  & -- & -- & 17.4 & -- & -- & -- &  & -- & -- & 12.5 & -- & -- & -- &  & -- & -- & 63.0 & -- & -- & -- &  & -- & -- & 31.5 & -- & -- & -- &  & \textcolor{gray}{--} & \textcolor{gray}{--} & \textcolor{gray}{59.0} & \textcolor{gray}{--} & \textcolor{gray}{--} & \textcolor{gray}{--} &  & -- & -- & 15.3 & -- & -- & --\\
V-DPM~\cite{sucar2026vdpm} &  & -- & -- & 48.1 & -- & -- & -- &  & -- & -- & 63.4 & -- & -- & -- &  & -- & -- & 42.9 & -- & -- & -- &  & -- & -- & 67.1 & -- & -- & -- &  & -- & -- & 19.7 & -- & -- & -- &  & \textcolor{gray}{--} & \textcolor{gray}{--} & \textcolor{gray}{29.7} & \textcolor{gray}{--} & \textcolor{gray}{--} & \textcolor{gray}{--} &  & -- & -- & 31.8 & -- & -- & --\\
4RC~\cite{luo2026fourrc} &  & -- & -- & 52.1 & -- & -- & -- &  & -- & -- & 77.6 & -- & -- & -- &  & -- & -- & 47.0 & -- & -- & -- &  & -- & -- & 72.1 & -- & -- & -- &  & -- & -- & 69.0 & -- & -- & -- &  & \textcolor{gray}{--} & \textcolor{gray}{--} & \textcolor{gray}{67.2} & \textcolor{gray}{--} & \textcolor{gray}{--} & \textcolor{gray}{--} &  & -- & -- & 42.5 & -- & -- & --\\
\midrule
\addlinespace[0.15em]
\multicolumn{50}{l}{2.b) \textbf{Multi-view 4D reconstruction and point tracking}}\\
\midrule
OmniX~\cite{jiang2026omnix} &  & -- & -- & 43.4 & 11.5 & -- & -- &  & -- & -- & 14.1 & 12.2 & -- & -- &  & -- & -- & 9.8 & 6.2 & -- & -- &  & -- & -- & 64.9 & 59.1 & -- & -- &  & -- & -- & 29.9 & 4.4 & -- & -- &  & \textcolor{gray}{--} & \textcolor{gray}{--} & \textcolor{gray}{59.7} & \textcolor{gray}{37.3} & \textcolor{gray}{--} & \textcolor{gray}{--} &  & -- & -- & 13.9 & 5.8 & -- & --\\
\bottomrule
\end{tabular}%
}
\end{table*}

\begin{table*}[!t]
\centering
\setlength{\tabcolsep}{2.1pt}
\caption{\textbf{Tracking results with dense ground-truth depth input.} This appendix table mirrors \cref{tab:main_results}, but only for splits where dense dataset depth is available. Replacing estimated geometry with dataset depth substantially improves the same trackers, showing that inaccurate depth, scale, and camera geometry are a major source of error in reconstruction-based tracking pipelines. Methods may use dataset depth either as direct model input or for lifting 2D tracks to 3D. Each dataset block reports \dmetric{W}, \dmetric{N}, and \dmetric{Q}. All 3D tracking metrics use query-median scaling. Best and second-best entries are bold and underlined. Gray MV-TAP rows use privileged oracle queries (\cref{tab:main_results}) and are excluded from ranking.}
\label{tab:gt_depth_results}
\vspace{-0.27cm}
\small
\begin{tabular}{lccccccccccccccccc}
\toprule
 &  &  & \multicolumn{3}{c}{\textbf{DROID}} &  & \multicolumn{3}{c}{\textbf{PACE}} &  & \multicolumn{3}{c}{\textbf{Hi4D}} &  & \multicolumn{3}{c}{\textbf{Perpetua}}\\
\cmidrule(lr){4-6}\cmidrule(lr){8-10}\cmidrule(lr){12-14}\cmidrule(lr){16-18}
Method & Depth &  & \dmetric{W} & \dmetric{N} & \dmetric{Q} &  & \dmetric{W} & \dmetric{N} & \dmetric{Q} &  & \dmetric{W} & \dmetric{N} & \dmetric{Q} &  & \dmetric{W} & \dmetric{N} & \dmetric{Q}\\
\midrule
\addlinespace[0.15em]
\multicolumn{18}{l}{1.a) \textbf{Monocular 2D point trackers} \emph{(lifted with GT depth)}}\\
\midrule
PIPs~\cite{harley2022particle} & GT &  & 23.2 & 23.3 & 27.1 &  & 39.8 & 39.0 & 39.9 &  & 28.4 & 27.5 & 40.7 &  & 17.5 & 7.7 & 24.0\\
PIPs++~\cite{zheng2023pointodyssey} & GT &  & 27.2 & 28.4 & 33.4 &  & 42.4 & 41.5 & 42.5 &  & 30.3 & 26.8 & 42.9 &  & 18.8 & 8.3 & 25.4\\
TAPIR~\cite{doersch2023tapir} & GT &  & 26.3 & 26.9 & 32.2 &  & 45.7 & 44.6 & 45.8 &  & 34.6 & 32.1 & 50.0 &  & 20.8 & 9.2 & 28.4\\
CoTracker2~\cite{karaev2024cotracker} & GT &  & 31.3 & 32.2 & 37.5 &  & 47.5 & 45.9 & 47.5 &  & 35.5 & 32.3 & 50.6 &  & 22.3 & 9.8 & 29.7\\
BootsTAPIR~\cite{doersch2024bootstap} & GT &  & 29.3 & 29.2 & 35.2 &  & 46.9 & 45.6 & 46.9 &  & 36.7 & 34.3 & 53.2 &  & 24.1 & 10.6 & 32.7\\
CoTracker1~\cite{karaev2024cotracker} & GT &  & 31.8 & 32.4 & 37.9 &  & 47.9 & 46.5 & 47.9 &  & 36.2 & 33.3 & 51.3 &  & 22.5 & 10.0 & 29.8\\
LocoTrack~\cite{cho2024locotrack} & GT &  & 31.2 & 31.1 & 38.0 &  & 49.4 & 47.9 & 49.5 &  & 38.9 & 35.8 & 56.8 &  & 23.5 & 10.2 & 31.6\\
CoTracker3~\cite{karaev2024cotracker3} & GT &  & 32.8 & 33.3 & 39.4 &  & 48.3 & 46.9 & 48.3 &  & 38.1 & 35.0 & 54.2 &  & 25.0 & 11.1 & 33.3\\
TAPNext++~\cite{jung2026tapnextpp} & GT &  & 34.7 & 34.5 & 41.6 &  & 48.6 & 47.2 & 48.7 &  & 41.0 & 38.3 & 59.1 &  & 30.3 & 13.5 & 40.3\\
TAPTRv3~\cite{li2024taptrv3} & GT &  & 31.3 & 31.2 & 38.4 &  & 46.9 & 45.6 & 46.9 &  & 35.6 & 32.6 & 50.3 &  & 20.2 & 8.6 & 27.2\\
DOT~\cite{lemoing2024dense} & GT &  & 34.7 & 35.3 & 41.0 &  & 48.6 & 47.0 & 48.5 &  & 39.1 & 35.4 & 56.6 &  & 24.8 & 11.0 & 33.3\\
MegaFlow~\cite{zhang2026megaflow} & GT &  & 33.8 & 34.0 & 40.5 &  & 50.1 & 48.4 & 50.1 &  & 39.5 & 37.1 & 57.7 &  & 28.1 & 12.4 & 37.5\\
Track-On-R~\cite{aydemir2025trackon} & GT &  & 34.2 & 34.2 & 40.9 &  & 50.2 & 48.5 & 50.1 &  & 39.9 & 37.4 & 57.3 &  & 27.6 & 12.3 & 36.7\\
AllTracker~\cite{harley2025alltracker} & GT &  & 33.7 & 34.5 & 39.9 &  & 49.7 & 48.1 & 49.6 &  & 39.9 & 36.9 & 57.5 &  & 30.4 & 13.7 & 40.2\\
CoWTracker~\cite{lai2026cowtracker} & GT &  & 35.9 & 35.7 & 42.5 &  & 51.3 & 49.5 & 51.1 &  & 41.8 & 39.0 & 60.4 &  & 31.3 & 14.2 & \underline{41.5}\\
\midrule
\addlinespace[0.15em]
\multicolumn{18}{l}{1.b) \textbf{Monocular 3D point trackers} \emph{(with GT depth input)}}\\
\midrule
SpatialTracker~\cite{xiao2024spatialtracker} & GT &  & 16.5 & 17.4 & 22.7 &  & 2.2 & 2.3 & 2.3 &  & 17.9 & 14.3 & 27.1 &  & 9.0 & 4.7 & 13.8\\
SceneTracker~\cite{wang2024scenetracker} & GT &  & 21.1 & 20.5 & 28.5 &  & 31.2 & 31.1 & 31.8 &  & 26.9 & 23.9 & 37.4 &  & 19.4 & 9.4 & 26.9\\
DELTA~\cite{ngo2025delta} & GT &  & 27.3 & 27.5 & 33.8 &  & 36.3 & 35.8 & 37.0 &  & 29.6 & 26.2 & 40.5 &  & 16.8 & 8.0 & 23.8\\
SpatialTrackerV2~\cite{xiao2025spatialtrackerv2} & GT &  & 29.0 & 29.9 & 33.7 &  & 42.7 & 41.9 & 43.1 &  & 34.1 & 31.5 & 45.4 &  & 22.6 & 11.3 & 30.3\\
TAPIP3D~\cite{zhang2025tapip3d} & GT &  & 39.3 & 41.3 & \underline{42.7} &  & \textbf{52.9} & \textbf{50.8} & \textbf{52.1} &  & \underline{53.8} & \underline{52.5} & \textbf{64.4} &  & \textbf{43.2} & \underline{26.2} & \textbf{47.6}\\
\midrule
\addlinespace[0.15em]
\multicolumn{18}{l}{1.c) \textbf{Multi-view 2D point trackers} \emph{(lifted with GT depth)}}\\
\midrule
MV-TAP~\cite{koo2025mvtap} & GT &  & \textbf{48.2} & \textbf{58.2} & \textbf{49.9} &  & 46.6 & 46.5 & 47.2 &  & 43.3 & 44.2 & 51.0 &  & 21.5 & 11.2 & 26.5\\
\textcolor{gray}{MV-TAP (MV)$^\ddagger$~\cite{koo2025mvtap}} & \textcolor{gray}{GT} &  & \textcolor{gray}{56.2} & \textcolor{gray}{67.1} & \textcolor{gray}{54.5} &  & \textcolor{gray}{55.7} & \textcolor{gray}{56.3} & \textcolor{gray}{56.1} &  & \textcolor{gray}{54.7} & \textcolor{gray}{55.2} & \textcolor{gray}{63.4} &  & \textcolor{gray}{22.9} & \textcolor{gray}{12.7} & \textcolor{gray}{30.1}\\
\textcolor{gray}{MV-TAP (MT)$^\dagger$~\cite{koo2025mvtap}} & \textcolor{gray}{GT} &  & \textcolor{gray}{52.6} & \textcolor{gray}{63.2} & \textcolor{gray}{52.9} &  & \textcolor{gray}{55.7} & \textcolor{gray}{56.3} & \textcolor{gray}{56.2} &  & \textcolor{gray}{55.3} & \textcolor{gray}{57.3} & \textcolor{gray}{61.3} &  & \textcolor{gray}{29.2} & \textcolor{gray}{25.8} & \textcolor{gray}{30.6}\\
\midrule
\addlinespace[0.15em]
\multicolumn{18}{l}{1.d) \textbf{Multi-view 3D point trackers} \emph{(with GT depth input)}}\\
\midrule
MVTracker~\cite{rajic2025mvtracker} & GT &  & 34.1 & 38.2 & 35.9 &  & 48.6 & 46.5 & 47.4 &  & 51.1 & 49.2 & 56.4 &  & \underline{42.3} & \textbf{32.8} & 39.7\\
MVTracker (finetuned)~\cite{rajic2025mvtracker} & GT &  & \underline{39.5} & \underline{45.2} & 39.8 &  & \underline{52.4} & \underline{50.1} & \underline{51.2} &  & \textbf{58.3} & \textbf{55.3} & \underline{62.1} &  & \textcolor{gray}{57.0} & \textcolor{gray}{44.2} & \textcolor{gray}{48.8}\\
\bottomrule
\end{tabular}%
\end{table*}

\begin{table*}[!t]
\centering
\setlength{\tabcolsep}{1.8pt}
\caption{\textbf{Sensitivity to the reconstruction source.} One representative tracker per family of \cref{tab:main_results} is run on every feed-forward reconstruction source, each evaluated in two configurations (Frames): \emph{all}, one forward over all views and timesteps of the sequence, windowed and stitched when they exceed the model's image budget; or $t_0{+}t$, one forward per timestep $t$ on the $(t_0, t)$ frame pair of every view, expressed in the $t_0$ frame. DUSt3R's pairwise graph with global alignment is too slow for the $t_0{+}t$ variant. The main tables use all frames for VGGT-$\Omega$, except on DROID, where they use the anchored variant. GT rows use dataset depth and cameras (\cref{tab:gt_depth_results}) and are listed for reference on the splits with dense depth; \cref{tab:reconstruction_quality} reports the camera and depth quality of every source. Cells report \dmetric{W}, \dmetric{N}, and \dmetric{Q} as in \cref{tab:main_results}, and Overall averages the seven splits. Best and second-best estimated sources per tracker and column are bold and underlined.}
\label{tab:depth_source_results}
\vspace{-0.27cm}
\resizebox{\linewidth}{!}{%
\begin{tabular}{llccccccccccccccccccccccccccccccccc}
\toprule
 &  &  &  & \multicolumn{3}{c}{\textbf{Overall}} &  & \multicolumn{3}{c}{\textbf{DROID}} &  & \multicolumn{3}{c}{\textbf{EgoExo4D}} &  & \multicolumn{3}{c}{\textbf{Harmony4D}} &  & \multicolumn{3}{c}{\textbf{PACE}} &  & \multicolumn{3}{c}{\textbf{Hi4D}} &  & \multicolumn{3}{c}{\textbf{Waymo}} &  & \multicolumn{3}{c}{\textbf{Perpetua}}\\
\cmidrule(lr){5-7}\cmidrule(lr){9-11}\cmidrule(lr){13-15}\cmidrule(lr){17-19}\cmidrule(lr){21-23}\cmidrule(lr){25-27}\cmidrule(lr){29-31}\cmidrule(lr){33-35}
Method & Depth & Frames &  & \dmetric{W} & \dmetric{N} & \dmetric{Q} &  & \dmetric{W} & \dmetric{N} & \dmetric{Q} &  & \dmetric{W} & \dmetric{N} & \dmetric{Q} &  & \dmetric{W} & \dmetric{N} & \dmetric{Q} &  & \dmetric{W} & \dmetric{N} & \dmetric{Q} &  & \dmetric{W} & \dmetric{N} & \dmetric{Q} &  & \dmetric{W} & \dmetric{N} & \dmetric{Q} &  & \dmetric{W} & \dmetric{N} & \dmetric{Q}\\
\midrule
\addlinespace[0.15em]
\multicolumn{35}{l}{1.a) \textbf{Monocular 2D point trackers}}\\
\midrule
\multirow[c]{10}{*}{CoTracker3~\cite{karaev2024cotracker3}} & VGGT-$\Omega$~\cite{wang2026vggtomega} & all &  & \textbf{20.4} & \textbf{17.7} & \textbf{25.5} &  & \underline{4.2} & \underline{8.2} & 9.3 &  & \textbf{45.4} & \underline{42.2} & \textbf{53.8} &  & \textbf{21.3} & \textbf{17.4} & \textbf{27.5} &  & \textbf{42.8} & \textbf{34.7} & \textbf{35.3} &  & \underline{16.3} & 12.2 & 18.7 &  & 2.4 & 4.2 & 14.8 &  & \underline{10.6} & 5.1 & \underline{18.9}\\
 & VGGT-$\Omega$~\cite{wang2026vggtomega} & $t_0{+}t$ &  & \underline{18.9} & \underline{17.3} & 23.9 &  & \textbf{5.7} & \textbf{10.7} & \textbf{12.7} &  & \underline{44.0} & \textbf{43.3} & \underline{51.2} &  & \underline{16.7} & \underline{16.5} & \underline{26.8} &  & 42.2 & 31.6 & 32.5 &  & 15.2 & 11.7 & 18.2 &  & 1.7 & 3.2 & 10.3 &  & 6.5 & 4.2 & 15.6\\
 & VGGT~\cite{wang2026vggt} & all &  & 5.3 & 4.9 & 8.8 &  & 0.6 & 0.5 & 2.4 &  & 13.1 & 13.6 & 17.3 &  & 1.3 & 1.9 & 7.3 &  & 12.2 & 10.4 & 12.8 &  & 4.7 & 2.7 & 5.8 &  & 1.7 & 2.8 & 4.3 &  & 3.7 & 2.5 & 11.9\\
 & VGGT~\cite{wang2026vggt} & $t_0{+}t$ &  & 6.5 & 6.3 & 10.3 &  & 0.6 & 0.5 & 2.7 &  & 22.0 & 22.0 & 26.0 &  & 1.9 & 4.3 & 10.5 &  & 12.8 & 9.8 & 11.6 &  & 3.5 & 2.1 & 5.0 &  & 1.1 & 2.8 & 3.5 &  & 3.8 & 2.8 & 12.3\\
 & Pi3~\cite{wang2025pi3} & all &  & 17.7 & 15.9 & 22.1 &  & 0.6 & 6.8 & 9.9 &  & 42.1 & 37.4 & 41.0 &  & 13.1 & 12.4 & 18.9 &  & \underline{42.5} & 33.8 & 34.6 &  & 15.0 & 11.9 & 19.1 &  & 2.9 & 4.0 & \underline{15.2} &  & 7.5 & 5.1 & 16.1\\
 & Pi3~\cite{wang2025pi3} & $t_0{+}t$ &  & 15.5 & 13.9 & 19.7 &  & 0.9 & 6.6 & 10.5 &  & 32.6 & 30.0 & 35.4 &  & 11.6 & 13.3 & 21.2 &  & 41.1 & 30.3 & 31.2 &  & 15.0 & 11.1 & 17.6 &  & 2.4 & 2.4 & 9.8 &  & 4.9 & 3.8 & 12.6\\
 & Pi3X~\cite{wang2025pi3} & all &  & 18.7 & 17.2 & \underline{24.9} &  & 1.0 & 7.5 & 9.9 &  & 43.4 & 39.7 & 46.4 &  & 13.1 & 12.8 & 20.1 &  & 41.4 & \underline{34.2} & \underline{35.2} &  & \textbf{16.5} & \textbf{13.7} & \textbf{21.7} &  & \textbf{4.0} & \textbf{6.3} & \textbf{19.7} &  & \textbf{11.7} & \textbf{6.5} & \textbf{21.1}\\
 & Pi3X~\cite{wang2025pi3} & $t_0{+}t$ &  & 16.8 & 16.1 & 23.2 &  & 1.1 & 7.0 & \underline{10.7} &  & 39.4 & 38.7 & 46.0 &  & 10.9 & 13.5 & 22.2 &  & 39.3 & 30.3 & 31.1 &  & \textbf{16.5} & \underline{12.7} & \underline{20.0} &  & \underline{3.2} & \underline{5.2} & \underline{15.2} &  & 7.5 & \underline{5.3} & 17.5\\
 & DUSt3R~\cite{wang2024dust3r} & all &  & 6.3 & 4.1 & 5.9 &  & 0.1 & 0.7 & 1.9 &  & 11.9 & 10.1 & 16.3 &  & 1.5 & 1.2 & 2.9 &  & 23.8 & 11.5 & 12.1 &  & 6.4 & 4.9 & 7.2 &  & 0.1 & 0.0 & 0.3 &  & 0.1 & 0.1 & 0.4\\
 & GT &  &  &  &  &  &  & 32.8 & 33.3 & 39.4 &  &  &  &  &  &  &  &  &  & 48.3 & 46.9 & 48.3 &  & 38.1 & 35.0 & 54.2 &  &  &  &  &  & 25.0 & 11.1 & 33.3\\
\midrule
\addlinespace[0.15em]
\multicolumn{35}{l}{1.b) \textbf{Monocular 3D point trackers}}\\
\midrule
\multirow[c]{10}{*}{TAPIP3D~\cite{zhang2025tapip3d}} & VGGT-$\Omega$~\cite{wang2026vggtomega} & all &  & \textbf{23.3} & \textbf{21.2} & \textbf{26.8} &  & \underline{4.6} & \underline{9.5} & 9.3 &  & \textbf{51.1} & \underline{48.0} & \textbf{56.5} &  & \textbf{23.5} & \textbf{19.7} & \textbf{27.9} &  & \underline{44.1} & \textbf{35.5} & \textbf{35.8} &  & 20.5 & 15.8 & 19.6 &  & 3.5 & \underline{10.1} & 16.5 &  & \underline{15.5} & \underline{10.0} & \underline{22.2}\\
 & VGGT-$\Omega$~\cite{wang2026vggtomega} & $t_0{+}t$ &  & 21.0 & \underline{19.9} & 24.7 &  & \textbf{5.8} & \textbf{11.6} & \textbf{12.7} &  & \underline{49.2} & \textbf{48.8} & \underline{53.9} &  & \underline{18.2} & \underline{18.0} & \underline{26.5} &  & 43.4 & 32.2 & 32.9 &  & 19.0 & 15.3 & 19.1 &  & 2.3 & 6.2 & 11.0 &  & 8.9 & 7.4 & 16.8\\
 & VGGT~\cite{wang2026vggt} & all &  & 5.8 & 5.5 & 8.9 &  & 0.6 & 0.6 & 2.3 &  & 13.5 & 13.7 & 17.3 &  & 1.2 & 1.8 & 6.5 &  & 12.6 & 10.8 & 13.0 &  & 5.4 & 3.2 & 6.0 &  & 2.1 & 4.1 & 4.7 &  & 5.1 & 4.3 & 12.7\\
 & VGGT~\cite{wang2026vggt} & $t_0{+}t$ &  & 7.1 & 7.0 & 10.1 &  & 0.6 & 0.5 & 2.7 &  & 23.6 & 23.4 & 26.4 &  & 1.7 & 4.1 & 9.2 &  & 13.3 & 10.1 & 11.8 &  & 4.0 & 2.5 & 5.2 &  & 1.5 & 4.2 & 3.6 &  & 5.0 & 4.2 & 11.8\\
 & Pi3~\cite{wang2025pi3} & all &  & 20.0 & 19.1 & 23.5 &  & 0.6 & 7.4 & 10.0 &  & 47.0 & 42.1 & 43.5 &  & 14.6 & 14.2 & 19.4 &  & \textbf{44.3} & 34.8 & \underline{35.4} &  & 18.7 & 15.9 & 20.1 &  & 4.1 & 9.6 & \underline{16.8} &  & 10.9 & 9.7 & 19.3\\
 & Pi3~\cite{wang2025pi3} & $t_0{+}t$ &  & 17.2 & 16.0 & 20.5 &  & 0.9 & 6.7 & 10.4 &  & 36.3 & 33.6 & 37.4 &  & 12.6 & 14.9 & 21.4 &  & 42.5 & 31.2 & 32.0 &  & 18.5 & 14.7 & 18.5 &  & 3.1 & 4.8 & 10.3 &  & 6.5 & 6.1 & 13.4\\
 & Pi3X~\cite{wang2025pi3} & all &  & \underline{21.6} & \textbf{21.2} & \underline{26.7} &  & 1.0 & 8.6 & 10.1 &  & 48.5 & 44.6 & 49.0 &  & 14.7 & 14.7 & 20.7 &  & 42.6 & \underline{35.0} & \textbf{35.8} &  & \textbf{20.9} & \textbf{18.5} & \textbf{23.1} &  & \textbf{5.7} & \textbf{14.2} & \textbf{22.2} &  & \textbf{17.5} & \textbf{13.1} & \textbf{26.1}\\
 & Pi3X~\cite{wang2025pi3} & $t_0{+}t$ &  & 19.0 & 19.1 & 24.4 &  & 1.1 & 7.6 & \underline{10.9} &  & 44.4 & 43.9 & 48.7 &  & 12.2 & 15.1 & 22.6 &  & 40.1 & 31.1 & 31.8 &  & \underline{20.7} & \underline{17.0} & \underline{21.2} &  & \underline{4.2} & 9.9 & 16.1 &  & 10.4 & 9.4 & 19.5\\
 & DUSt3R~\cite{wang2024dust3r} & all &  & 6.5 & 4.3 & 5.8 &  & 0.1 & 0.6 & 1.6 &  & 12.6 & 11.0 & 16.5 &  & 1.4 & 1.1 & 2.2 &  & 24.4 & 11.7 & 12.2 &  & 7.2 & 5.8 & 7.4 &  & 0.1 & 0.1 & 0.3 &  & 0.1 & 0.1 & 0.3\\
 & GT &  &  &  &  &  &  & 39.3 & 41.3 & 42.7 &  &  &  &  &  &  &  &  &  & 52.9 & 50.8 & 52.1 &  & 53.8 & 52.5 & 64.4 &  &  &  &  &  & 43.2 & 26.2 & 47.6\\
\midrule
\addlinespace[0.15em]
\multicolumn{35}{l}{1.c) \textbf{Multi-view 2D point trackers}}\\
\midrule
\multirow[c]{10}{*}{MV-TAP~\cite{koo2025mvtap}} & VGGT-$\Omega$~\cite{wang2026vggtomega} & all &  & \textbf{19.8} & \textbf{18.3} & \textbf{23.1} &  & \underline{4.4} & \underline{9.4} & 8.9 &  & \textbf{47.6} & \underline{45.1} & \textbf{51.8} &  & \textbf{18.9} & \textbf{17.8} & \textbf{23.3} &  & \textbf{40.7} & \textbf{33.4} & \textbf{33.6} &  & \underline{17.0} & 13.8 & 17.0 &  & 2.1 & 4.0 & 12.7 &  & \underline{7.8} & 4.4 & \underline{14.1}\\
 & VGGT-$\Omega$~\cite{wang2026vggtomega} & $t_0{+}t$ &  & \underline{18.5} & \underline{17.9} & 21.7 &  & \textbf{5.8} & \textbf{12.0} & \textbf{11.9} &  & \underline{46.0} & \textbf{46.2} & \underline{49.5} &  & \underline{15.5} & \underline{16.9} & \underline{22.8} &  & \underline{40.2} & 30.5 & 31.0 &  & 15.7 & 13.3 & 16.4 &  & 1.5 & 3.0 & 9.0 &  & 4.9 & 3.5 & 11.5\\
 & VGGT~\cite{wang2026vggt} & all &  & 4.6 & 4.4 & 7.2 &  & 0.5 & 0.5 & 2.2 &  & 12.1 & 12.6 & 15.1 &  & 0.9 & 1.5 & 5.6 &  & 9.9 & 8.9 & 10.8 &  & 4.4 & 2.8 & 4.7 &  & 1.5 & 2.4 & 3.8 &  & 2.7 & 2.0 & 8.5\\
 & VGGT~\cite{wang2026vggt} & $t_0{+}t$ &  & 5.9 & 5.9 & 8.5 &  & 0.6 & 0.5 & 2.5 &  & 21.6 & 21.9 & 23.7 &  & 1.4 & 3.6 & 8.0 &  & 10.2 & 8.4 & 9.6 &  & 3.4 & 2.2 & 4.1 &  & 1.0 & 2.3 & 3.1 &  & 2.9 & 2.3 & 8.8\\
 & Pi3~\cite{wang2025pi3} & all &  & 16.9 & 16.0 & 19.9 &  & 0.6 & 7.6 & 9.4 &  & 43.2 & 38.6 & 39.4 &  & 11.7 & 12.3 & 15.9 &  & 39.7 & 32.3 & 32.3 &  & 15.3 & 13.4 & 16.8 &  & 2.5 & 3.7 & \underline{13.2} &  & 5.4 & 4.3 & 12.0\\
 & Pi3~\cite{wang2025pi3} & $t_0{+}t$ &  & 14.6 & 13.8 & 17.5 &  & 0.9 & 7.1 & 9.8 &  & 31.8 & 29.9 & 31.9 &  & 10.5 & 12.9 & 18.0 &  & 37.9 & 29.0 & 29.3 &  & 15.2 & 12.4 & 15.4 &  & 2.1 & 2.1 & 8.5 &  & 3.6 & 3.2 & 9.3\\
 & Pi3X~\cite{wang2025pi3} & all &  & 18.1 & 17.4 & \underline{22.3} &  & 0.9 & 8.5 & 9.4 &  & 45.6 & 41.6 & 44.9 &  & 12.0 & 12.3 & 16.8 &  & 38.5 & \underline{32.5} & \underline{32.7} &  & \textbf{17.2} & \textbf{15.6} & \textbf{19.3} &  & \textbf{3.5} & \textbf{5.7} & \textbf{16.9} &  & \textbf{8.8} & \textbf{5.7} & \textbf{16.0}\\
 & Pi3X~\cite{wang2025pi3} & $t_0{+}t$ &  & 16.2 & 16.3 & 20.8 &  & 1.0 & 7.9 & \underline{10.1} &  & 40.8 & 40.7 & 44.1 &  & 9.5 & 12.8 & 18.5 &  & 36.8 & 29.2 & 29.4 &  & \underline{17.0} & \underline{14.4} & \underline{17.5} &  & \underline{2.7} & \underline{4.5} & 12.9 &  & 5.7 & \underline{4.6} & 13.1\\
 & DUSt3R~\cite{wang2024dust3r} & all &  & 5.9 & 4.1 & 5.3 &  & 0.1 & 0.7 & 1.7 &  & 11.3 & 10.1 & 14.4 &  & 1.3 & 1.2 & 2.3 &  & 22.4 & 11.6 & 11.9 &  & 6.4 & 5.3 & 6.5 &  & 0.1 & 0.0 & 0.3 &  & 0.0 & 0.1 & 0.3\\
 & GT &  &  &  &  &  &  & 48.2 & 58.2 & 49.9 &  &  &  &  &  &  &  &  &  & 46.6 & 46.5 & 47.2 &  & 43.3 & 44.2 & 51.0 &  &  &  &  &  & 21.5 & 11.2 & 26.5\\
\midrule
\addlinespace[0.15em]
\multicolumn{35}{l}{1.d) \textbf{Multi-view 3D point trackers}}\\
\midrule
\multirow[c]{10}{*}{MVTracker~\cite{rajic2025mvtracker}} & VGGT-$\Omega$~\cite{wang2026vggtomega} & all &  & \textbf{22.6} & \underline{21.4} & \underline{24.2} &  & \underline{3.9} & \underline{9.0} & 8.1 &  & \textbf{48.8} & \underline{45.7} & \textbf{50.7} &  & \textbf{23.2} & \textbf{20.4} & \textbf{24.0} &  & \underline{43.7} & \textbf{35.2} & \textbf{35.4} &  & 19.1 & 15.1 & 17.8 &  & 3.5 & 11.2 & 15.1 &  & \underline{16.1} & \underline{13.4} & \underline{18.4}\\
 & VGGT-$\Omega$~\cite{wang2026vggtomega} & $t_0{+}t$ &  & 20.4 & 19.8 & 21.8 &  & \textbf{5.2} & \textbf{11.0} & \textbf{11.0} &  & 46.9 & \textbf{45.9} & \underline{47.9} &  & \underline{17.7} & \underline{18.4} & \underline{21.4} &  & 42.4 & 32.0 & 32.5 &  & 17.9 & 14.5 & 17.2 &  & 2.5 & 6.9 & 9.5 &  & 10.0 & 10.2 & 13.5\\
 & VGGT~\cite{wang2026vggt} & all &  & 5.7 & 5.9 & 7.6 &  & 0.5 & 0.6 & 2.1 &  & 13.3 & 14.3 & 16.2 &  & 1.0 & 1.8 & 3.9 &  & 12.4 & 10.5 & 12.2 &  & 5.2 & 3.4 & 4.7 &  & 2.3 & 4.7 & 4.7 &  & 5.6 & 6.1 & 9.4\\
 & VGGT~\cite{wang2026vggt} & $t_0{+}t$ &  & 7.1 & 7.2 & 8.4 &  & 0.6 & 0.5 & 2.4 &  & 23.7 & 23.6 & 24.2 &  & 1.5 & 3.5 & 5.5 &  & 12.8 & 9.7 & 11.0 &  & 3.9 & 2.7 & 4.0 &  & 1.5 & 4.6 & 3.5 &  & 5.8 & 5.8 & 8.1\\
 & Pi3~\cite{wang2025pi3} & all &  & 19.4 & 19.1 & 21.4 &  & 0.6 & 7.2 & 9.1 &  & 44.1 & 39.3 & 39.9 &  & 14.1 & 14.4 & 17.0 &  & \textbf{43.9} & \underline{33.4} & \underline{33.9} &  & 17.6 & 15.1 & 17.8 &  & \underline{4.3} & \underline{11.7} & \underline{15.9} &  & 11.6 & 12.5 & 16.5\\
 & Pi3~\cite{wang2025pi3} & $t_0{+}t$ &  & 17.0 & 15.7 & 17.9 &  & 0.8 & 6.6 & 9.0 &  & 34.7 & 30.8 & 32.0 &  & 12.5 & 14.7 & 17.7 &  & 43.1 & 30.1 & 30.5 &  & 17.0 & 13.6 & 16.0 &  & 3.1 & 5.7 & 8.9 &  & 7.6 & 8.2 & 11.0\\
 & Pi3X~\cite{wang2025pi3} & all &  & \underline{21.0} & \textbf{21.6} & \textbf{24.4} &  & 0.8 & 8.2 & 9.0 &  & \underline{47.2} & 43.3 & 46.1 &  & 14.7 & 14.9 & 18.5 &  & 39.5 & 33.2 & 33.1 &  & \textbf{19.9} & \textbf{17.7} & \textbf{20.8} &  & \textbf{5.9} & \textbf{16.8} & \textbf{21.2} &  & \textbf{18.8} & \textbf{17.3} & \textbf{22.1}\\
 & Pi3X~\cite{wang2025pi3} & $t_0{+}t$ &  & 18.4 & 19.0 & 21.5 &  & 1.0 & 7.1 & \underline{9.3} &  & 42.4 & 41.5 & 44.0 &  & 12.2 & 15.2 & 19.1 &  & 38.0 & 29.8 & 30.2 &  & \underline{19.5} & \underline{16.0} & \underline{18.4} &  & \underline{4.3} & 10.9 & 14.1 &  & 11.6 & 12.2 & 15.5\\
 & DUSt3R~\cite{wang2024dust3r} & all &  & 6.4 & 4.2 & 5.0 &  & 0.1 & 0.6 & 1.0 &  & 12.6 & 10.4 & 13.5 &  & 1.5 & 1.0 & 1.8 &  & 23.7 & 11.6 & 11.9 &  & 6.9 & 5.6 & 6.5 &  & 0.1 & 0.1 & 0.3 &  & 0.1 & 0.1 & 0.2\\
 & GT &  &  &  &  &  &  & 34.1 & 38.2 & 35.9 &  &  &  &  &  &  &  &  &  & 48.6 & 46.5 & 47.4 &  & 51.1 & 49.2 & 56.4 &  &  &  &  &  & 42.3 & 32.8 & 39.7\\
\bottomrule
\end{tabular}%
}
\end{table*}

\begin{table*}[!t]
\centering
\setlength{\tabcolsep}{2.1pt}
\caption{\textbf{Detailed query-view location-accuracy thresholds.} Each cell reports \dthresh{k}{Q}, the percentage of ground-truth-visible track positions within the specified 3D threshold under query-median camera-space scaling. Rows with off-the-shelf or self-estimated depth are listed first within each method family, followed by the corresponding GT-depth rows. Best and second-best entries are bold and underlined. Gray cells belong to methods trained on that subset's source data (\cref{tab:training_data}) and are excluded from ranking.}
\label{tab:query_view_thresholds}
\vspace{-0.27cm}
\resizebox{\linewidth}{!}{%
\begin{tabular}{lccccccccccccccccccccccccccccccccccccccccccc}
\toprule
 &  &  & \multicolumn{5}{c}{\textbf{DROID}} &  & \multicolumn{5}{c}{\textbf{EgoExo4D}} &  & \multicolumn{5}{c}{\textbf{Harmony4D}} &  & \multicolumn{5}{c}{\textbf{PACE}} &  & \multicolumn{5}{c}{\textbf{Hi4D}} &  & \multicolumn{5}{c}{\textbf{Waymo}} &  & \multicolumn{5}{c}{\textbf{Perpetua}}\\
\cmidrule(lr){4-8}\cmidrule(lr){10-14}\cmidrule(lr){16-20}\cmidrule(lr){22-26}\cmidrule(lr){28-32}\cmidrule(lr){34-38}\cmidrule(lr){40-44}
Method & Depth &  & 1 & 2 & 4 & 8 & 16 &  & 1 & 2 & 4 & 8 & 16 &  & 1 & 2 & 4 & 8 & 16 &  & 1 & 2 & 4 & 8 & 16 &  & 1 & 2 & 4 & 8 & 16 &  & 1 & 2 & 4 & 8 & 16 &  & 1 & 2 & 4 & 8 & 16\\
\midrule
\addlinespace[0.15em]
\multicolumn{44}{l}{1.a) \textbf{Monocular 2D point trackers}}\\
\midrule
PIPs~\cite{harley2022particle} & VGGT-$\Omega$ &  & 0.1 & 0.4 & 2.9 & 14.2 & 37.0 &  & 3.5 & 15.8 & 42.5 & 67.1 & 80.3 &  & 0.7 & 3.8 & 14.1 & 32.8 & 51.4 &  & 0.3 & 2.6 & 15.7 & 48.5 & 87.8 &  & 0.0 & 0.1 & 1.4 & 21.0 & 54.7 &  & \textbf{0.1} & 0.6 & 4.0 & 18.0 & 47.0 &  & 0.4 & 1.9 & 7.3 & 19.7 & 39.7\\
PIPs++~\cite{zheng2023pointodyssey} & VGGT-$\Omega$ &  & 0.1 & 0.4 & 3.4 & 15.8 & 38.3 &  & 7.7 & 24.6 & 53.4 & 76.9 & 86.7 &  & 1.4 & 6.3 & 18.8 & 38.2 & 56.9 &  & 0.4 & 3.3 & 18.1 & 50.0 & 88.6 &  & 0.0 & 0.1 & 1.7 & 23.9 & 54.1 &  & \textbf{0.1} & 0.5 & 3.5 & 15.8 & 42.6 &  & 0.5 & 2.4 & 8.5 & 21.5 & 41.1\\
TAPTRv3~\cite{li2024taptrv3} & VGGT-$\Omega$ &  & 0.1 & 0.5 & 3.8 & 16.9 & 40.6 &  & 9.9 & 28.2 & 56.6 & 77.5 & 86.7 &  & 1.6 & 6.7 & 17.9 & 35.6 & 52.8 &  & 0.5 & 4.4 & 22.5 & 53.7 & 90.9 &  & 0.0 & 0.1 & 1.5 & 27.1 & 57.9 &  & \textbf{0.1} & \underline{0.7} & 4.1 & 17.5 & 45.9 &  & 0.6 & 2.8 & 9.0 & 21.6 & 40.6\\
\bottomrule
\end{tabular}%
}
\end{table*}

\begin{table*}[!t]
\centering
\setlength{\tabcolsep}{2.1pt}
\caption{\textbf{Detailed non-query-view location-accuracy thresholds.} Each cell reports \dthresh{k}{N}, the percentage of ground-truth-visible track positions within the specified 3D threshold under query-median camera-space scaling. Rows with off-the-shelf or self-estimated depth are listed first within each method family, followed by the corresponding GT-depth rows. Best and second-best entries are bold and underlined. Gray cells belong to methods trained on that subset's source data (\cref{tab:training_data}) and are excluded from ranking.}
\label{tab:non_query_view_thresholds}
\vspace{-0.27cm}
\resizebox{\linewidth}{!}{%
\begin{tabular}{lccccccccccccccccccccccccccccccccccccccccccc}
\toprule
 &  &  & \multicolumn{5}{c}{\textbf{DROID}} &  & \multicolumn{5}{c}{\textbf{EgoExo4D}} &  & \multicolumn{5}{c}{\textbf{Harmony4D}} &  & \multicolumn{5}{c}{\textbf{PACE}} &  & \multicolumn{5}{c}{\textbf{Hi4D}} &  & \multicolumn{5}{c}{\textbf{Waymo}} &  & \multicolumn{5}{c}{\textbf{Perpetua}}\\
\cmidrule(lr){4-8}\cmidrule(lr){10-14}\cmidrule(lr){16-20}\cmidrule(lr){22-26}\cmidrule(lr){28-32}\cmidrule(lr){34-38}\cmidrule(lr){40-44}
Method & Depth &  & 1 & 2 & 4 & 8 & 16 &  & 1 & 2 & 4 & 8 & 16 &  & 1 & 2 & 4 & 8 & 16 &  & 1 & 2 & 4 & 8 & 16 &  & 1 & 2 & 4 & 8 & 16 &  & 1 & 2 & 4 & 8 & 16 &  & 1 & 2 & 4 & 8 & 16\\
\midrule
\addlinespace[0.15em]
\multicolumn{44}{l}{1.a) \textbf{Monocular 2D point trackers}}\\
\midrule
PIPs~\cite{harley2022particle} & VGGT-$\Omega$ &  & 0.0 & 0.3 & 2.1 & 11.8 & 32.2 &  & 1.2 & 7.7 & 31.2 & 58.9 & 75.4 &  & 0.3 & 1.8 & 7.8 & 21.5 & 38.7 &  & 0.3 & 2.3 & 13.4 & 47.1 & 86.9 &  & 0.0 & 0.2 & 1.8 & 12.5 & 37.0 &  & \underline{0.0} & 0.1 & 0.7 & 4.0 & 14.6 &  & 0.0 & 0.2 & 1.4 & 5.0 & 11.7\\
PIPs++~\cite{zheng2023pointodyssey} & VGGT-$\Omega$ &  & 0.0 & 0.3 & 2.3 & 12.5 & 33.0 &  & 1.6 & 9.7 & 37.2 & 67.2 & 81.3 &  & 0.5 & 2.5 & 9.8 & 23.9 & 40.6 &  & 0.4 & 2.7 & 15.8 & 51.3 & 88.1 &  & 0.0 & 0.2 & 1.9 & 13.0 & 35.8 &  & \underline{0.0} & 0.1 & 0.6 & 3.5 & 13.5 &  & 0.0 & 0.3 & 1.7 & 5.6 & 12.3\\
TAPTRv3~\cite{li2024taptrv3} & VGGT-$\Omega$ &  & 0.0 & 0.3 & 2.4 & 13.1 & 34.3 &  & 1.8 & 10.9 & 39.8 & 68.4 & 81.8 &  & 0.5 & 2.6 & 9.5 & 23.2 & 39.8 &  & 0.5 & 3.4 & 19.1 & 55.2 & 90.1 &  & 0.0 & 0.3 & 2.2 & 14.5 & 39.2 &  & \underline{0.0} & 0.1 & 0.8 & 4.2 & 15.6 &  & 0.0 & 0.3 & 1.6 & 5.4 & 12.0\\
\bottomrule
\end{tabular}%
}
\end{table*}

\begin{table*}[!t]
\centering
\setlength{\tabcolsep}{2.1pt}
\caption{\textbf{Detailed world-space metrics.} \dmetric{W} and \dthresh{k}{W} are percentages of ground-truth-visible track positions under query-median scaling, while EPE is their mean Euclidean world-space error in meters under the same scaling. Rows with off-the-shelf or self-estimated depth are listed first within each method family, followed by the corresponding GT-depth rows. Best and second-best entries are bold and underlined, with lower being better for EPE. EPE is $\infty$ where a method returned no prediction for some ground-truth-visible positions. Such positions count as misses in the percentage metrics, but make the mean error infinite. Entries marked -- are not defined by the method output, world-space tracks for that dataset or a visibility estimate for OA\textsubscript{glob}. Gray cells belong to methods trained on that subset's source data and are excluded from ranking.}
\label{tab:world_metrics}
\vspace{-0.27cm}
\resizebox{\linewidth}{!}{%
\begin{tabular}{lcccccccccccccccccccccccccccccccccccccccccccccccccccccccccccccccc}
\toprule
 &  &  & \multicolumn{8}{c}{\textbf{DROID}} &  & \multicolumn{8}{c}{\textbf{EgoExo4D}} &  & \multicolumn{8}{c}{\textbf{Harmony4D}} &  & \multicolumn{8}{c}{\textbf{PACE}} &  & \multicolumn{8}{c}{\textbf{Hi4D}} &  & \multicolumn{8}{c}{\textbf{Waymo}} &  & \multicolumn{8}{c}{\textbf{Perpetua}}\\
\cmidrule(lr){4-11}\cmidrule(lr){13-20}\cmidrule(lr){22-29}\cmidrule(lr){31-38}\cmidrule(lr){40-47}\cmidrule(lr){49-56}\cmidrule(lr){58-65}
Method & Depth &  & \dmetric{W}\,\(\uparrow\) & OA\textsubscript{glob}\,\(\uparrow\) & EPE\,\(\downarrow\) & $<1$ & $<2$ & $<4$ & $<8$ & $<16$ &  & \dmetric{W}\,\(\uparrow\) & OA\textsubscript{glob}\,\(\uparrow\) & EPE\,\(\downarrow\) & $<1$ & $<2$ & $<4$ & $<8$ & $<16$ &  & \dmetric{W}\,\(\uparrow\) & OA\textsubscript{glob}\,\(\uparrow\) & EPE\,\(\downarrow\) & $<1$ & $<2$ & $<4$ & $<8$ & $<16$ &  & \dmetric{W}\,\(\uparrow\) & OA\textsubscript{glob}\,\(\uparrow\) & EPE\,\(\downarrow\) & $<1$ & $<2$ & $<4$ & $<8$ & $<16$ &  & \dmetric{W}\,\(\uparrow\) & OA\textsubscript{glob}\,\(\uparrow\) & EPE\,\(\downarrow\) & $<1$ & $<2$ & $<4$ & $<8$ & $<16$ &  & \dmetric{W}\,\(\uparrow\) & OA\textsubscript{glob}\,\(\uparrow\) & EPE\,\(\downarrow\) & $<1$ & $<2$ & $<4$ & $<8$ & $<16$ &  & \dmetric{W}\,\(\uparrow\) & OA\textsubscript{glob}\,\(\uparrow\) & EPE\,\(\downarrow\) & $<1$ & $<2$ & $<4$ & $<8$ & $<16$\\
\midrule
\addlinespace[0.15em]
\multicolumn{65}{l}{1.a) \textbf{Monocular 2D point trackers}}\\
\midrule
PIPs~\cite{harley2022particle} & VGGT-$\Omega$ &  & 4.8 & 87.5 & 0.17 & 0.0 & 0.1 & 0.6 & 3.9 & 19.4 &  & 36.9 & 91.4 & 0.73 & 2.0 & 10.8 & 33.9 & 61.6 & 76.1 &  & 16.9 & 78.6 & 2.47 & 0.3 & 1.7 & 8.8 & 26.9 & 46.9 &  & 36.6 & 97.7 & \underline{0.03} & 0.5 & 4.1 & 24.4 & 66.0 & 88.1 &  & 13.5 & 80.0 & 3.00 & 0.1 & 0.5 & 4.2 & 20.7 & 42.0 &  & 2.2 & 66.5 & 8.59 & \textbf{0.0} & \underline{0.0} & 0.3 & 1.9 & 8.9 &  & 7.7 & 56.6 & 2.59 & 0.1 & 0.6 & 3.2 & 10.9 & 23.6\\
\bottomrule
\end{tabular}%
}
\end{table*}

\section{Extended Task and Metric Formalization}
\label{sec:supp_metrics}

This section provides the full mathematical formulations for the task inputs, outputs, query sampling, and metric evaluations deferred from \cref{sec:task} of the main text.

\paragraph{Inputs and Outputs.}
A model is given $V$ time-synchronized RGB video streams of a dynamic scene, captured from different viewpoints. Each view $v \in \{1, \dots, V\}$ is a sequence of $F$ frames $I^v_1, \dots, I^v_F$, with $I^v_t \in \mathbb{R}^{H \times W \times 3}$. Corresponding frames across views are time-synchronized to within 1/FPS. Accompanying the videos, the model receives a set of $P$ query points to define the tracking task. Each query $q_p = (x_p, y_p, t_p, v_p)$ specifies a single pixel location $(x_p, y_p)$ observed in view $v_p$ at frame $t_p$; that is, every query is anchored in \emph{exactly one} view. 

For each query point $p$, the model must predict its 3D trajectory $\hat{\mathbf{X}}_{p,1}, \dots, \hat{\mathbf{X}}_{p,F}$, with $\hat{\mathbf{X}}_{p,t} \in \mathbb{R}^3$, in a single world coordinate frame shared across all views, together with a per-view visibility $\hat{o}^v_{p,t} \in \{0, 1\}$ indicating whether point $p$ is visible from view $v$ at frame $t$. Since no ground-truth extrinsics or depth are provided, this frame is the model's own. World-space scoring expresses predictions and ground truth relative to their respective first-camera poses and fits only one scale factor, as detailed below. For the per-view metrics the model additionally reports each query's position $\hat{\mathbf{X}}^v_{p,t}$ in the camera frame of every view $v$, which a model with a consistent world frame obtains by applying its own estimated pose for that view. Baselines that do not produce a shared world frame are evaluated only on the per-view metrics they support, and their unsupported metrics are left undefined. Per-view visibility is optional, and models that do not predict it are scored on position accuracy alone.

\paragraph{Query sampling.}
Ground-truth tracks $\mathbf{X}_{p,t} \in \mathbb{R}^3$ are given in world space and shared across all views, accompanied by per-view, per-frame visibility labels $o^v_{p,t}$. For each track, the query $(x_p, y_p, t_p, v_p)$ is sampled uniformly at random from the set of $(\text{frame}, \text{view})$ pairs in which the track is visible. Every track is visible in at least one such pair; tracks that are never visible in any view are excluded from the benchmark. The complete array and camera conventions are specified in \cref{sec:file_spec}.

\paragraph{Per-view metrics (Q and N).}
We adapt the TAP-3D metrics~\cite{koppula2024tapvid3d} to the multi-view setting by scoring every query in every camera view. We retain its metric definitions and depth-adaptive threshold $\tau_{3D} = \tau_{2D} \cdot Z / f$, with $\tau_{2D} \in \{1,2,4,8,16\}$, and compute $\mathrm{AJ}^{3D}$, $\delta^{3D}$, and $\mathrm{OA}$ per view before averaging across views. Because no ground-truth geometry is given as input, predictions are recovered only up to scale. For each camera view, the query-median scale is the median ground-truth 3D distance from the camera divided by the median predicted distance at query times, computed over tracks visible in that view. We apply this one factor to both track subsets scored in the view: tracks whose query lies there (\dmetric{Q}) and tracks queried in another view (\dmetric{N}). Which tracks fall into either subset is fixed by the dataset's queries, one per track, and does not depend on how many views a method consumes, so monocular and multi-view methods, and runs with fewer input views in \cref{fig:views_scaling}, are scored on the same tracks. Q and N nevertheless score different sets of visible 3D locations, so their numerical difference does not isolate the effect of crossing views. Queries a model declines to predict are scored as failures rather than excluded.

\paragraph{World-space metrics.}
While N probes cross-view correspondence, the per-view metrics score each view in its own camera frame, under a scale estimated independently per view and against a threshold adapted to that camera's viewing distance. They therefore do not evaluate whether the recovered trajectory remains correct in a fixed world frame as the cameras move. We additionally score all $P$ queries in one world frame, defined by the pose and orientation of the first camera at the first frame with metric scale following~\cite{feng2025st4rtrack}.

Points are weighted by union visibility $g_{p,t} = 1 - \prod_v (1 - o^v_{p,t})$, so a model must place a track whenever it is visible in at least one view. We fix one threshold per sequence, $\tau_{3D} = \tau_{2D} \cdot \mathrm{median}_p (Z_p / f^{v_p})$, where $Z_p$ is the ground-truth depth of query $p$ in its query view and $f^{v_p}=\sqrt{f_x^{v_p}f_y^{v_p}}$ is that view's geometric-mean focal length at the metric evaluation resolution. The threshold is applied uniformly across all tracks and frames: a depth-adaptive threshold would measure accuracy relative to each camera rather than in the common frame, while the fixed absolute thresholds used in prior work~\cite{feng2025st4rtrack,rajic2025mvtracker} do not transfer across scenes of widely varying scale. We report \dmetric{W}, threshold-specific location accuracy, and EPE after expressing both predictions and ground truth relative to their respective first-camera poses and applying one query-median scale across the sequence. We additionally report global occlusion accuracy when a method predicts visibility. Models that estimate their own cameras are anchored by their predicted reference pose, so pose drift is charged to the reconstruction rather than removed by construction.

\section{Baseline Implementation Details}
\label{sec:baseline_impl}

\paragraph{Shared inputs for Group~1.}
All group-1 point-tracker baselines share one evaluation pipeline. Each tracker is paired with a reconstruction source. The source supplies per-view depth maps and cameras. Every sequence is resized to one shared evaluation resolution with 512 pixels on the longer side and the shorter side rounded to the aspect-preserving multiple of 8. The landscape exocentric views of Harmony4D and EgoExo4D are center-cropped to square at load and become $512\times512$. All trackers and reconstruction sources consume these same images.

\paragraph{Reconstruction source.}
In \cref{tab:main_results} the source is the shared VGGT-$\Omega$ reconstruction. One forward pass covers all views and timesteps jointly, and sequences beyond 500 images run in overlapping windows with eight shared timesteps. Consecutive windows are aligned on their shared frames: both windows unproject the same pixel grid over those frames, and the similarity transform that maps one point set onto the other is fitted by least squares on the resulting dense correspondences, with the largest fifth of the residuals dropped and the fit repeated.

On DROID we instead serve a per-timestep variant anchored on the first timestep. Nothing in VGGT-$\Omega$ can state that two images were captured at the same timestep, since it carries no frame-index embeddings. Its training data also holds little synchronized multi-view footage. The per-timestep variant works around both by grouping the views by timestep, so that each pass covers two timesteps rather than the whole sequence. A single forward pass over the views of timestep~0 defines the reference geometry. Every other timestep~$t$ then runs its own forward pass over the views of timestep~0 together with the views of timestep~$t$. The half of that pass covering timestep~0 is aligned onto the reference by a similarity transform, and the same transform places the half covering timestep~$t$. Each timestep therefore inherits the reference frame through a reconstruction that saw both timesteps at once. Although this produces worse camera poses, it reconstructs DROID's moving robot arm and gripper more faithfully than the all-frames pass.

The benchmark therefore fixes the per-timestep variant for DROID and the all-frames variant for every other subset. The reconstruction is handed to the trackers in the model's own coordinate frame and scale, so no ground-truth geometry enters their input, and the evaluation rescales every prediction as described in \cref{sec:models}. In the GT-depth tables the source is the dataset geometry. The methods of group~2 estimate their own geometry instead.

\paragraph{Queries and tracking direction.}
The benchmark provides one query observation per track: one pixel in one view at one timestep. This query is unprojected through the source depth into a 3D query point. Every group-1 tracker runs twice, once on the original video and once on the time-reversed video, and the two half-tracks are merged at the query timestep. Predicted world-space trajectories are transformed into the coordinate frame of every source camera for the camera-space metrics. The following subsections match the method groups of \cref{tab:main_results}, and \cref{tab:training_data} lists the training data of every baseline.

\begin{table*}[t]
\centering
\caption{\textbf{Training data of the evaluated baselines.} Bold marks training data that overlaps the source data of an evaluation subset. Such results are gray in the result tables. Overlap inherited through a pretrained backbone, namely St4RTrack's frozen MASt3R encoder, is listed but not grayed.}
\label{tab:training_data}
\vspace{-0.27cm}
\footnotesize
\begin{tabular}{l p{0.72\linewidth}}
\toprule
Method & Training data\\
\midrule
\multicolumn{2}{l}{1.a) \textbf{Monocular 2D point trackers} \emph{(lifted with VGGT-$\Omega$~\cite{wang2026vggtomega})}}\\
\midrule
CoTracker1~\cite{karaev2024cotracker} & TAP-Vid-Kubric~\cite{doersch2022tapvid} (MOVi-f)\\
CoTracker2~\cite{karaev2024cotracker} & TAP-Vid-Kubric~\cite{doersch2022tapvid}\\
CoTracker3~\cite{karaev2024cotracker3} & Kubric~\cite{karaev2024cotracker3} + 15k pseudo-labelled real web videos~\cite{karaev2024cotracker3}\\
LocoTrack~\cite{cho2024locotrack} & Panning MOVi-E~\cite{doersch2023tapir}\\
PIPs~\cite{harley2022particle} & FlyingThings++~\cite{harley2022pips,mayer2016flyingthings}\\
PIPs++~\cite{zheng2023pointodyssey} & PointOdyssey~\cite{zheng2023pointodyssey}\\
TAPIR~\cite{doersch2023tapir} & Panning MOVi-E~\cite{doersch2023tapir}\\
BootsTAPIR~\cite{doersch2024bootstap} & Kubric~\cite{greff2022kubric} + 15M real web clips (TAPTube~\cite{doersch2024bootstap}, unreleased)\\
DOT~\cite{lemoing2024dense} & Kubric CVO + MOVi-F~\cite{lemoing2024dense}, for its RAFT and CoTracker2 components\\
TAPTRv3~\cite{li2024taptrv3} & TAP-Vid-Kubric~\cite{doersch2022tapvid} (MOVi-F)\\
Track-On-R~\cite{aydemir2025trackon} & Kubric~\cite{karaev2024cotracker3} + pseudo-labelled TAO~\cite{dave2020tao}, OVIS~\cite{qi2022ovis}, VSPW~\cite{miao2021vspw}~\cite{aydemir2025trackon}\\
MegaFlow~\cite{zhang2026megaflow} & FlyingChairs~\cite{dosovitskiy2015flownet}, TartanAir~\cite{wang2020tartanair}, FlyingThings3D~\cite{mayer2016flyingthings}, Sintel~\cite{butler2012sintel}, KITTI~\cite{geiger2012kitti}, HD1K~\cite{kondermann2016hd1k}, then a Kubric fine-tune~\cite{zhang2026megaflow,harley2025alltracker}\\
AllTracker~\cite{harley2025alltracker} & Kubric~\cite{greff2022kubric}, DynamicReplica~\cite{karaev2023dynamicstereo}, PointOdyssey~\cite{zheng2023pointodyssey}, FlyingChairs~\cite{dosovitskiy2015flownet}, FlyingThings3D, Monkaa, Driving~\cite{mayer2016flyingthings,harley2025alltracker}, AutoFlow~\cite{sun2021autoflow}, Spring~\cite{mehl2023spring}, VIPER~\cite{richter2017viper}, HD1K~\cite{kondermann2016hd1k}, KITTI~\cite{geiger2012kitti}, TartanAir~\cite{wang2020tartanair}\\
CoWTracker~\cite{lai2026cowtracker} & Kubric~\cite{greff2022kubric} (tracker); its VGGT~\cite{wang2025vggt} backbone is pretrained on a mixture including PointOdyssey~\cite{zheng2023pointodyssey}\\
TAPNext++~\cite{jung2026tapnextpp} & Kubric~\cite{greff2022kubric} + TAPTube~\cite{doersch2024bootstap} init, fine-tuned on PointOdyssey~\cite{zheng2023pointodyssey} + Kubric\\
\midrule
\multicolumn{2}{l}{1.b) \textbf{Monocular 3D point trackers} \emph{(with VGGT-$\Omega$~\cite{wang2026vggtomega} depth input)}}\\
\midrule
SceneTracker~\cite{wang2024scenetracker} & LSFOdyssey~\cite{wang2024scenetracker} (augmented PointOdyssey~\cite{zheng2023pointodyssey})\\
DELTA~\cite{ngo2025delta} & Kubric MOVi-F RGB-D~\cite{ngo2025delta}\\
SpatialTracker~\cite{xiao2024spatialtracker} & TAP-Vid-Kubric RGB-D~\cite{xiao2024spatialtracker}\\
TAPIP3D~\cite{zhang2025tapip3d} & Kubric MOVi-F~\cite{zhang2025tapip3d} (CoTracker3~\cite{karaev2024cotracker3} encoder init)\\
\midrule
\multicolumn{2}{l}{1.c) \textbf{Multi-view 2D point trackers} \emph{(lifted with VGGT-$\Omega$~\cite{wang2026vggtomega})}}\\
\midrule
MV-TAP~\cite{koo2025mvtap} & Multi-view Kubric~\cite{koo2025mvtap} (5k scenes $\times$ 4 views)\\
\midrule
\multicolumn{2}{l}{1.d) \textbf{Multi-view 3D point trackers} \emph{(with VGGT-$\Omega$~\cite{wang2026vggtomega} depth input)}}\\
\midrule
MVTracker~\cite{rajic2025mvtracker} & MV-Kubric~\cite{rajic2025mvtracker}\\
MVTracker (finetuned)~\cite{rajic2025mvtracker} & MV-Kubric~\cite{rajic2025mvtracker} + \textbf{Perpetua}\\
LAPA~\cite{galoaa2025lookaround} & Calibrated multi-camera splits it derives from Panoptic Studio~\cite{joo2015panoptic,koppula2024tapvid3d} and the PointOdyssey~\cite{zheng2023pointodyssey} robots subset~\cite{galoaa2025lookaround}; frozen DINOv2~\cite{oquab2024dinov2} features and CoTracker~\cite{karaev2024cotracker} 2D observations\\
\midrule
\multicolumn{2}{l}{2.a) \textbf{Monocular 4D reconstruction and point tracking}}\\
\midrule
St4RTrack~\cite{feng2025st4rtrack} & PointOdyssey~\cite{zheng2023pointodyssey}, DynamicReplica~\cite{karaev2023dynamicstereo}, Kubric~\cite{greff2022kubric} (frozen MASt3R~\cite{leroy2024mast3r} encoder trained incl. \textbf{Waymo}~\cite{sun2020waymo})\\
SpatialTrackerV2~\cite{xiao2025spatialtrackerv2} & 17-dataset mixture~\cite{xiao2025spatialtrackerv2} incl. Kubric~\cite{greff2022kubric}, PointOdyssey~\cite{zheng2023pointodyssey}, DynamicReplica~\cite{karaev2023dynamicstereo}, TartanAir~\cite{wang2020tartanair}, BEDLAM~\cite{black2023bedlam}, Ego4D~\cite{grauman2022ego4d}, Stereo4D~\cite{jin2025stereo4d}\\
V-DPM~\cite{sucar2026vdpm} & VGGT~\cite{wang2025vggt} fine-tune on ScanNet++~\cite{yeshwanth2023scannetpp}, BlendedMVS~\cite{yao2020blendedmvs} (static), Kubric-F/G~\cite{greff2022kubric}, PointOdyssey~\cite{zheng2023pointodyssey}, \textbf{Waymo}~\cite{sun2020waymo} (dynamic)\\
Any4D~\cite{karhade2025any4d} & MapAnything init + BlendedMVS~\cite{yao2020blendedmvs}, MegaDepth~\cite{li2018megadepth}, ScanNet++~\cite{yeshwanth2023scannetpp}, VKITTI2~\cite{cabon2020vkitti2}, \textbf{Waymo}~\cite{sun2020waymo,balasingam2024drivetrack} (DriveTrack, 1.5k scenes), Kubric~\cite{greff2022kubric} rendered by CoTracker3~\cite{karaev2024cotracker3} and GCD~\cite{vanhoorick2024gcd}, DynamicReplica~\cite{karaev2023dynamicstereo}, PointOdyssey~\cite{zheng2023pointodyssey}\\
L4P~\cite{badki2026l4p} & VideoMAEv2~\cite{wang2023videomaev2} encoder (1.35M-clip masked-autoencoding pretraining), fine-tuned on Kubric~\cite{greff2022kubric} (15k MOVi-E/F videos), PointOdyssey~\cite{zheng2023pointodyssey}, DynamicReplica~\cite{karaev2023dynamicstereo}, TartanAir~\cite{wang2020tartanair}\\
4RC~\cite{luo2026fourrc} & DA3~\cite{depthanything3} encoder and geometry-decoder init, fine-tuned on PointOdyssey~\cite{zheng2023pointodyssey}, DynamicReplica~\cite{karaev2023dynamicstereo}, Kubric~\cite{greff2022kubric} (MOVi-F and CoTracker3~\cite{karaev2024cotracker3} renders), \textbf{Waymo}~\cite{sun2020waymo}, DL3DV~\cite{ling2024dl3dv}, ScanNet++~\cite{yeshwanth2023scannetpp}, MVS-Synth~\cite{huang2018deepmvs}\\
\midrule
\multicolumn{2}{l}{2.b) \textbf{Multi-view 4D reconstruction and point tracking}}\\
\midrule
OmniX~\cite{jiang2026omnix} & UE5 data engine~\cite{jiang2026omnix} (80k scenes, 1.28M multi-view videos), DynamicReplica~\cite{karaev2023dynamicstereo}, PointOdyssey~\cite{zheng2023pointodyssey}, Spring~\cite{mehl2023spring}, OmniGame~\cite{jiang2026omnix}, HOI4D~\cite{liu2022hoi4d}, \textbf{Waymo}~\cite{sun2020waymo}, DL3DV~\cite{ling2024dl3dv}, Stereo4D~\cite{jin2025stereo4d}\\
\bottomrule
\end{tabular}
\end{table*}

\subsection{Monocular 2D point trackers (1.a)}
\label{sec:monocular_impl}
An adapter runs each monocular model only in its track's query view. The model never sees the other cameras. Its lifted 3D trajectory is still evaluated in all views and in world space, so this limitation directly hurts its non-query-view and world metrics. The benchmark also scores whether a track is visible or occluded. When a monocular model predicts visibility, its query-view prediction is used as its single visibility estimate in every view.

2D trackers return pixel trajectories. The adapter lifts them to 3D by sampling the source depth map at the predicted location in every frame. Samples that fall out of frame or on depth holes take the track's most recent valid depth, or its first valid depth when no earlier one exists. Tracks with no valid depth anywhere remain invalid and score as incorrect.

Each wrapper resizes the video to the resolution its model expects and maps queries and returned tracks between the two resolutions. TAPIR, BootsTAPIR, and TAPNext++ run at $256\times256$, following their official protocol, which is fixed to their $256\times256$ training resolution and puts them at a resolution disadvantage (higher-resolution inference did not improve either model in our tests, and the TAPIR-family ports additionally require square inputs). Track-On runs at $384\times512$. CoWTracker runs at $448\times560$, and MegaFlow at its demo resolution with a fixed width of 518 pixels. PIPs, PIPs++, AllTracker, CoTracker, LocoTrack, TAPTRv3, and DOT run at the benchmark resolution without further resizing, though DOT's internal CoTracker2 seeder keeps DOT's own $384\times512$ resolution. TAPTRv3 additionally requires each image side to be divisible by~32; on subsets whose benchmark resolution is not (Waymo, which is 344 pixels tall), each side is resized to the nearest multiple of~32 and the returned tracks are mapped back to the benchmark resolution.

Sparse trackers that attend across tracks use auxiliary support points, and we follow each method's official protocol for them. CoTracker adds a $10\times10$ support grid, and Track-On its default $20\times20$ grid of first-frame queries. TAPTRv3's official evaluation pads each query with 63 local and 36 global support points. We keep only the 36-point global grid, since the dense benchmark queries already provide the local context its padding emulates. PIPs, PIPs++, TAPIR, BootsTAPIR, TAPNext++, and LocoTrack use no support points in their official protocols. The dense trackers need none, since the dense field is its own support. DOT seeds 8192 dense tracks internally as part of its method. All support tracks are discarded before lifting and evaluation.

Most point trackers natively accept queries at arbitrary timesteps. TAPIR, BootsTAPIR, TAPNext++, CoTracker, LocoTrack, Track-On, and TAPTRv3 therefore need one run per query view. PIPs chains fixed 8-frame windows. Its appearance template is anchored at the window start and re-anchors adaptively. PIPs++ slides variable-length windows over a per-frame track initialization. The dense trackers (AllTracker, DOT, MegaFlow, CoWTracker) only track from a fixed source frame. The adapter therefore runs one dense pass per distinct query timestep and reads out the tracks at the query pixels. Benchmark sequences have a median of 131 distinct query timesteps, so these methods run their tracker over a hundred times per sequence. MegaFlow additionally cannot fit long sequences into memory in one pass, so its video is split into segments of up to 100 frames. The source frame is prepended to each segment, so every segment is still tracked against the same source frame.

\subsection{Monocular 3D point trackers (1.b)}
These trackers run through the same per-query-view adapter but consume the source geometry directly in their native format, and they output 3D trajectories themselves, so no depth lifting is needed. SceneTracker reads camera-space query depth and runs at the benchmark resolution. DELTA predicts per-trajectory depth and runs at $384\times512$. SpatialTracker consumes the source depth as its RGB-D input. TAPIP3D operates directly in the source world frame. SpatialTrackerV2 receives the supplied RGB-D and cameras, and the wrapper restores its internal camera normalization so trajectories return in the source world frame. TAPIP3D adds a support grid of size 40, and SpatialTrackerV2 tracks an internal grid of points alongside the queries for its visual odometry.

\subsection{Multi-view 2D point trackers (1.c): MV-TAP}
\label{sec:mvtap_variants}
MV-TAP tracks a set of query points across a fixed group of camera views. It natively expects each query to be specified in every view of that group. \benchmarkname{} provides only a single query observation per track. The MV-TAP rows in \cref{tab:main_results} differ only in how the participating views and their per-view queries are derived from this observation. The model weights, the inference settings, and the 3D lifting described next are identical across all reported rows.

\paragraph{Lifting per-view 2D tracks to 3D.}
MV-TAP outputs a 2D trajectory and visibility estimate per view. \benchmarkname{} evaluates a single 3D trajectory. Each track and frame is therefore lifted independently. The views with visible, finite, in-bounds predictions form the observation set. Candidate 3D points are generated by DLT triangulation of every observation pair. Additional candidates come from backprojecting each single observation through the source depth map. Every candidate is reprojected into all observed views. Views with reprojection error below 4\,px and positive depth count as inliers. Candidates are ranked by inlier count, then by median and mean inlier error. A triangulated candidate needs at least two inlier views. A depth-backprojected candidate is accepted with a single inlier because the depth map already supplies the ray depth. Frames with no accepted candidate remain invalid and count as incorrect in the 3D metrics. Accepted world-space points are transformed into each camera's coordinate frame to produce the camera-space tracks.

\paragraph{Support points.}
All variants append auxiliary support queries. These give the model cross-view context beyond the sparse evaluation points. At each timestep, a coarse pixel grid in every view is backprojected through the source depth. The resulting points are reprojected into the other views under a stricter 1\% depth-consistency tolerance. The four candidates with the widest cross-view coverage are added as extra queries. Support tracks are discarded before lifting and evaluation.

\paragraph{Plain MV-TAP.}
The benchmark query pixel is backprojected through its view's source depth into a world-space point. This point is then reprojected into every other view with the source cameras. A view receives a query only if the point lands in bounds and in front of the camera. Its depth must also match that view's source depth map within a 5\% relative tolerance. A view is thus used only where the source geometry itself predicts the point to be unoccluded. The query view always participates. Tracks with identical accepted view sets are batched. MV-TAP runs once per view group. This variant uses no ground-truth signal. Errors in the estimated depth or cameras can misplace queries or drop views entirely.

\paragraph{Oracle multi-view queries (\(^{\ddagger}\), MV).}
This variant replaces the estimated view selection with ground truth. A view participates exactly when the track is GT-visible in it at the query timestep. Its query is the GT 3D track point projected with the GT cameras. Tracks are again grouped by identical view sets and tracked per group.

\paragraph{Oracle multi-timestep queries (\(^{\dagger}\), MT).}
All views participate at once. Each view's query is placed at that view's GT-visible timestep nearest to the benchmark query time. Query timesteps may therefore differ across views. This is the most privileged variant. A view is excluded only when the track is never visible in it.

\subsection{Multi-view 3D point trackers (1.d)}
MVTracker consumes all views' RGB-D and cameras jointly and tracks the 3D query point directly in world space. Its scene normalization assumes an upright world with gravity along $-z$, so the scene is first rotated into a gravity-aligned frame estimated from the reconstruction alone: the mean image-down axis of the cameras serves as a prior, refined by the normal of a RANSAC plane fitted to the lowest quarter of the first-frame points when that plane lies within $30^\circ$ of the prior. The rotated scene is then normalized in scale and translation (the floor at the twelfth height percentile, the median camera distance to a fixed radius), and the whole transform is undone on the output. It adds a $10\times10$ support grid at the first timestep of each view.

LAPA~\cite{galoaa2025lookaround} is an end-to-end transformer for multi-camera point tracking. Given calibrated cameras and per-view 2D observations of each query, it populates a normalized volumetric grid from the views, attends over it with cross-view attention carrying geometric priors, and regresses the 3D position of every point with a temporal decoder, in place of triangulation. It is trained on static, calibrated multi-camera splits derived from Panoptic Studio and PointOdyssey (\cref{tab:training_data}). We run the released checkpoint unchanged, with its default CoTracker 2D observations and the cameras of the shared VGGT-$\Omega$ reconstruction. Two adaptations were necessary. First, the released forward pass hands one pose per view to every timestep, so a moving camera cannot be expressed; since every geometric primitive already takes the pose as an argument and the grid is defined in normalized world space, we run the per-frame forward pass with the pose of each frame, which leaves the weights untouched and reproduces the original outputs exactly on static rigs. Second, the volume normalizes pixel distances by a single image size, so views of different resolution (the Aria and exocentric cameras of Ego-Exo4D) are first canonicalized to the reference view's resolution by scaling their intrinsics and observations alike. We do not work around its fixed 16-voxel-per-axis scene grid: the scene bounding box is normalized into it, so on Waymo, where the cameras travel tens of meters, each voxel spans meters and the volume, rather than the camera motion, limits its accuracy.

\subsection{Monocular 4D reconstruction and point tracking (2.a)}
These methods estimate their own geometry instead of consuming a reconstruction source, and the quality of that native reconstruction is reported in \cref{tab:reconstruction_quality}. They run through the same per-query-view adapter as the other monocular trackers. Unlike the lifted monocular trackers, whose shared reconstruction source places every track in a frame common to all cameras, their native geometry has no correspondence to the other benchmark views. Only query-view metrics are therefore defined for this group.

Most of these methods predict no visibility, since they were not trained for occlusion reasoning. They report no visibility estimates and are compared on the localization metrics only. L4P is the exception and predicts per-frame visibility directly.

\paragraph{St4RTrack.}
St4RTrack predicts a dense 3D point map for every frame in the coordinate system of an anchor camera, up to one scale per pass. It runs once per distinct query anchor frame (which makes it very slow to evaluate), and the tracks are read out from the predicted point maps at the query pixels. The benchmark scores camera-space tracks in every frame's own camera, so the model's cameras are recovered with St4RTrack's own solver (the focal length from the anchor frame's point map, per-frame poses by PnP on the reconstruction branch of one pass anchored at the first frame). Each anchor pass is brought to the scale of that pass by the median ratio between the depth its trajectories have in every frame and the depth maps of that pass, and the fitted poses then move the trajectories from the anchor camera into each frame's camera.

\paragraph{SpatialTrackerV2.}
SpatialTrackerV2 jointly estimates video depth, camera intrinsics and poses, and 3D trajectories from the monocular video, tracking an internal point grid alongside the queries for its visual odometry. This is in contrast to its entry in 1.b, which runs on top of the off-the-shelf depth and cameras.

\paragraph{Any4D.}
Any4D predicts per-frame point maps together with camera intrinsics and poses, and the tracks are read out from the point maps at the query pixels. It reconstructs a whole view against a single source frame in one forward pass. Its memory grows with the view's total pixel count, so a view runs in one pass when its frame count times its per-frame pixels stays below $3{\times}10^7$ (about 200 frames at DROID's $288{\times}512$ resolution and 114 frames at $512{\times}512$).

\paragraph{V-DPM.}
V-DPM predicts 3D trajectories in the query camera's coordinate frame, together with per-view depth, intrinsics, and poses. It reconstructs a view in one attention pass over all of its frames. Its memory grows with the frame count alone, and a view of up to 160 frames runs in one pass.

\paragraph{Long-video inference.}
Longer views are split into overlapping windows sized by the same limits (V-DPM uses 128-frame windows) and chained with the construction used for OmniX (\cref{subsec:mv4d}): consecutive windows overlap by half their span, we fit the rotation, translation, and scale that map one window's point maps onto the previous window's from their shared frames, and each track is handed over at a window transition by projecting its current 3D point into the next window. Within a window each model runs exactly as its released inference, and views short enough to fit are reconstructed in a single pass unchanged.

\paragraph{L4P.}
L4P estimates dense depth, camera rays, and query-prompted tracks with per-frame visibility and depth from a video encoder frozen at 16 frames of $224\times224$. It slides this window with stride 8 over the video. All geometry is up-to-scale. We align the sliding windows with the official similarity alignment of depth and camera trajectories, and bring per-track depths onto the dense-depth scale with the official median-ratio correction. The two directional runs (forward in time and backward in time) each estimate their own depth scale, but both predict the query frame. For every track, the backward run's depths are multiplied by the ratio of the two query-frame depth estimates, so the merged track has no depth jump at the query frame.

\paragraph{4RC.}
4RC encodes the whole video once and decodes dense world-space motion maps conditioned on a chosen source frame. For each view that owns queries we encode its video at a 512-pixel long-edge resolution and run one decoder query per distinct query timestamp. Tracks are read out from the motion maps at the query pixels. Depth and cameras come from its own reconstruction heads, and the camera-space track depth equals its depth maps by construction. Tracks live in the query view's coordinate frame.

The monocular variant of OmniX is described together with the multi-view setup in the next subsection.

\subsection{Multi-view 4D reconstruction and point tracking (2.b)}
\label{subsec:mv4d}
These methods consume the RGB of all benchmark views jointly and report tracks that are consistent across views.

OmniX predicts point maps, cameras, and a dense trajectory field for a set of input images in one shared coordinate frame with arbitrary scale. Its demo and its released evaluations run one forward pass per scene, on short clips. Attention cost grows with the square of the image count, and our sequences are far longer, so we run OmniX in overlapping temporal windows and chain the windows. Windows of 48 images scored better than 16, 24, and 32 on four datasets, because fewer windows leave fewer chaining steps. A 48-image forward pass peaks at about 55~GB of GPU memory.

\paragraph{Window composition.}
Each window is an independent reconstruction with its own arbitrary scale. Frames from a single static camera give a window no parallax, which leaves that scale poorly constrained. A window therefore holds up to 40 consecutive frames of one view plus 4 frames from each of two other views, sampled evenly across the same time span; the added camera baseline constrains the scale. On the highest-resolution multi-view subsets, where a $48$-image forward exceeds memory, we shrink the window (to as few as 16 dense frames on Harmony4D) so that one forward fits.

\paragraph{Chaining windows.}
Windows advance by half their span, so consecutive windows share half their frames. From eight of the shared frames we fit the rotation, translation, and scale that map one window's point maps onto the previous window's, with a trimmed least-squares fit, and express every window in the frame of the first one. The overlap is what makes this alignment work, and a one-frame overlap scored far worse for both variants. The number of shared frames that enter the fit does not matter, and one, eight, and twenty all scored the same.

\paragraph{Handing tracks over.}
Tracks are handed over at window transitions. The track's current 3D point is projected into the first image of the next window. Among the pixels in a $9\times9$ neighbourhood of the projection we pick the one whose 3D point is closest to the track's 3D point, and the track continues along that pixel's trajectory field. At object boundaries the surrounding pixels cover both the object and the background behind it, and this choice prevents the track from jumping onto the background surface. When the projection falls outside the image, the track keeps its last 3D position and retries the projection at later windows.

\paragraph{Merging the views.}
The per-view results are then merged. One additional forward pass contains a few frames of every view across the whole time span and reconstructs all views in one frame. For each view we fit the same rotation, translation, and scale mapping between its windowed point maps and this pass's point maps of the same frames, and apply it to the view's cameras, depth, and tracks. We report the model's own cameras. OmniX solves all cameras of a window in one pass, so they are mutually consistent, and replacing them with per-frame cameras fitted from the point maps scored worse on every metric. OmniX reports its point maps to the benchmark directly, so \cref{tab:reconstruction_quality} scores those points and not points unprojected from its depth.

\paragraph{Resolution and time encoding.}
We resize each video so that its long side is 504 pixels, the long side OmniX was trained on, and whose short side is the source aspect rounded to a multiple of the patch size. Each axis is resized independently, so no image content is cropped away and no query point is lost. Portrait sequences are rotated to landscape first, since the model is trained on landscape video only, and every output is rotated back. Each frame carries its true frame offset in the temporal encoding, so the strided support frames share the clock of the dense span they sit in. OmniX's released inference code numbers the frames of every input video consecutively, which coincides with their true times only when all videos in a window are dense, as in its own inputs; for our strided support frames it would place them at the wrong times. The true offsets scored better on all five datasets we tested. OmniX encodes time as a video index plus a frame index, summed into the tokens. Frames of different videos that carry the same frame index therefore share a clock, and OmniX's released code uses this same encoding for synchronized videos. The model also takes a separate global time index, but that index only selects the slot at which each image is normalized. It never enters the features, the released checkpoint holds no weights for it, and declaring frames simultaneous through it left the output unchanged, bit for bit.

\paragraph{Outputs.}
The trajectory field is predicted per image, so a query point's positions come from the field of the view that owns it, and the support frames of the other views enter only through attention. OmniX predicts no visibility, so it reports none and is compared on the localization metrics only. The monocular 2.a variant uses no cross-view frames and no merging pass. Its windows hold 48 frames of the query view, each view keeps its own coordinate frame, and only query-view metrics are defined, as for the rest of group 2.a.

\section{Dataset Statistics}
\label{sec:dataset_stats}
\Cref{tab:subset_preview} summarizes the seven \benchmarkname{} splits of \cref{sec:tapvidmv}: their domains, real or synthetic provenance, camera configurations, depth sources, and sizes. \Cref{tab:dataset_stats} reports per-dataset size statistics for the TAP benchmarks compared in \cref{tab:related_work_comparison}.

\begin{table}[th]
\centering
\caption{Per-dataset size statistics for TAP benchmarks in \cref{tab:related_work_comparison}. Scenes are sequences, videos are camera streams (sequences $\times$ cameras), and tracks are annotated point trajectories; \emph{dense} marks datasets that provide correspondence for every surface point rather than a fixed track set.}
\label{tab:dataset_stats}
\setlength{\tabcolsep}{3pt}
\renewcommand{\arraystretch}{1.2}
\footnotesize
\begin{tabular}{l r r r}
\toprule
\textbf{Benchmark}
&
\textbf{Scenes}
&
\textbf{Videos}
&
\textbf{Tracks}
\\
\midrule

Kubric~\cite{greff2022kubric} (TAP-Vid)
& 39,124 & 39,124 & dense$^{\dagger}$ \\

PointOdyssey~\cite{zheng2023pointodyssey}
& 159 & 159 & $\sim$3.2M$^{\S}$ \\

SynthVerse~\cite{zhao2026synthverse}
& 48K & $\geq$48K$^{\|}$ & dense$^{\dagger}$ \\

Syn4D~\cite{jiang2026syn4d}
& 4.7K & 37.6K & dense$^{\dagger}$ \\

\hdashline

TAPVid-DAVIS~\cite{doersch2022tapvid}
& 30 & 30 & 650 \\

EgoPoints~\cite{egopoints}
& 517 & 517 & 4,703 \\

RoboTAP~\cite{vecerik2024robotap}
& 265 & 265 & 11,592 \\

\hdashline

TAPVid-3D~\cite{koppula2024tapvid3d}
& 255 & 4,569 & 2.1M \\

DriveTrack~\cite{balasingam2024drivetrack}
& 1,000 & $\sim$10,000 & 1B \\

\hdashline

DexYCB~\cite{chao2021dexycb}
& 10$^{\ddagger}$ & 80 & 5,120 \\

Panoptic Studio~\cite{joo2015panoptic}
& 6$^{\ddagger}$ & 162 & 3,072 \\

Stereo4D~\cite{jin2025stereo4d}
& $\sim$110K & $\sim$220K & dense$^{\dagger}$ \\

PACE~\cite{you2024pace}
& 100 & 300 & --$^{\P}$ \\

\midrule

\rowcolor{oursrow}
\textbf{\benchmarkname}
& 238 & 1,004 & 74,147 \\

\quad\(\triangleright\) DROID\(^{\ast}\)
& 48 & 144 & 4,181 \\

\quad\(\triangleright\) EgoExo4D
& 18 & 54 & 8,940 \\

\quad\(\triangleright\) Harmony4D
& 29 & 116 & 3,104 \\

\quad\(\triangleright\) PACE
& 24 & 72 & 1,819 \\

\quad\(\triangleright\) Hi4D
& 48 & 384 & 10,456 \\

\quad\(\triangleright\) Waymo
& 50 & 150 & 24,143 \\

\quad\(\triangleright\) Perpetua
& 21 & 84 & 21,504 \\

\bottomrule
\multicolumn{4}{@{}p{\linewidth}@{}}{\footnotesize \(^{\ast}\)For DROID, we also release the full pool of 5,371 scenes (16,113 videos, $\sim$3M+ tracks) for training.}\\
\multicolumn{4}{@{}p{\linewidth}@{}}{\footnotesize $^{\dagger}$Any surface point can be queried (TAP-Vid samples 256 per Kubric video).}\\
\multicolumn{4}{@{}p{\linewidth}@{}}{\footnotesize $^{\S}$About 20K trajectories per video.}\\
\multicolumn{4}{@{}p{\linewidth}@{}}{\footnotesize $^{\ddagger}$The multi-view evaluation split of MVTracker~\cite{rajic2025mvtracker}, also used by MV-TAP~\cite{koo2025mvtap}, with 512 query points per scene; the source datasets are larger (DexYCB: 1,000 sequences).}\\
\multicolumn{4}{@{}p{\linewidth}@{}}{\footnotesize $^{\P}$PACE provides 258K 6-DoF pose annotations but no point tracks; our split derives tracks from them.}\\
\multicolumn{4}{@{}p{\linewidth}@{}}{\footnotesize $^{\|}$Instance-level sequences are rendered from five cameras (four fixed and one orbiting); scene-level camera setups vary and no stream count is given.}
\end{tabular}
\end{table}

\section{Per-Sequence Track Visualizations}
\label{sec:per_sequence_visuals}
Every sequence of \benchmarkname{} was manually inspected before release (\cref{sec:tapvidmv}). For each sequence, we rendered the ground-truth 3D tracks projected into every camera view, side by side and synchronized in time, together with the per-view visibility flags, and a human reviewed the full video, removing the tracks that the annotation pipeline got wrong and correcting the trajectories where the source signal (mesh fit, object pose, LiDAR box, or kinematic model) had drifted from the observed surface. Visualizations (for anonymous and blind review) of each sequence are released with the benchmark at \url{https://storage.googleapis.com/dm-tapnet/tapvidmv/rrd/index.html} as Rerun files. The reader can download and view in 3D, to verify any annotation themselves and see the camera motion, occlusions, and track density of every sequence rather than the samples of \cref{fig:mvtap_splits}.

\section{DROID Generation Pipeline and Dataset Details}
\label{sec:supp_droid}

In this section, we describe the full technical formulation, stage-by-stage methodology, quality metrics, and dataset statistics for the DROID 3D point trajectory generation pipeline.

\subsection{Setup and Challenges}
The DROID dataset~\cite{khazatsky2024droid} captures diverse tabletop robotic manipulation tasks using synchronized ZED 2 stereo camera rigs (typically two static exocentric cameras and one wrist-mounted camera), accompanied by 7-DOF Franka Emika Panda robot joint telemetries and Robotiq 2F-85 gripper states. The stereo cameras have factory-calibrated baselines ($63$\,mm for the wrist camera, $120$\,mm for the external cameras). Generating metric, multi-view consistent 3D point tracks from raw DROID episodes poses three primary challenges:
\begin{enumerate}
    \item \textbf{Close-range stereo matching degradation on specular surfaces:} At close distance ($<15$\,cm), the large disparity exceeds typical stereo matching search bounds. Furthermore, specular reflections on the metallic robot gripper cause severe depth holes and outliers on the wrist view.
    \item \textbf{Camera extrinsic calibration drift:} Multi-camera extrinsics calibrated per-setup can drift. Optimizing each camera's pose independently against robot CAD depth does not guarantee cross-view geometric consistency of the surrounding environment point clouds.
    \item \textbf{Articulated vs.\ static scene structure:} The tabletop environment is largely stationary, while the robot undergoes fast articulated motion. Appearance-based 2D trackers drift on specular and self-similar robot links, while introducing unnecessary temporal noise on static background surfaces.
\end{enumerate}

\subsection{Stage 1: Metric Depth with Gripper Distillation}
\paragraph{Stereo Depth Estimation.}
Stereo pairs are processed using the S2M2 stereo matching network~\cite{s2m2} with aggressive confidence thresholding ($\tau = 0.95$). Pixels with confidence below $\tau$ are zeroed out, prioritizing high-confidence depth measurements over density.

\paragraph{Temporal Gripper Depth Distillation.}
On the wrist camera, stereo matching fails on the robot gripper. We leverage a domain prior: at the beginning of each episode, the gripper remains in an open, unarticulated configuration, forming a rigid surface with time-invariant local depth. We distill clean gripper geometry via:
\begin{enumerate}
    \item \textbf{Consensus Segmentation:} Zero-shot SAM~\cite{sam} generates candidate gripper masks. A temporal majority vote across initial open-gripper frames ($g < 0.05$) produces a stabilized consensus mask $M_{\text{grip}}$.
    \item \textbf{Temporal Median Distillation:} Within $M_{\text{grip}}$, we compute the per-pixel temporal median depth across initial open frames:
    \begin{equation}
        D_{\text{distill}}(u, v) = \operatorname{median}_{t \in \mathcal{T}_{\text{open}}} D_{\text{raw}}(t, u, v).
    \end{equation}
    The median suppresses non-Gaussian specular matching noise and dropouts.
    \item \textbf{Depth Injection:} For all open-gripper frames, the distilled depth replaces raw stereo depth within $M_{\text{grip}}$, providing clean geometry for downstream extrinsic calibration and 3D point sampling.
\end{enumerate}

\subsection{Stage 2: Differentiable Multi-Camera Extrinsics Calibration}
Camera extrinsics are parameterized as 6-DOF Lie algebra transforms $\mathbf{\xi} \in \mathfrak{se}(3)$ mapping camera coordinates to the robot base frame:

\paragraph{Camera-to-Robot CAD Alignment.}
Each camera's pose $T_c \in \mathrm{SE}(3)$ is initialized from dataset metadata and refined against the Franka Panda URDF surface model. Given joint angles $\mathbf{q}(t)$, forward kinematics places CAD surface points $\mathbf{X}_{\text{CAD}}$ in the base frame. Surface points are projected onto camera $c$ with intrinsics $\mathbf{K}_c$:
\begin{equation}
    \mathcal{L}_{\text{robot}}^{(c)} = \frac{1}{|\mathcal{V}_c|} \sum_{(t, i) \in \mathcal{V}_c} \left| Z_{\text{proj}}(t, i) - D_{\text{obs}}(t, u_i, v_i) \right|,
\end{equation}
where $\mathcal{V}_c$ denotes valid visible front-facing CAD points (culling back-facing normals with $\mathbf{n}_i \cdot \mathbf{d}_i \ge 0$ and depth tolerance $|Z_{\text{proj}} - D_{\text{obs}}| < 15$\,cm). For the wrist camera, the optimization operates in end-effector coordinates to solve for the hand--eye transform $T_{\text{cam}\to\text{ee}}$.

\paragraph{Global Joint Multi-View Optimization.}
To guarantee that background environment surfaces overlap across different viewpoints, we jointly optimize all camera extrinsics simultaneously using a composite loss:
\begin{equation}
    \mathcal{L}_{\text{total}} = \mathcal{L}_{\text{chamfer}} + \lambda_{\text{robot}} \sum\nolimits_{c} \mathcal{L}_{\text{robot}}^{(c)},
\end{equation}
where $\mathcal{L}_{\text{chamfer}}$ is the pairwise truncated Chamfer distance between unprojected environment point clouds $\mathcal{P}_i, \mathcal{P}_j$ from cameras $i$ and $j$:
\begin{equation}
\begin{aligned}
    \mathcal{L}_{\text{chamfer}} &= \sum_{i < j} \big( d_\delta(\mathcal{P}_i, \mathcal{P}_j) + d_\delta(\mathcal{P}_j, \mathcal{P}_i) \big), \\
    d_\delta(\mathcal{P}, \mathcal{Q}) &= \frac{1}{|\mathcal{P}|} \sum_{\mathbf{p} \in \mathcal{P}} \; \min_{\substack{\mathbf{q} \in \mathcal{Q} \\ \|\mathbf{p}-\mathbf{q}\| < \delta}} \|\mathbf{p}-\mathbf{q}\|,
\end{aligned}
\end{equation}
where $d_\delta$ is the one-sided truncated distance, with truncation distance $\delta = 5$\,cm to prevent non-overlapping fields of view from dominating gradients.

\begin{figure*}[t]
  \centering
  \begin{subfigure}[b]{0.495\linewidth}
    \centering
    \includegraphics[width=\linewidth]{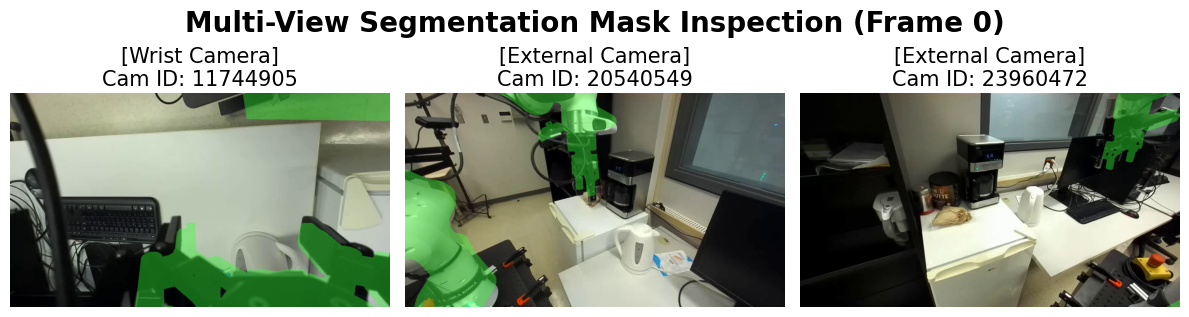}
    \caption{Multi-view mask inspection (\textbf{Before} calibration)}
    \label{fig:droid_cam_before}
  \end{subfigure}
  \hfill
  \begin{subfigure}[b]{0.495\linewidth}
    \centering
    \includegraphics[width=\linewidth]{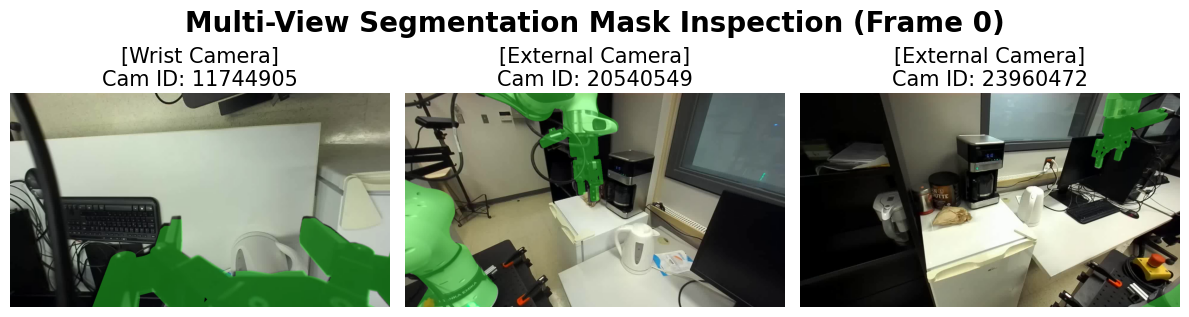}
    \caption{Multi-view mask inspection (\textbf{After} calibration)}
    \label{fig:droid_cam_after}
  \end{subfigure}

  \vspace{2mm}

  \begin{subfigure}[b]{0.495\linewidth}
    \centering
    \includegraphics[height=0.8\linewidth, width=\linewidth]{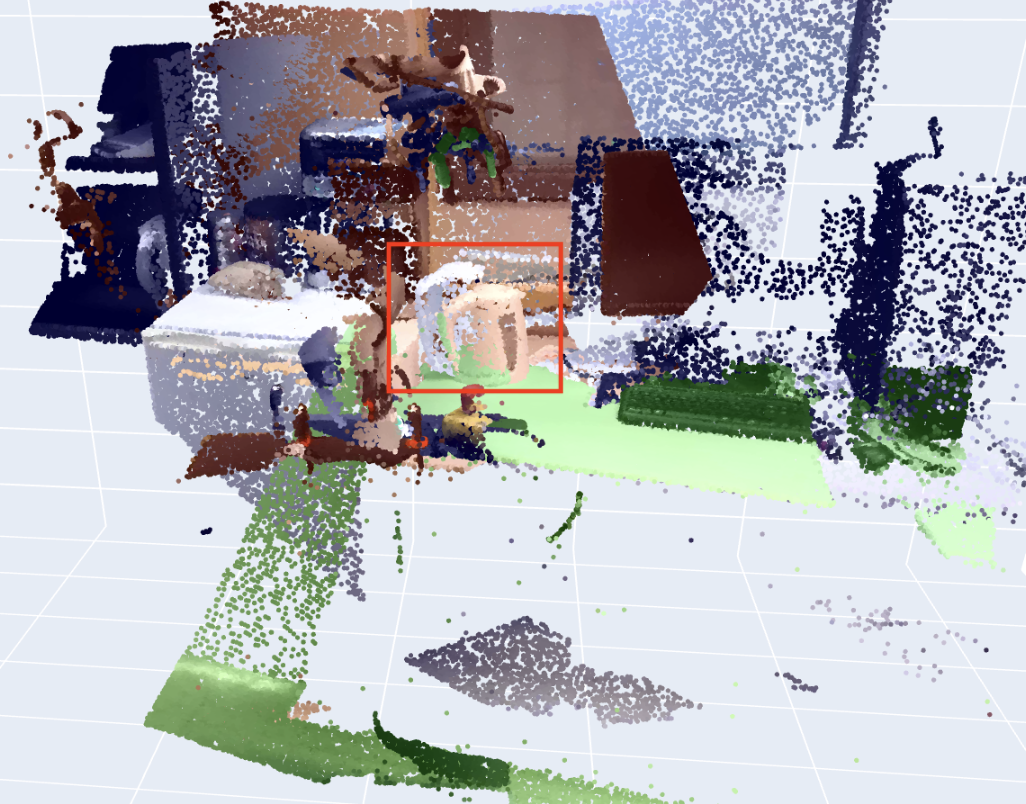}
    \caption{Unprojected scene point cloud (\textbf{Before} calibration)}
    \label{fig:droid_pc_before}
  \end{subfigure}
  \hfill
  \begin{subfigure}[b]{0.495\linewidth}
    \centering
    \includegraphics[height=0.8\linewidth, width=\linewidth]{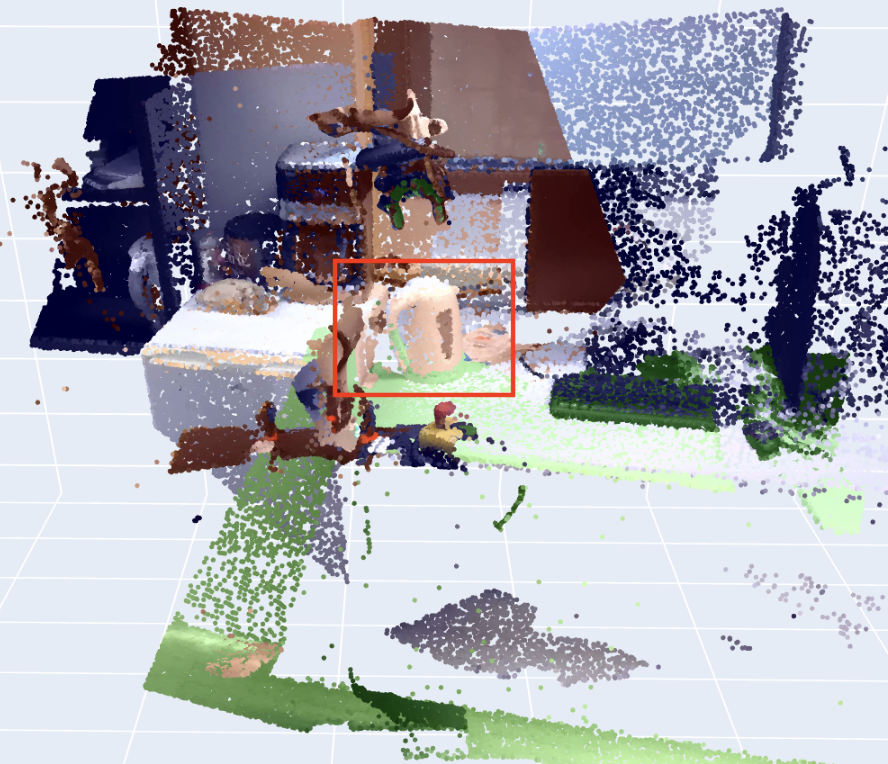}
    \caption{Unprojected scene point cloud (\textbf{After} calibration)}
    \label{fig:droid_pc_after}
  \end{subfigure}

  \caption{\textbf{Qualitative comparison of multi-camera extrinsic calibration on DROID.} Top row: Projected robot segmentation masks across the wrist camera and two static external views before (a) and after (b) calibration. Bottom row: Fused 3D environment point clouds unprojected from all views. Before joint optimization (c), calibration drift causes severe ghosting and duplicated surfaces (highlighted in the red box around the tabletop kettle). Our differentiable joint optimization (d) resolves cross-view discrepancies, yielding a sharp, geometrically consistent static background representation.}
  \label{fig:droid_calibration_comparison}
\end{figure*}

As qualitatively illustrated in Fig.~\ref{fig:droid_calibration_comparison}, optimizing camera extrinsics jointly with the multi-view Chamfer loss successfully eliminates severe ghosting and surface duplication in the unprojected environment point clouds (highlighted in the red box), while ensuring accurate projection consistency across both the wrist-mounted and static external cameras.

\subsection{Stage 3: Dual-Track 3D Tracking Architecture}
DROID provides tracks only for the robot arm, gripper, and static environment, excluding other moving objects such as objects being manipulated.
\paragraph{Track A: Static Background via Cross-View Depth Consensus.}
In tabletop manipulation, the workspace background is stationary. We exploit this prior:
\begin{enumerate}
    \item \textbf{Dense Initialization at $t=0$:} All non-robot pixels at frame 0 are unprojected into 3D world points $\mathbf{X} \in \mathbb{R}^3$.
    \item \textbf{Cross-View Depth Verification:} A 3D candidate $\mathbf{X}$ is projected into all other camera views. If its projected depth agrees with the observed sensor depth in at least one other camera within 5\,mm, it is verified as physically consistent.
    \item \textbf{Multi-Frame Static Verification:} Candidate points are verified across 5 evenly spaced keyframes. Points whose projected depth matches observed depth across multiple keyframes are retained, filtering out transient dynamic objects.
    \item \textbf{Voxel Deduplication and Projection:} Verified points are deduplicated via voxel hashing and subsampled to $\sim$300 static points per sequence. 2D trajectories across all views are obtained by geometric projection $\pi(\mathbf{X}, \mathbf{K}_c, T_c(t))$, with visibility governed by image bounds, sensor depth consistency, and robot occlusion masks.
\end{enumerate}

\paragraph{Track B: Articulated Robot Tracking via Forward Kinematics.}
Because tracking reflective, self-similar robot links with 2D appearance trackers is ill-posed, we assign robot seed points at $t=0$ to their nearest URDF CAD link. Point positions over time $t$ are computed directly via forward kinematics:
\begin{equation}
    \mathbf{X}_{\text{robot}}(t) = T_{\text{base}}(t) \cdot T_{\text{link}_k}(\mathbf{q}(t)) \cdot \mathbf{x}_{\text{local}},
\end{equation}
providing exact 3D trajectories across all 7 arm links and gripper fingers. Visibility is determined by checking against rendered robot CAD depth (self-occlusion) and raw sensor depth (environment occlusion).

\subsection{Quality Validation and Dataset Statistics}
\paragraph{Physical Depth Residual as Self-Consistency Metric.}
Because 2D tracks in our static-prior pipeline are directly projected from 3D points, 2D reprojection error against stored tracks is an identity check ($\sim$0\,px). Instead, we evaluate physical quality via \textbf{Depth Residual}: comparing the projected 3D track depth against raw stereo sensor depth across all visible frames:
\begin{equation}
    \Delta Z(t, p, c) = \left| Z_{\text{proj}}(t, p, c) - D_{\text{sensor}}(t, u_{p,c}, v_{p,c}) \right|.
\end{equation}
This metric assesses physical agreement with raw sensor measurements. Static background points exhibit low depth residual (reflecting sensor noise and sub-centimetre extrinsics precision), while robot tracks reflect forward kinematics precision and fast dynamic motion.

\paragraph{Full Dataset and Reviewed Evaluation Split.}
Our pipeline successfully processes \textbf{5,371 episodes} (96.3\% completion rate across 5,580 attempted episodes), yielding over 16,000 calibrated video streams and millions of 3D trajectories. From this full pool, we initially select \textbf{50 evaluation sequences} by applying quality thresholds ($\text{Chamfer} < 0.05$, $\text{Depth Residual} < 20$\,mm) and stratified sampling across diverse recording sites and interaction complexity. Before manual filtering, each generated sequence provides 3 synchronized camera views, $\sim$300 static background tracks, $\sim$300 dynamic robot tracks, and full 3D calibration.

\paragraph{Manual Filtering.}
We conduct two manual filtering passes and exclude 2 of the 50 sequences for calibration or multi-view consistency issues. Starting from 120 candidate tracks per retained sequence, the first pass accepts 5,215 of 5,760 candidates across the remaining 48 sequences (90.5\%), and the second accepts 4,181 of those 5,215 tracks (80.2\%), leaving \textbf{4,181 tracks across 48 sequences}, or \textbf{72.6\%} of the original candidates in the retained sequences.

\section{Dataset Generation: Additional Details}
\label{sec:supp_datasets}

This section provides additional technical details for dataset construction, verification, and cleaning deferred from the main text. Details for the DROID subset are provided independently in \cref{sec:supp_droid}.

\subsection{Harmony4D Track Verification and Cleaning}
\label{sec:supp_datasets_harmony4d}
As described in \cref{sec:tapvidmv:harmony4d}, the fitted meshes do not always remain perfectly aligned with the observed body or clothing, particularly around highly articulated parts such as the hands and feet. We therefore verify the mesh tracks against image-based tracking.

\begin{figure}[tb]
  \centering
  \includegraphics[width=\linewidth]{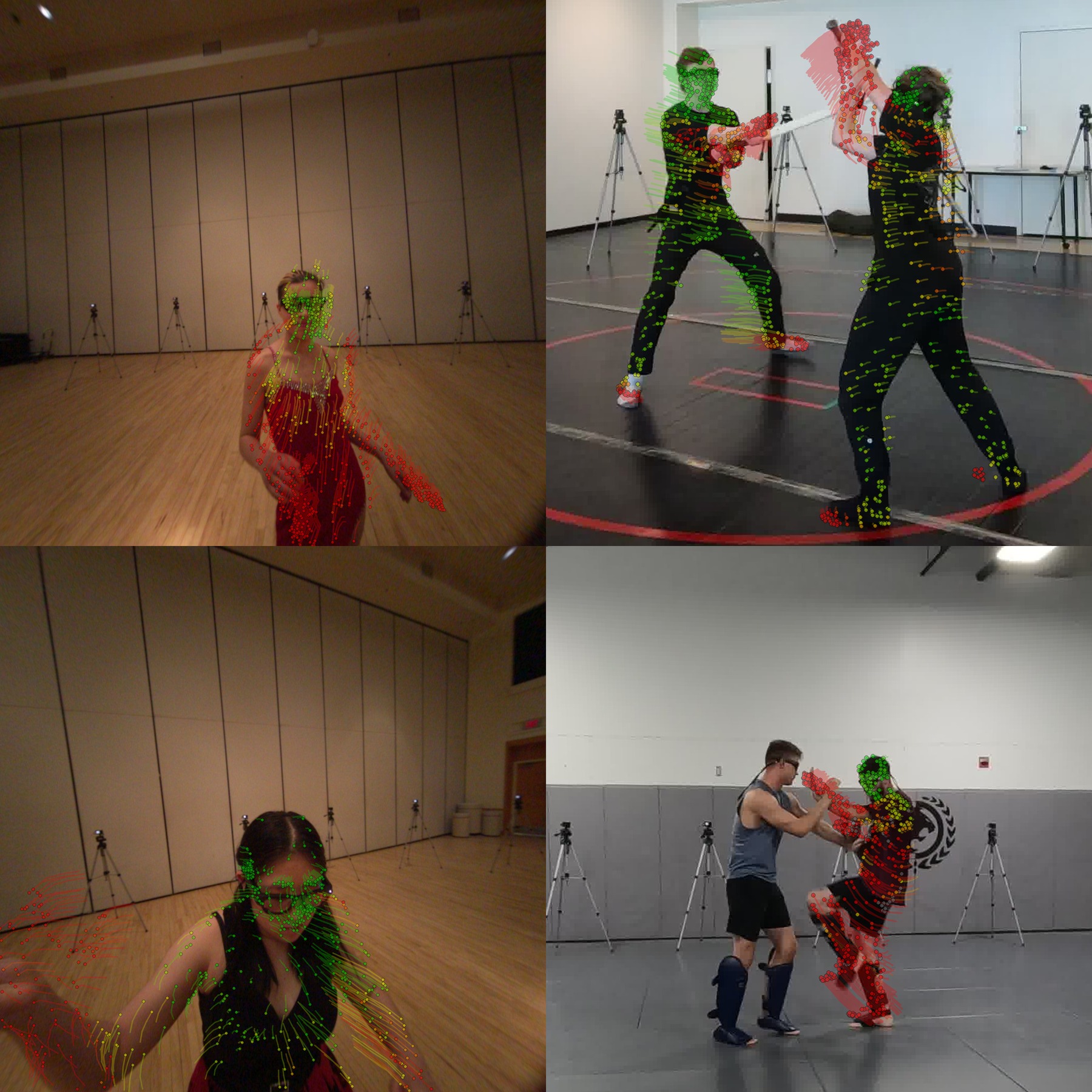}
  \caption{%
    \textbf{Track cleaning examples.} Green tracks agree with our robust 2D tracker consensus and pass verification, while red tracks disagree and are removed. Rejected tracks occur mainly on extremities such as hands and feet, or where the fitted SMPL mesh is misaligned with the observed surface, often floating slightly off the person rather than lying on the visible body or clothing.
  }
  \label{fig:harmony4d_track_cleaning}
\end{figure}

\paragraph{Track verification.}
We seed TAPNext++~\cite{jung2026tapnextpp} at the projected mesh locations and track them forward over short overlapping temporal windows. Each window is also tracked backward to measure cycle consistency~\cite{wang2019cycletime}. The overlapping predictions are combined using a robust geometric-median consensus, which downweights windows that drift onto other image content. We compare this consensus with the projected mesh trajectory, converting the image-space discrepancy $e_{\mathrm{px}}$ to an approximate metric error $e_{\mathrm{m}}=e_{\mathrm{px}}z/f$ using the vertex depth $z$ and focal length $f$. Tracks that remain consistent over the sequence are retained, while tracks with persistent disagreement are removed. In sword-sparring sequences, we additionally segment the sword with SAM~3~\cite{sam3} and mark a point as occluded when its projection falls inside the sword mask. \Cref{fig:harmony4d_track_cleaning} illustrates examples of tracks removed by this pipeline.

\subsection{PACE Coordinate Recovery and Track Cleaning}
\label{sec:supp_datasets_pace}

\paragraph{World frame recovery.}
PACE is recorded with a moving handheld camera rig, but provides only rig-relative camera extrinsics and per-frame 6-DoF object poses in camera coordinates, rather than camera poses in a fixed world frame. Recovering camera motion from the object annotations requires distinguishing changes caused by camera motion from those caused by object motion. A stationary object provides a fixed reference because changes in its pose relative to the camera can then be attributed to the camera's motion. We therefore evaluate every annotated object as a potential stationary reference, comparing the camera motion implied by this assumption with background motion estimated from SIFT feature matches and sensor depth. We select the best-supported candidate and derive camera extrinsics solely from its annotated poses, assuming it remains stationary throughout the sequence. We retain the fixed relative positions and orientations between cameras and define the world frame as camera 0's coordinate frame at frame 0. Because camera extrinsics are derived from the reference object's annotated poses, annotation noise transfers directly to the recovered camera motion, producing world-frame jitter even when the object is physically stationary. To identify sequences where this jitter is pronounced, we backproject held-out background correspondences using sensor depth and measure their frame-to-frame displacement in the recovered world frame. This background-shift diagnostic guides manual 3D review and the removal of affected sequences.

\paragraph{Track filtering and visibility estimation.}
PACE tracks are initialized from sensor depth and propagated across frames using the annotated object poses. Visibility is determined by comparing each track's camera-space depth with the sensor depth at its projected pixel. Before manual review, we filter invalid tracks, fill missing depth for visibility estimation, and flag potentially off-surface tracks. First, a subset of tracks collapse onto a camera's optical centre rather than onto a scene surface. Projecting such a point involves division by zero or near-zero camera-space depth, producing invalid query pixels, and many of the collapsed tracks are byte-identical. We compute each camera's optical centre from its recovered world-frame extrinsics and measure each track's minimum distance to any camera centre across all frames. We discard tracks whose minimum distance is below 1\,mm, together with tracks containing non-finite coordinates or query pixels. The precise cutoff is not critical because collapsed points lie approximately $10^{-16}$\,m from a camera centre, whereas our audit of all 69 PACE sequences found the nearest retained point approximately 338\,mm away. This filter removes 1{,}456 of 20{,}328 PACE tracks (7.2\%).

Second, we determine visibility where sensor depth is missing by filling depth holes with Depth Anything 3~\cite{depthanything3} predictions. To align the predictions with sensor depth, we first rescale them using the supplied camera poses via Umeyama alignment, then fit a per-camera, per-frame scale and offset to the measured depth, excluding outliers. We release the resulting visibility labels alongside the original sensor depth maps. Third, we flag potentially off-surface tracks with two separate checks. We compare each visible-labelled track's camera-space depth, derived from the annotated object poses, with the sensor depth at its projected pixel, flagging points more than 5\,cm in front of the measured surface. Box-prompted SAM2.1~\cite{sam2} masks also flag tracks whose projections fall outside their object's mask too often. Finally, we manually review every sequence and its tracks across all three views, using these flags to decide which tracks or sequences to remove.

\paragraph{Balancing object motion.}
Most PACE tracks lie on objects that stay in place, so after cleaning only 10.9\% of the tracks are dynamic, counting a track as dynamic when its world-space position moves by more than 5\,cm within the sequence. We therefore keep every dynamic track of a sequence together with an equal number of static tracks drawn uniformly at random, which leaves 564 dynamic and 554 static tracks in the 14 sequences with object motion. The remaining ten sequences contain no dynamic track at all (\texttt{scene00\_video00}, \texttt{scene00\_video02}, \texttt{scene00\_video03}, \texttt{scene00\_video04}, \texttt{scene00\_video05}, \texttt{scene00\_video06}, \texttt{scene00\_video08}, \texttt{scene00\_video09}, \texttt{scene01\_video00}, and \texttt{scene01\_video04}), and we keep 72 static tracks in each of them (all 53 in \texttt{scene01\_video04}). The split then holds 1,819 tracks, of which 31.0\% are dynamic.

\subsection{Hi4D Track Construction and Actions}
\label{sec:supp_datasets_hi4d}
We construct the Hi4D tracks entirely from files in the public release. The released scan meshes are reconstructed independently at each frame and carry no vertex correspondence over time, so correspondence comes from the fitted SMPL bodies instead. On the first frame, we sample 4,000 points area-uniformly over the two fitted SMPL bodies. Each sample retains its person, triangle, and barycentric coordinates, which propagate a body-surface anchor and geometric triangle normal through the per-frame fits. To place the track on clothing and hair rather than inside the fitted body, we cast from the anchor in both normal directions and keep the nearest public-scan intersection within 5\,cm. We reject an intersection if the ray reaches either fitted body before the scan or if the candidate lies closer to the other person. Frames without a valid intersection retain the SMPL anchor only as a placeholder and are forced invisible.

We discard candidates with valid scan intersections in at most 10\% of frames or visibility in at least one camera for less than 20\% of frames, then select 600 spatially diverse tracks for review. Visibility is recomputed by camera-to-track ray casting against the public scan mesh with a 1\,mm surface tolerance and strict image bounds. We observed occasional frame-to-frame visibility flicker near silhouettes. A single annotator reviews the candidates in a fixed random order in an interactive 3D viewer, which shows the candidate trajectory together with the per-frame scan geometry it should adhere to. The annotator rejects a candidate that drifts along the surface, jumps between surfaces, or detaches from the body it started on, targeting a mean surface deviation below 2\,cm and a maximum drift below 5\,cm. Review stops once at least 200 candidates have been reviewed and at least 100 have been accepted, so most of the 600 are never reached. Four additional sequences are excluded because their released SMPL fits are too noisy or temporally unstable to support reliable tracks. This leaves 5,228 human tracks over 48 sequences, with 100--180 per sequence.

The released sequences cover tai chi, basketball, back-hugging, bending, arguing, high-fives, dancing, posing, talking, kissing, jumping, piggybacking, football, fighting, hugging, games, leg interactions, cheering, side-hugging, fever acting, handshakes, and related close-contact motions.

\begin{figure*}[t]
  \centering
  \setlength{\tabcolsep}{2pt}
  \begin{tabular}{*{8}{c}}
    \begin{tabular}[t]{@{}c@{}}\includegraphics[width=0.112\textwidth]{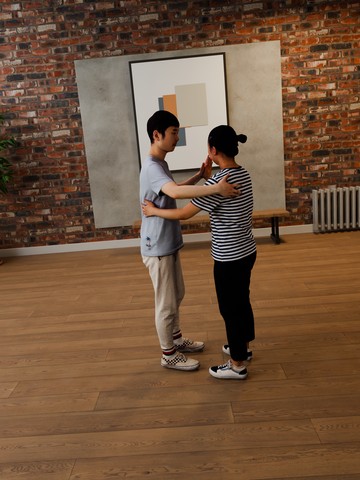}\\[-2pt]\footnotesize pair00\_\\[-3.5pt]\footnotesize dance00\end{tabular} & \begin{tabular}[t]{@{}c@{}}\includegraphics[width=0.112\textwidth]{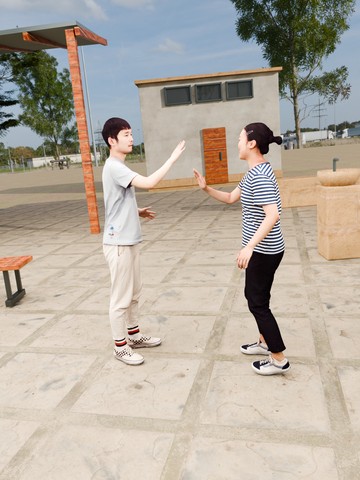}\\[-2pt]\footnotesize pair00\_\\[-3.5pt]\footnotesize highfive00\end{tabular} & \begin{tabular}[t]{@{}c@{}}\includegraphics[width=0.112\textwidth]{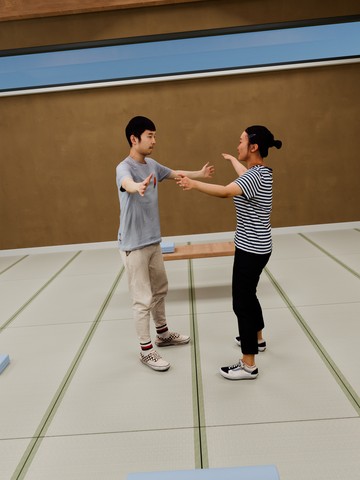}\\[-2pt]\footnotesize pair00\_\\[-3.5pt]\footnotesize taichi00\end{tabular} & \begin{tabular}[t]{@{}c@{}}\includegraphics[width=0.112\textwidth]{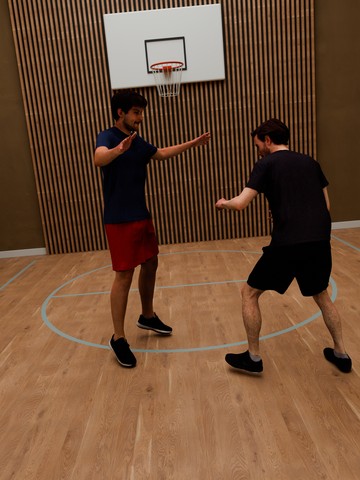}\\[-2pt]\footnotesize pair01\_\\[-3.5pt]\footnotesize basketball01\end{tabular} & \begin{tabular}[t]{@{}c@{}}\includegraphics[width=0.112\textwidth]{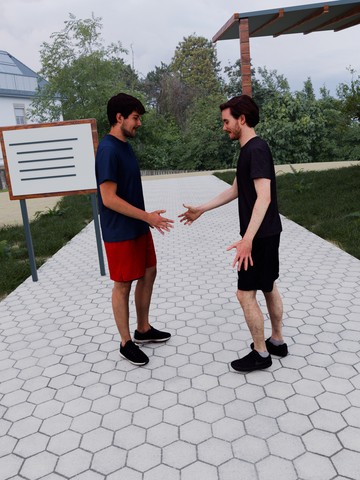}\\[-2pt]\footnotesize pair01\_\\[-3.5pt]\footnotesize highfive01\end{tabular} & \begin{tabular}[t]{@{}c@{}}\includegraphics[width=0.112\textwidth]{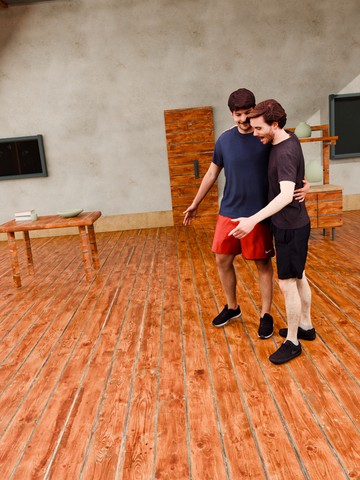}\\[-2pt]\footnotesize pair01\_\\[-3.5pt]\footnotesize talk01\end{tabular} & \begin{tabular}[t]{@{}c@{}}\includegraphics[width=0.112\textwidth]{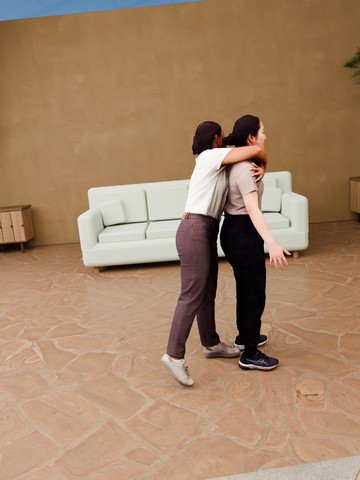}\\[-2pt]\footnotesize pair02\_\\[-3.5pt]\footnotesize backhug02\end{tabular} & \begin{tabular}[t]{@{}c@{}}\includegraphics[width=0.112\textwidth]{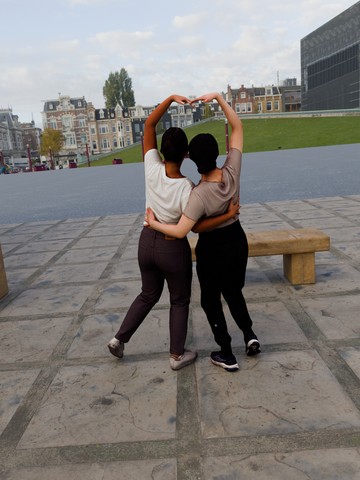}\\[-2pt]\footnotesize pair02\_\\[-3.5pt]\footnotesize pose02\end{tabular}\\[3pt]
    \begin{tabular}[t]{@{}c@{}}\includegraphics[width=0.112\textwidth]{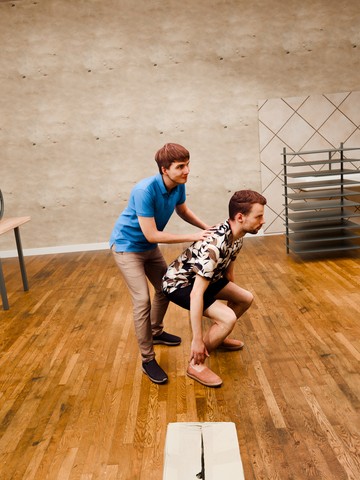}\\[-2pt]\footnotesize pair09\_\\[-3.5pt]\footnotesize bend09\end{tabular} & \begin{tabular}[t]{@{}c@{}}\includegraphics[width=0.112\textwidth]{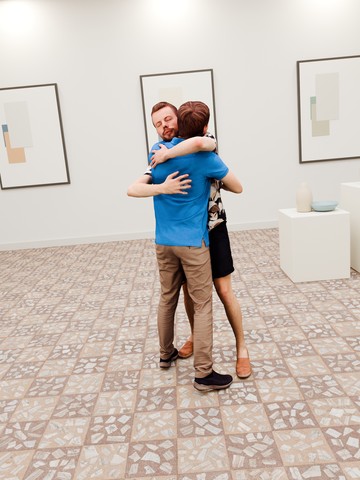}\\[-2pt]\footnotesize pair09\_\\[-3.5pt]\footnotesize hug09\end{tabular} & \begin{tabular}[t]{@{}c@{}}\includegraphics[width=0.112\textwidth]{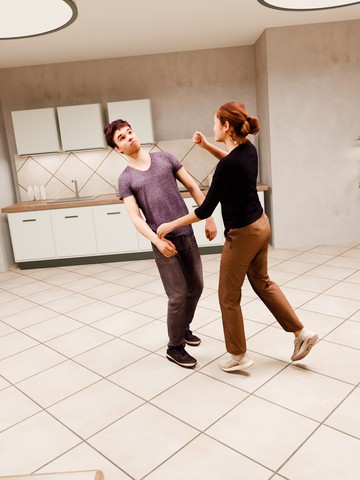}\\[-2pt]\footnotesize pair10\_\\[-3.5pt]\footnotesize argue10\end{tabular} & \begin{tabular}[t]{@{}c@{}}\includegraphics[width=0.112\textwidth]{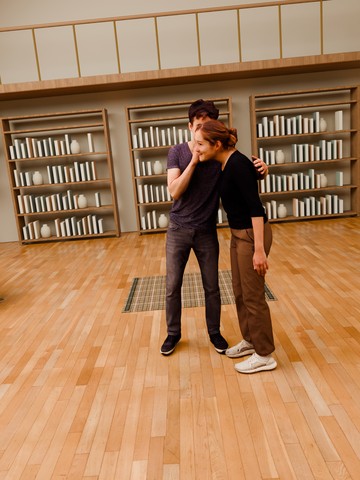}\\[-2pt]\footnotesize pair10\_\\[-3.5pt]\footnotesize talk10\end{tabular} & \begin{tabular}[t]{@{}c@{}}\includegraphics[width=0.112\textwidth]{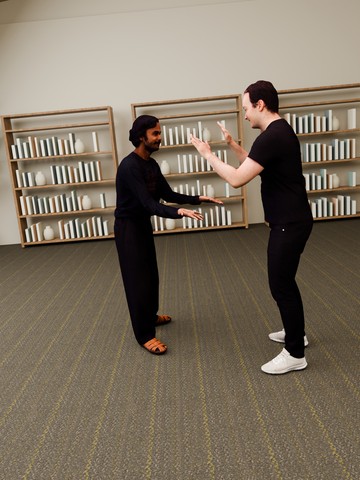}\\[-2pt]\footnotesize pair12\_\\[-3.5pt]\footnotesize highfive12\end{tabular} & \begin{tabular}[t]{@{}c@{}}\includegraphics[width=0.112\textwidth]{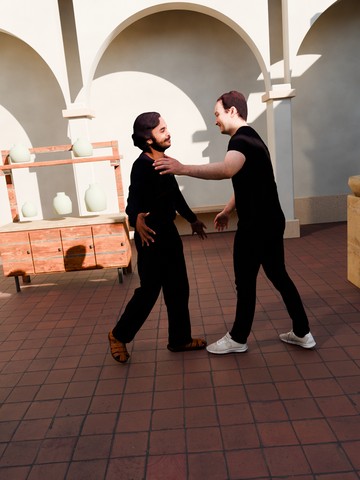}\\[-2pt]\footnotesize pair12\_\\[-3.5pt]\footnotesize hug12\end{tabular} & \begin{tabular}[t]{@{}c@{}}\includegraphics[width=0.112\textwidth]{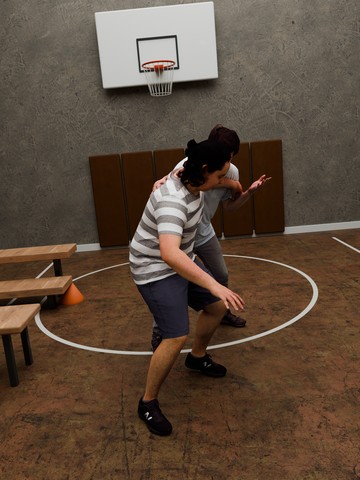}\\[-2pt]\footnotesize pair13\_\\[-3.5pt]\footnotesize basketball13\end{tabular} & \begin{tabular}[t]{@{}c@{}}\includegraphics[width=0.112\textwidth]{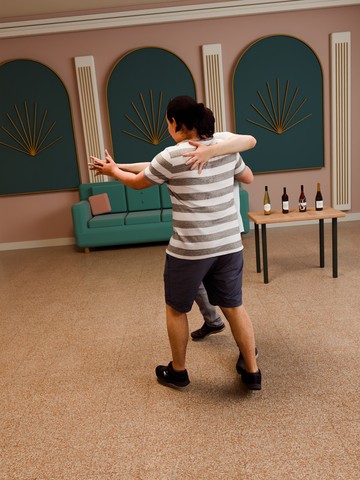}\\[-2pt]\footnotesize pair13\_\\[-3.5pt]\footnotesize dance13\end{tabular}\\[3pt]
    \begin{tabular}[t]{@{}c@{}}\includegraphics[width=0.112\textwidth]{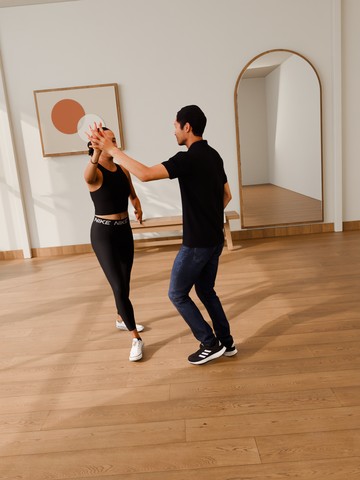}\\[-2pt]\footnotesize pair14\_\\[-3.5pt]\footnotesize dance14\end{tabular} & \begin{tabular}[t]{@{}c@{}}\includegraphics[width=0.112\textwidth]{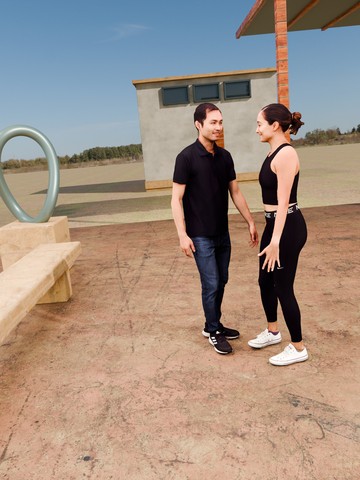}\\[-2pt]\footnotesize pair14\_\\[-3.5pt]\footnotesize pose14\end{tabular} & \begin{tabular}[t]{@{}c@{}}\includegraphics[width=0.112\textwidth]{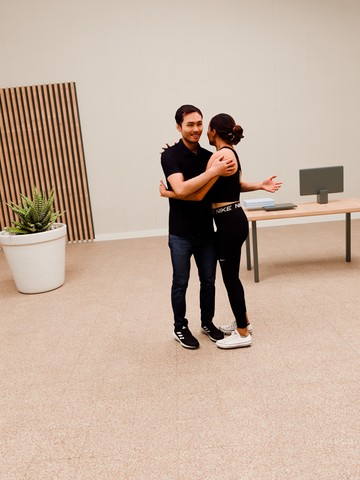}\\[-2pt]\footnotesize pair14\_\\[-3.5pt]\footnotesize talk14\end{tabular} & \begin{tabular}[t]{@{}c@{}}\includegraphics[width=0.112\textwidth]{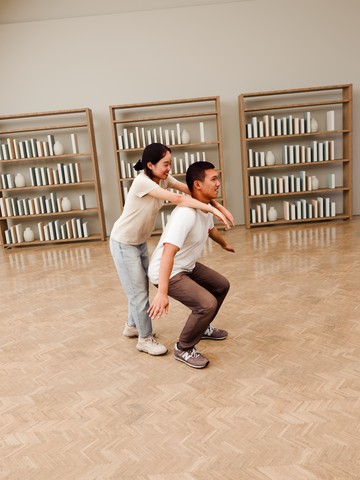}\\[-2pt]\footnotesize pair15\_\\[-3.5pt]\footnotesize backhug15\end{tabular} & \begin{tabular}[t]{@{}c@{}}\includegraphics[width=0.112\textwidth]{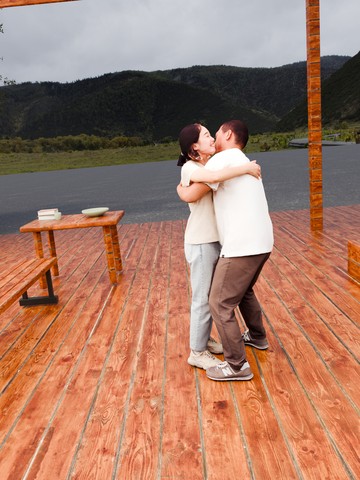}\\[-2pt]\footnotesize pair15\_\\[-3.5pt]\footnotesize kiss15\end{tabular} & \begin{tabular}[t]{@{}c@{}}\includegraphics[width=0.112\textwidth]{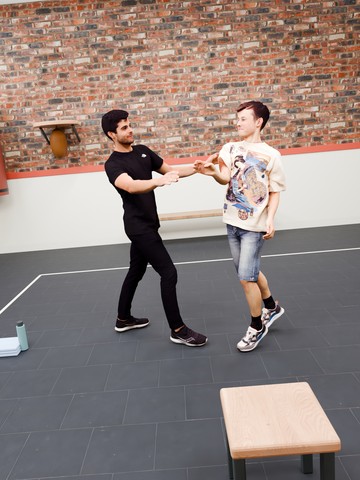}\\[-2pt]\footnotesize pair16\_\\[-3.5pt]\footnotesize fight16\end{tabular} & \begin{tabular}[t]{@{}c@{}}\includegraphics[width=0.112\textwidth]{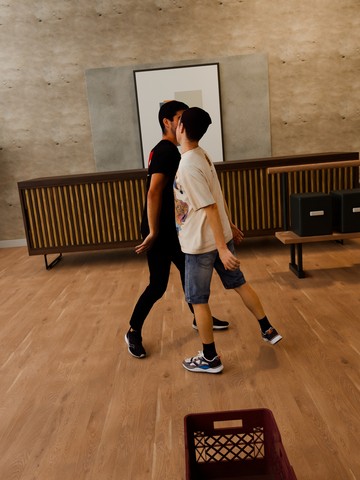}\\[-2pt]\footnotesize pair16\_\\[-3.5pt]\footnotesize jump16\end{tabular} & \begin{tabular}[t]{@{}c@{}}\includegraphics[width=0.112\textwidth]{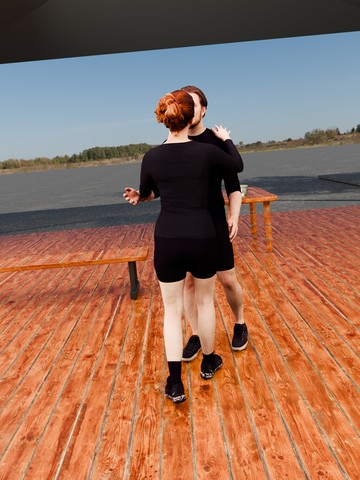}\\[-2pt]\footnotesize pair17\_\\[-3.5pt]\footnotesize kiss17\end{tabular}\\[3pt]
    \begin{tabular}[t]{@{}c@{}}\includegraphics[width=0.112\textwidth]{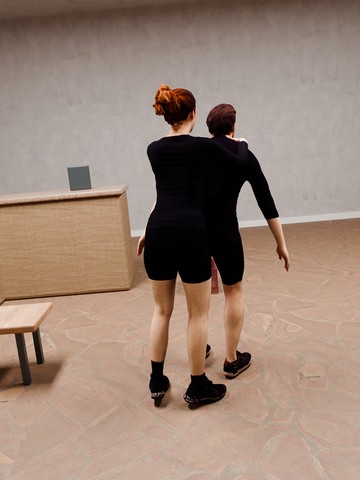}\\[-2pt]\footnotesize pair17\_\\[-3.5pt]\footnotesize piggyback17\end{tabular} & \begin{tabular}[t]{@{}c@{}}\includegraphics[width=0.112\textwidth]{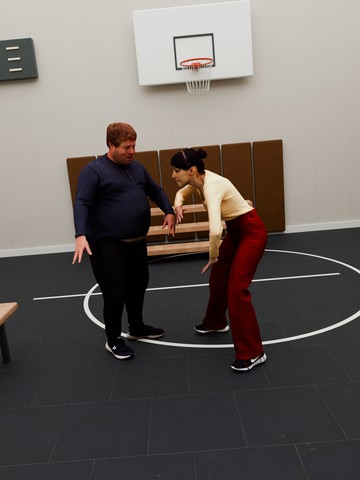}\\[-2pt]\footnotesize pair18\_\\[-3.5pt]\footnotesize basketball18\end{tabular} & \begin{tabular}[t]{@{}c@{}}\includegraphics[width=0.112\textwidth]{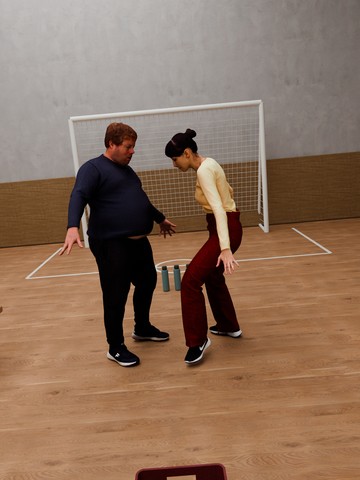}\\[-2pt]\footnotesize pair18\_\\[-3.5pt]\footnotesize football18\end{tabular} & \begin{tabular}[t]{@{}c@{}}\includegraphics[width=0.112\textwidth]{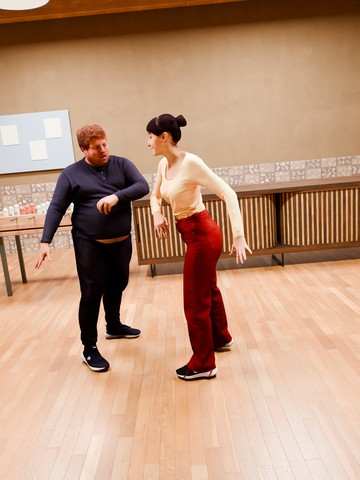}\\[-2pt]\footnotesize pair18\_\\[-3.5pt]\footnotesize highfive18\end{tabular} & \begin{tabular}[t]{@{}c@{}}\includegraphics[width=0.112\textwidth]{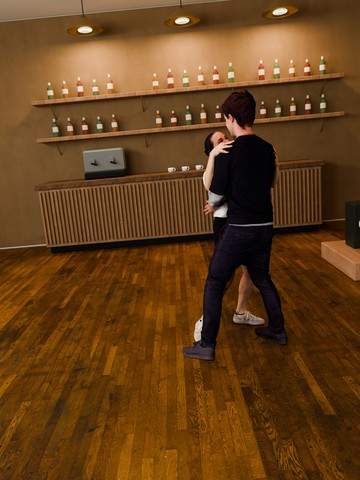}\\[-2pt]\footnotesize pair19\_\\[-3.5pt]\footnotesize dance19\end{tabular} & \begin{tabular}[t]{@{}c@{}}\includegraphics[width=0.112\textwidth]{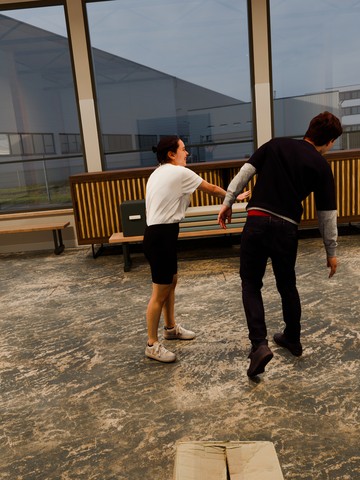}\\[-2pt]\footnotesize pair19\_\\[-3.5pt]\footnotesize fight19\end{tabular} & \begin{tabular}[t]{@{}c@{}}\includegraphics[width=0.112\textwidth]{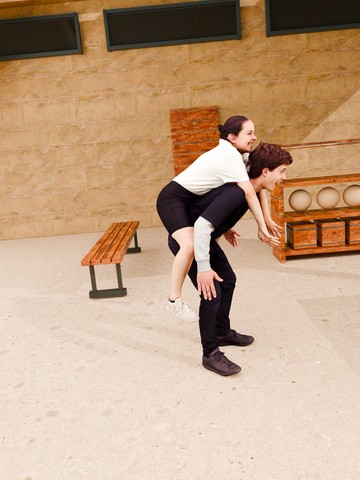}\\[-2pt]\footnotesize pair19\_\\[-3.5pt]\footnotesize piggyback19\end{tabular} & \begin{tabular}[t]{@{}c@{}}\includegraphics[width=0.112\textwidth]{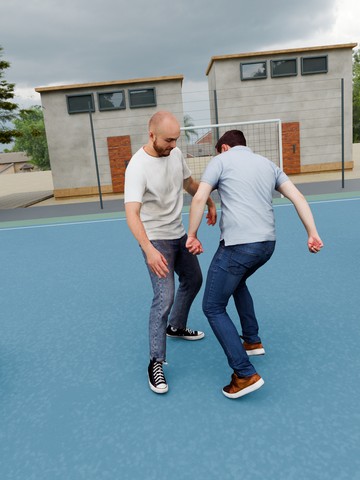}\\[-2pt]\footnotesize pair21\_\\[-3.5pt]\footnotesize football21\end{tabular}\\[3pt]
    \begin{tabular}[t]{@{}c@{}}\includegraphics[width=0.112\textwidth]{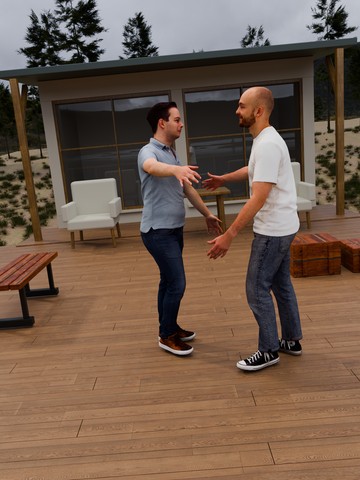}\\[-2pt]\footnotesize pair21\_\\[-3.5pt]\footnotesize hug21\end{tabular} & \begin{tabular}[t]{@{}c@{}}\includegraphics[width=0.112\textwidth]{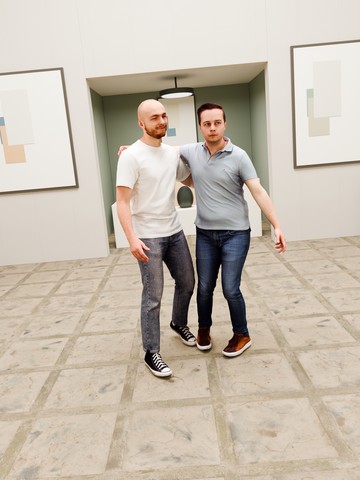}\\[-2pt]\footnotesize pair21\_\\[-3.5pt]\footnotesize pose21\end{tabular} & \begin{tabular}[t]{@{}c@{}}\includegraphics[width=0.112\textwidth]{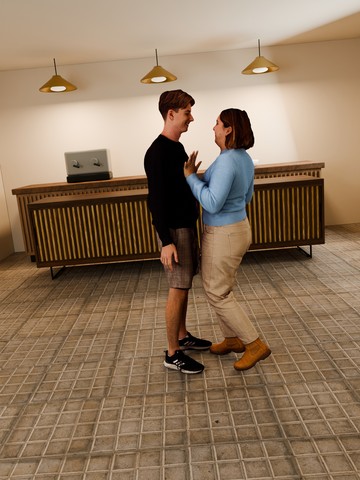}\\[-2pt]\footnotesize pair22\_\\[-3.5pt]\footnotesize game22\end{tabular} & \begin{tabular}[t]{@{}c@{}}\includegraphics[width=0.112\textwidth]{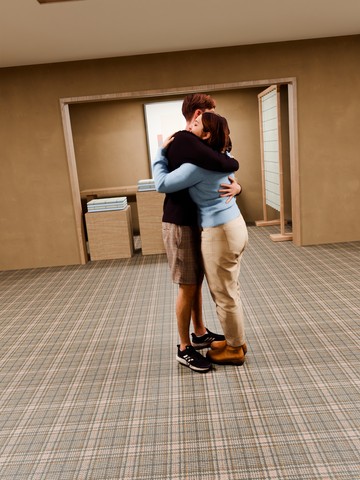}\\[-2pt]\footnotesize pair22\_\\[-3.5pt]\footnotesize hug22\end{tabular} & \begin{tabular}[t]{@{}c@{}}\includegraphics[width=0.112\textwidth]{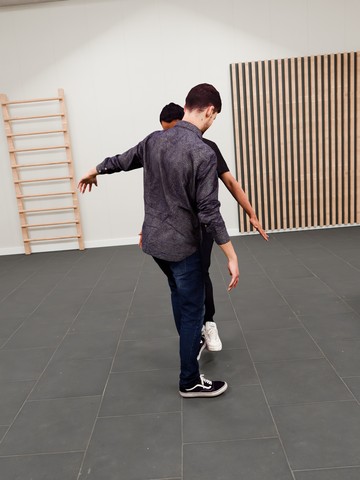}\\[-2pt]\footnotesize pair23\_\\[-3.5pt]\footnotesize leg23\end{tabular} & \begin{tabular}[t]{@{}c@{}}\includegraphics[width=0.112\textwidth]{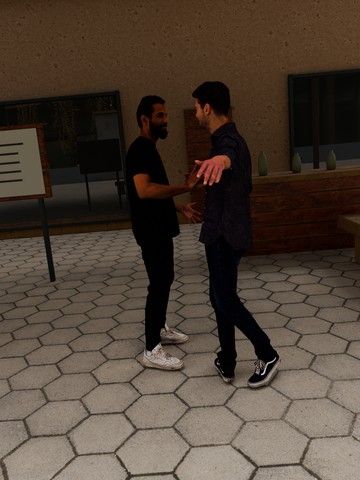}\\[-2pt]\footnotesize pair23\_\\[-3.5pt]\footnotesize talk23\end{tabular} & \begin{tabular}[t]{@{}c@{}}\includegraphics[width=0.112\textwidth]{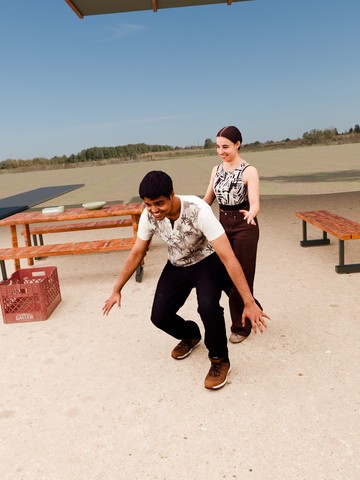}\\[-2pt]\footnotesize pair27\_\\[-3.5pt]\footnotesize piggy27\end{tabular} & \begin{tabular}[t]{@{}c@{}}\includegraphics[width=0.112\textwidth]{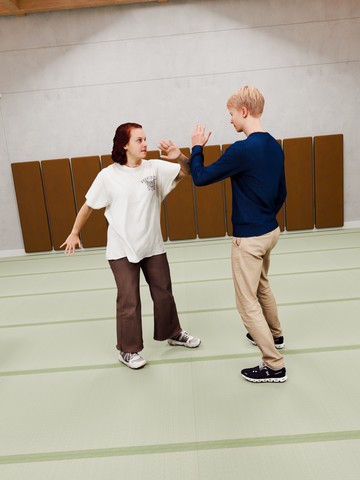}\\[-2pt]\footnotesize pair28\_\\[-3.5pt]\footnotesize fight28\end{tabular}\\[3pt]
    \begin{tabular}[t]{@{}c@{}}\includegraphics[width=0.112\textwidth]{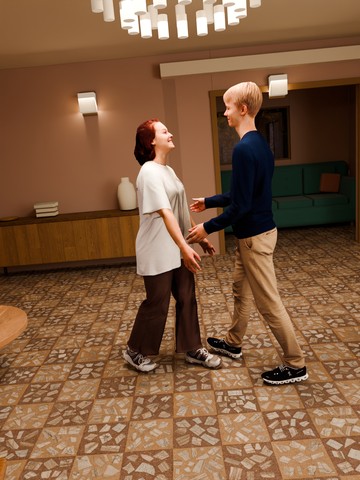}\\[-2pt]\footnotesize pair28\_\\[-3.5pt]\footnotesize kiss28\end{tabular} & \begin{tabular}[t]{@{}c@{}}\includegraphics[width=0.112\textwidth]{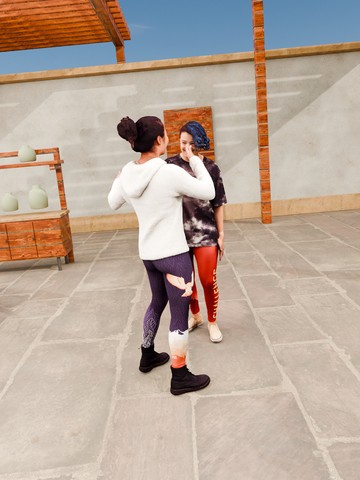}\\[-2pt]\footnotesize pair32\_\\[-3.5pt]\footnotesize cheers32\end{tabular} & \begin{tabular}[t]{@{}c@{}}\includegraphics[width=0.112\textwidth]{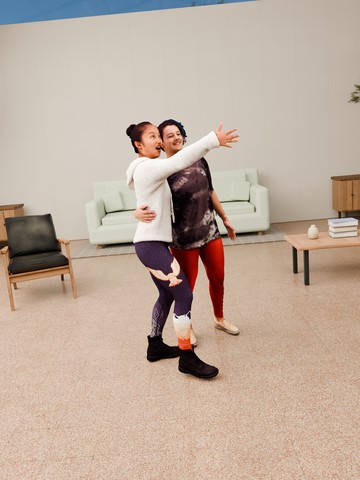}\\[-2pt]\footnotesize pair32\_\\[-3.5pt]\footnotesize pose32\end{tabular} & \begin{tabular}[t]{@{}c@{}}\includegraphics[width=0.112\textwidth]{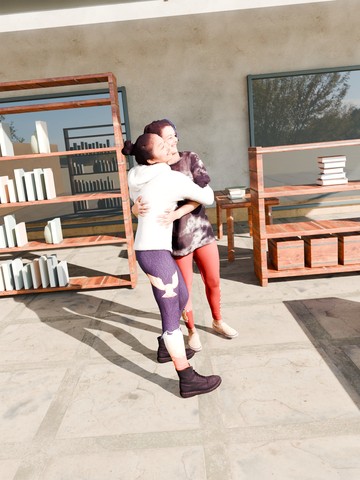}\\[-2pt]\footnotesize pair32\_\\[-3.5pt]\footnotesize sidehug32\end{tabular} & \begin{tabular}[t]{@{}c@{}}\includegraphics[width=0.112\textwidth]{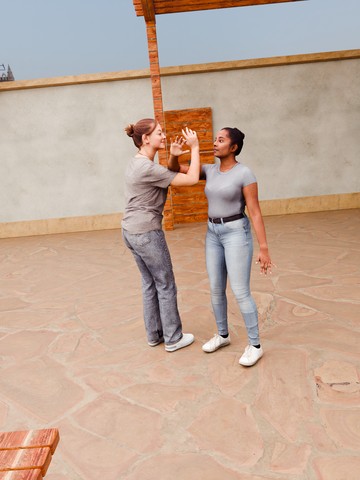}\\[-2pt]\footnotesize pair37\_\\[-3.5pt]\footnotesize cheers37\end{tabular} & \begin{tabular}[t]{@{}c@{}}\includegraphics[width=0.112\textwidth]{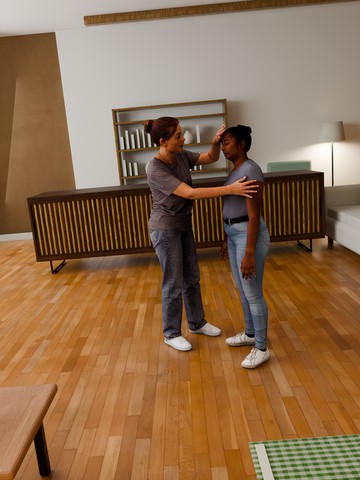}\\[-2pt]\footnotesize pair37\_\\[-3.5pt]\footnotesize fever37\end{tabular} & \begin{tabular}[t]{@{}c@{}}\includegraphics[width=0.112\textwidth]{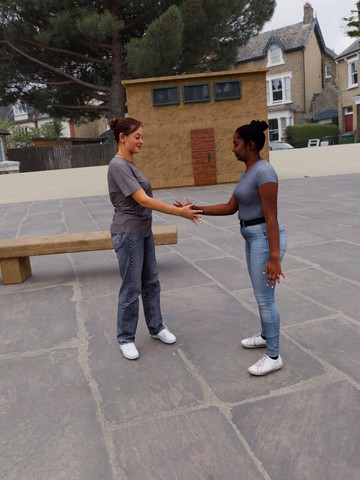}\\[-2pt]\footnotesize pair37\_\\[-3.5pt]\footnotesize handshake37\end{tabular} & \begin{tabular}[t]{@{}c@{}}\includegraphics[width=0.112\textwidth]{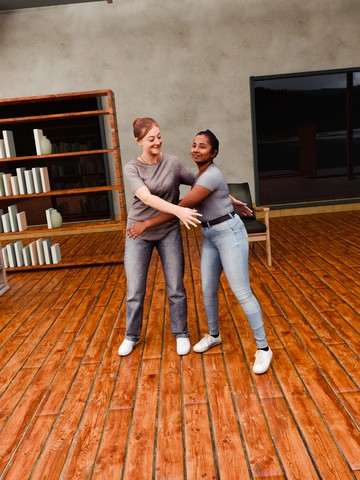}\\[-2pt]\footnotesize pair37\_\\[-3.5pt]\footnotesize sidehug37\end{tabular}\\
  \end{tabular}
  \vspace{-0.2cm}
  \caption{\textbf{The 48 rendered Hi4D environments.} One frame per sequence, from static camera 0 at one third of the clip, with the sequence name below it. Every sequence has its own environment, 30 indoor and 18 outdoor or semi-outdoor.}
  \label{fig:hi4d_environments}
\end{figure*}

We place each sequence in its own synthetic environment, assembled in Blender from procedural architecture and CC0 assets from Poly Haven (textures, models, and HDRIs). The 48 setups comprise 30 indoor and 18 outdoor or semi-outdoor scenes drawn from 19 shared environment families, such as studios, lofts, libraries, kitchens, sports halls, terraces, gardens, and a lakeside deck, and were refined through visual review (\cref{fig:hi4d_environments}). The public per-frame scans enter at metric scale with their original geometry and $4096\times4096$ texture atlases, and one constant translation per sequence places them on the environment floor, so the released cameras, tracks, and visibility stay in their original coordinates. People and environment are rendered together with Blender~4.2 Cycles at $1440\times1920$ resolution from the four fixed and four moving cameras, using 128 samples per pixel, guided OpenImageDenoise, and no motion blur or depth of field. Because the scan atlases contain baked illumination, the people are shaded as matte surfaces, which relights the captured appearance only approximately. Ground-truth depth is the camera-axis depth of the first surface hit, rendered in a separate undenoised pass. Pixels without modeled geometry, glass, mirror, and water surfaces, and depths beyond 20\,m are marked invalid and excluded from evaluation. The human tracks and their visibility are unchanged by the environments: a ray-casting audit of all 1,757,587 originally visible track observations found no environment surface in front of any of them, with a 1\,mm tolerance.

Each sequence also receives as many static tracks on the environment as it has human tracks, for 10,456 tracks in total. We sample candidate points area-uniformly on opaque environment surfaces, compute their visibility in every frame and view by ray casting against the environment and the current scan with the same 1\,mm tolerance, and keep candidates that are visible in at least two views and in at least 20\% of the frames. From these we select the tracks by farthest-point sampling for spatial coverage and draw each query uniformly from the visible view and frame pairs at which the rendered depth at the query pixel agrees with the point to within 2\,cm and 1\%.

\begin{figure}[t]
  \centering
  \includegraphics[width=\linewidth]{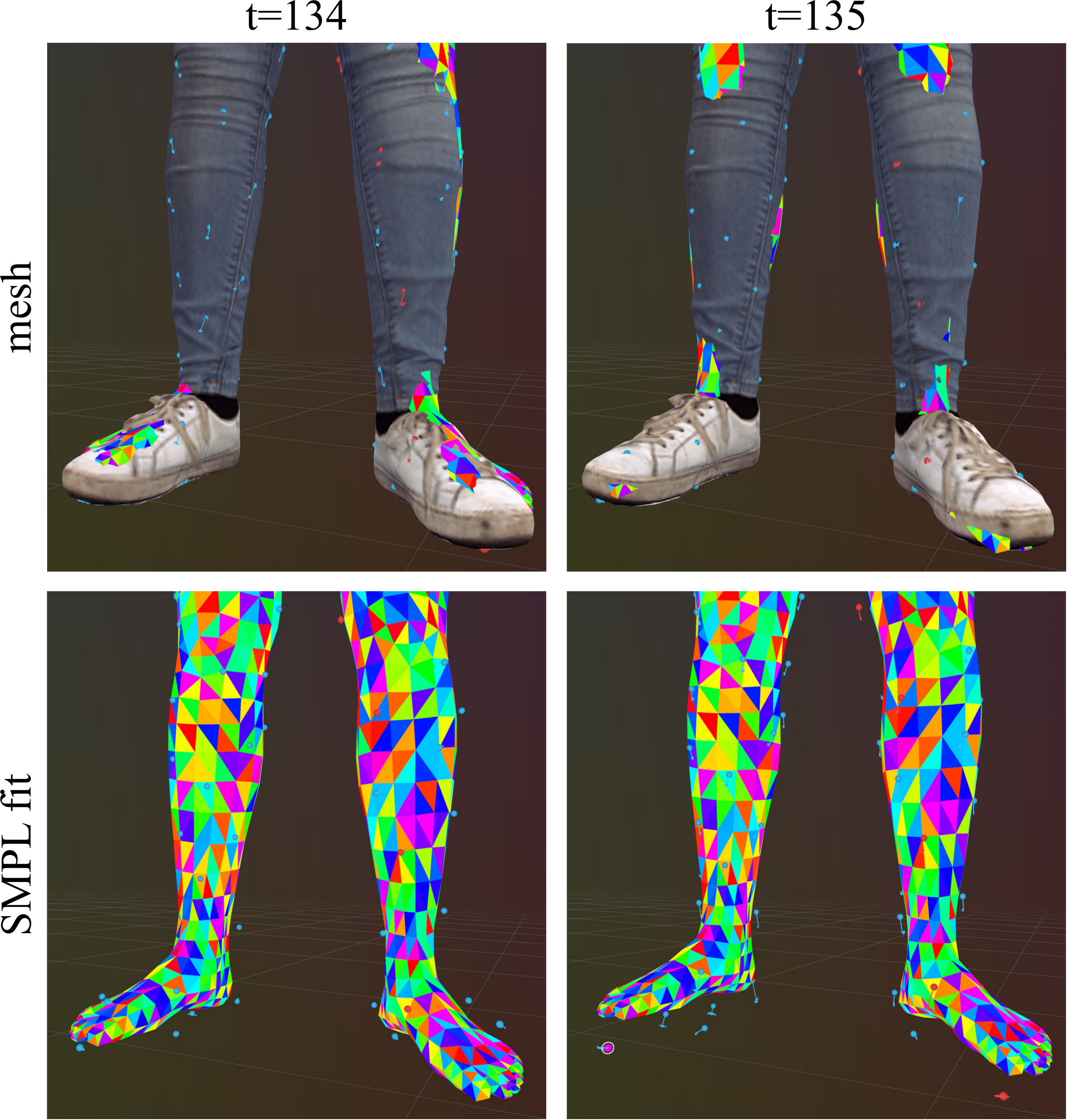}
  \caption{\textbf{Noisy Hi4D body fits create invalid track candidates.} The public scan mesh (top) and fitted SMPL bodies (bottom) are shown at two consecutive frames. Here, the fitted feet translate upward by approximately 5\,cm in the second frame, although the feet in the public scan remain stationary. Such temporal instability can make a fixed body-surface anchor slide over or detach from the observed surface, causing the candidate to be rejected during manual review.}
  \label{fig:hi4d_labeling}
\end{figure}

\paragraph{Manual review statistics.}
\Cref{tab:hi4d_labeling} reports the distribution of the review effort over the 48 sequences. Across the split the annotator reviewed 12,857 of the 28,800 generated candidates (44.6\%) and rejected 7,629 of them (59.3\%), spending 20.7 hours in total. The rejection rate varies widely by sequence (27.7\% to 74.0\%) because it is set primarily by the quality and temporal stability of the released SMPL fits. Misaligned or jittering fits make candidate tracks slide over or detach from the observed scan surface, as illustrated in \cref{fig:hi4d_labeling}.

\begin{table}[h]
  \centering
  \setlength{\tabcolsep}{4pt}
  \caption{\textbf{Hi4D manual review effort, over the 48 sequences.} Candidates are reviewed in a fixed random order until at least 200 have been reviewed and at least 100 have been accepted, so the number reviewed varies with the rejection rate. Review time excludes gaps longer than three minutes.}
  \label{tab:hi4d_labeling}
  \vspace{-0.27cm}
  \small
  \begin{tabular}{lrrrrrr}
    \toprule
    & Min & P25 & P50 & P75 & P90 & Max \\
    \midrule
    Candidates reviewed        & 200  & 225  & 264  & 304  & 327  & 385  \\
    Tracks accepted            & 100  & 100  & 100  & 110  & 132  & 180  \\
    Rejection rate (\%)        & 27.7 & 49.3 & 61.8 & 65.7 & 69.4 & 74.0 \\
    Review time (min)          & 11.1 & 19.8 & 23.1 & 27.2 & 36.7 & 89.6 \\
    Time per decision (s)      & 1.6  & 2.9  & 3.9  & 4.9  & 5.6  & 9.1  \\
    \bottomrule
  \end{tabular}
\end{table}

\subsection{Perpetua Procedural Generation Details}
\label{sec:supp_datasets_perpetua}
As described in \cref{sec:tapvidmv:perpetua}, Perpetua generates complete indoor scenes, animated actors, moving cameras, and metric point trajectories.

\paragraph{Environment and actor motion.}
Infinigen Indoors~\cite{infinigen2024indoors} generates a furnished room and exposes its final collision geometry. We rasterize the walkable surfaces into a 2.5D navigation grid and seed four actor routes with independent geometry-aware random walks. Adam jointly optimizes their dense root trajectories for scene clearance, actor separation, bounded speed and acceleration, and smooth, spatially varied motion. Routes that violate the navigation or motion constraints are rejected. For each accepted route, Kimodo~\cite{Kimodo2026} generates full-body motion from a natural-language action description while following the prescribed root trajectory. The resulting motion is applied to a dressed SMPL-X actor~\cite{SMPL-X:2019,tesch2025bedlam2}, assembled in the scene, and validated before the actor stage is accepted.

\paragraph{Camera trajectories.}
We construct a 3D signed-distance field from the scene collision mesh and sample camera controls from free space. Each view receives an actor-focus schedule that may switch targets over time. From multiple initial candidates, Adam jointly optimizes four camera trajectories for scene and actor clearance, actor framing and coverage, smooth translation and rotation, spatially varied motion, limited revisiting, level roll, and viewpoint diversity. We retain the lowest-loss candidate that passes a final collision-geometry audit.

\paragraph{Tracks and visibility.}
We render synchronized RGB, metric depth, and object-index images for all four cameras. Static candidates are sampled from rendered environment surfaces, while dynamic candidates are sampled directly from the animated SMPL-X meshes. Their world-space positions are propagated using the exact scene transforms and deforming actor geometry. Visibility is determined by camera-to-point ray casting against the scene geometry, accounting for self-occlusion and occlusion by other scene elements. This yields exact 3D trajectories and per-view visibility labels for both actors and the environment.

\subsection{Ego-Exo4D Track Construction}
\label{sec:supp_datasets_egoexo4d}
As described in \cref{sec:tapvidmv:egoexo4d}, we select soccer sequences in which the head-camera wearer interacts with another participant visible from the head-mounted view and at least two static cameras. For dynamic tracks, we apply THFM~\cite{wang2026thfm} independently to the static cameras, then fuse and optimize the predicted human meshes to match both calibrated views. Static candidates come from the dataset-provided SLAM point cloud and are filtered by agreement with dense VGGT-$\Omega$~\cite{wang2026vggtomega} depth maps.

We determine visibility using agreement with the human-mesh depth and additionally mask occlusions from other humans and the ball using SAM~3~\cite{sam3}. From the surviving candidates, we select 500 tracks that are jointly visible from all three cameras for at least 50 frames.

\subsection{Waymo Track Construction}
\label{sec:supp_datasets_waymo}
As described in \cref{sec:tapvidmv:waymo}, we extend the DriveTrack-style trajectory construction~\cite{balasingam2024drivetrack,koppula2024tapvid3d} to a synchronized triplet of adjacent front cameras. At each frame, LiDAR points inside a vehicle's manually annotated 3D bounding box are associated with that vehicle. The vehicle rigidity assumption and annotated per-frame box pose propagate these points through time; composing with the ego-vehicle pose places the resulting trajectories in a world frame shared by all views.

For visibility, we compare each point's expected camera distance with the depth of the closest LiDAR measurement at the corresponding timestamp. We finally sample 500 tracks per sequence that are visible in at least two views, including tracks on vehicles and on the background scene. Visual spot checks reveal moderate cross-view misalignment for some tracks and occasional incorrect visibility labels.

\section{Hi4D Static-Only Split}
\label{sec:hi4d_static_only}
\begin{figure*}[t]
  \centering
  \setlength{\tabcolsep}{1pt}
  \begin{tabular}{c*{8}{c}}
    & \footnotesize cam 0 & \footnotesize cam 1 & \footnotesize cam 2 & \footnotesize cam 3 & \footnotesize cam 4 & \footnotesize cam 5 & \footnotesize cam 6 & \footnotesize cam 7\\
    \raisebox{-0.5\height}{\rotatebox{90}{\footnotesize Hi4D, $t{=}0$}} & \raisebox{-0.5\height}{\includegraphics[width=0.114\textwidth]{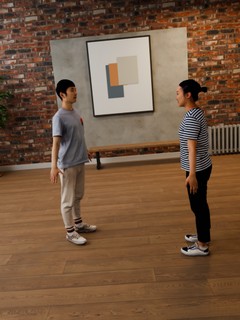}} & \raisebox{-0.5\height}{\includegraphics[width=0.114\textwidth]{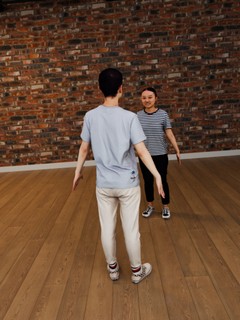}} & \raisebox{-0.5\height}{\includegraphics[width=0.114\textwidth]{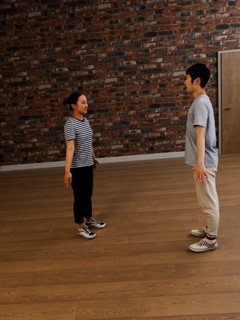}} & \raisebox{-0.5\height}{\includegraphics[width=0.114\textwidth]{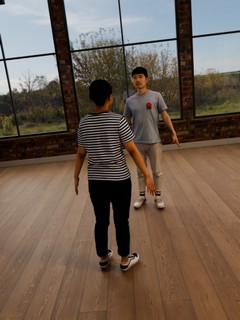}} & \raisebox{-0.5\height}{\includegraphics[width=0.114\textwidth]{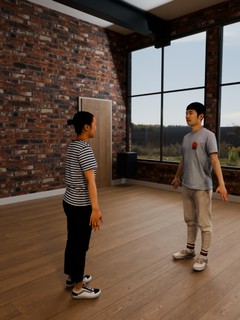}} & \raisebox{-0.5\height}{\includegraphics[width=0.114\textwidth]{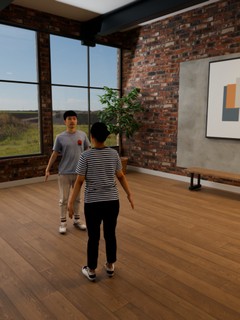}} & \raisebox{-0.5\height}{\includegraphics[width=0.114\textwidth]{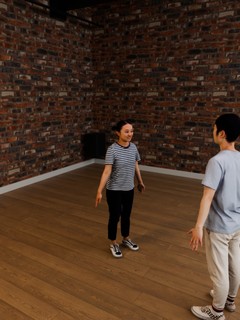}} & \raisebox{-0.5\height}{\includegraphics[width=0.114\textwidth]{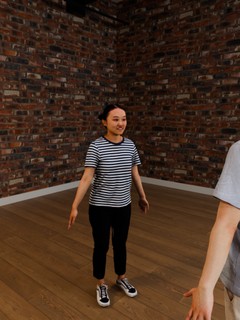}}\\
    \addlinespace[2pt]
    \raisebox{-0.5\height}{\rotatebox{90}{\footnotesize Hi4D, $t{=}T$}} & \raisebox{-0.5\height}{\includegraphics[width=0.114\textwidth]{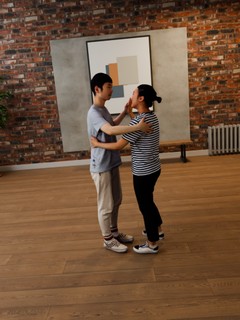}} & \raisebox{-0.5\height}{\includegraphics[width=0.114\textwidth]{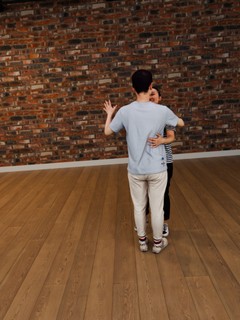}} & \raisebox{-0.5\height}{\includegraphics[width=0.114\textwidth]{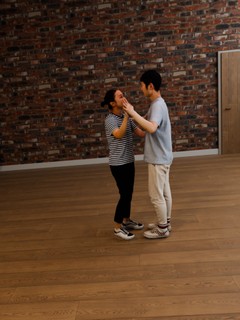}} & \raisebox{-0.5\height}{\includegraphics[width=0.114\textwidth]{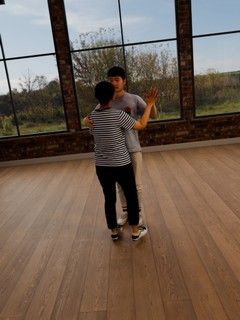}} & \raisebox{-0.5\height}{\includegraphics[width=0.114\textwidth]{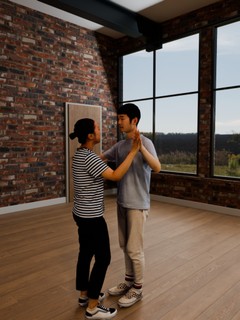}} & \raisebox{-0.5\height}{\includegraphics[width=0.114\textwidth]{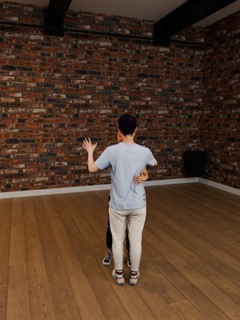}} & \raisebox{-0.5\height}{\includegraphics[width=0.114\textwidth]{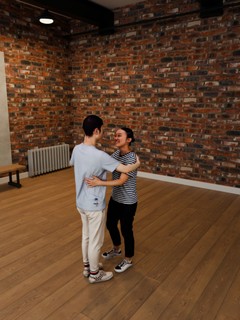}} & \raisebox{-0.5\height}{\includegraphics[width=0.114\textwidth]{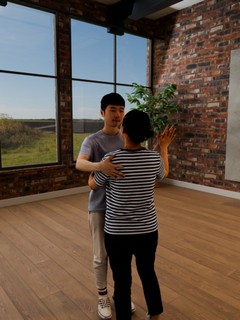}}\\
    \addlinespace[4pt]
    \raisebox{-0.5\height}{\rotatebox{90}{\footnotesize Hi4D-static, $t{=}0$}} & \raisebox{-0.5\height}{\includegraphics[width=0.114\textwidth]{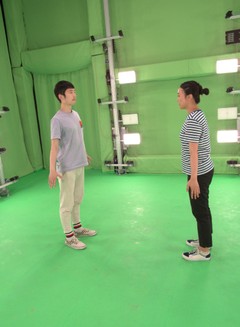}} & \raisebox{-0.5\height}{\includegraphics[width=0.114\textwidth]{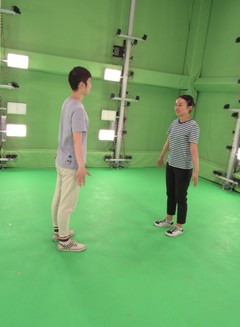}} & \raisebox{-0.5\height}{\includegraphics[width=0.114\textwidth]{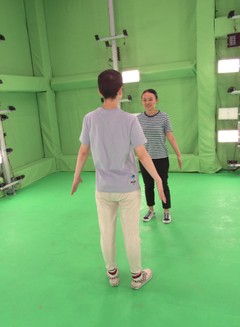}} & \raisebox{-0.5\height}{\includegraphics[width=0.114\textwidth]{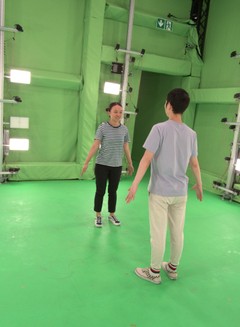}} & \raisebox{-0.5\height}{\includegraphics[width=0.114\textwidth]{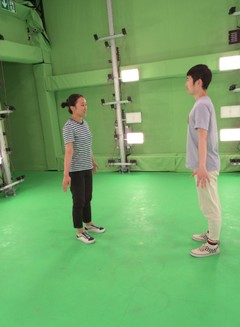}} & \raisebox{-0.5\height}{\includegraphics[width=0.114\textwidth]{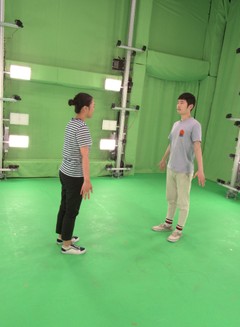}} & \raisebox{-0.5\height}{\includegraphics[width=0.114\textwidth]{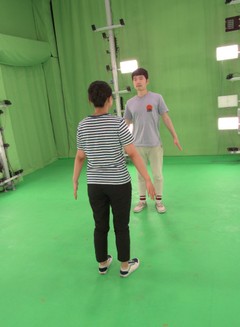}} & \raisebox{-0.5\height}{\includegraphics[width=0.114\textwidth]{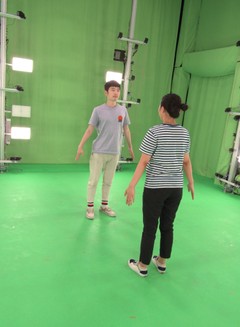}}\\
    \addlinespace[2pt]
    \raisebox{-0.5\height}{\rotatebox{90}{\footnotesize Hi4D-static, $t{=}T$}} & \raisebox{-0.5\height}{\includegraphics[width=0.114\textwidth]{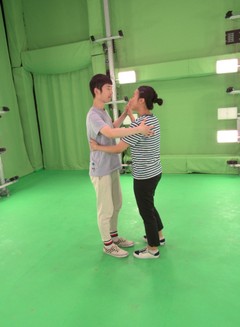}} & \raisebox{-0.5\height}{\includegraphics[width=0.114\textwidth]{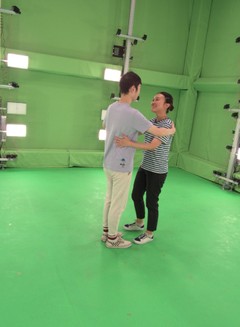}} & \raisebox{-0.5\height}{\includegraphics[width=0.114\textwidth]{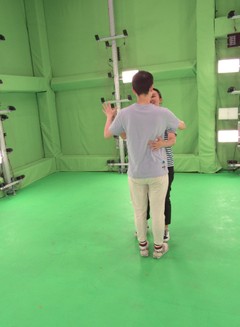}} & \raisebox{-0.5\height}{\includegraphics[width=0.114\textwidth]{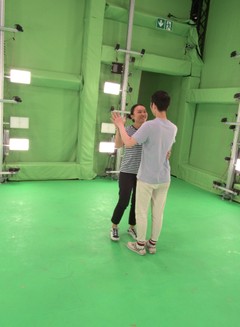}} & \raisebox{-0.5\height}{\includegraphics[width=0.114\textwidth]{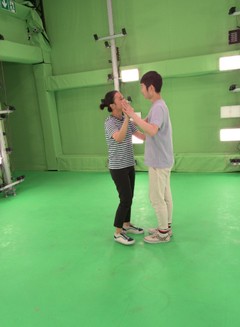}} & \raisebox{-0.5\height}{\includegraphics[width=0.114\textwidth]{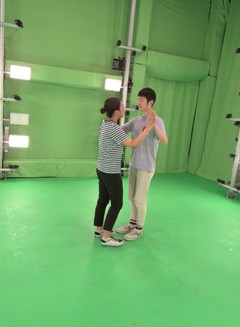}} & \raisebox{-0.5\height}{\includegraphics[width=0.114\textwidth]{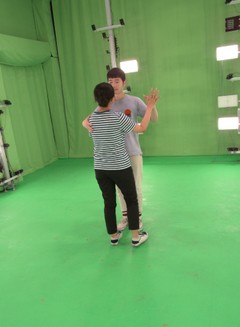}} & \raisebox{-0.5\height}{\includegraphics[width=0.114\textwidth]{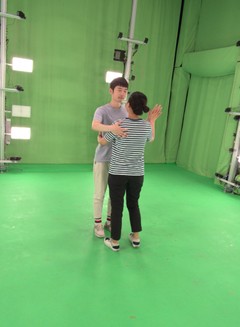}}\\
  \end{tabular}
  \vspace{-0.2cm}
  \caption{\textbf{Hi4D rendered split versus the static-only split} on \texttt{pair00\_dance00}, all eight cameras at the first and the last frame. Rows 1--2: the rendered split of \cref{tab:main_results}, the textured scans rendered in a synthetic environment; cameras 0--3 are the public static cameras and cameras 4--7 simulated moving cameras, whose viewpoint changes between the two rows. Rows 3--4: the static-only companion split, the original RGB captures of the eight static cameras; its cameras 0--3 coincide with cameras 0--3 of the rendered split.}
  \label{fig:hi4d_static_split}
\end{figure*}

To evaluate the same manually verified 3D tracks on real-world RGB captures rather than only on our renders, we add a static-only Hi4D companion split built from the original public 8-view RGB videos distributed by Hi4D~\cite{yin2023hi4d}. This split pairs those RGB videos with matching \benchmarkname labels, cameras, and metadata under the same Hi4D-derived license terms. \Cref{fig:hi4d_static_split} shows both splits on one sequence: the rendered split of the main benchmark views the textured scans in a synthetic environment from the four public static cameras and four simulated moving cameras, while the static-only split keeps the original captures, so the two splits share cameras 0--3 and differ in appearance (rendered scan versus real footage) and in camera motion. \Cref{tab:hi4d_static_reconstruction,tab:hi4d_static_results} report reconstruction and tracking quality on both splits.

\begin{figure*}[t]
  \centering
  \setlength{\tabcolsep}{1.5pt}
  \begin{tabular}{cccc@{\hspace{6pt}}ccc}
    & \multicolumn{3}{c}{\footnotesize Hi4D (rendered)} & \multicolumn{3}{c}{\footnotesize Hi4D-static (original captures)}\\
    & \footnotesize Ground truth & \footnotesize VGGT-$\Omega$ & \footnotesize OmniX & \footnotesize Ground truth & \footnotesize VGGT-$\Omega$ & \footnotesize OmniX\\
    \raisebox{-0.5\height}{\rotatebox{90}{\footnotesize scene, orbit}} & \raisebox{-0.5\height}{\includegraphics[width=0.155\textwidth]{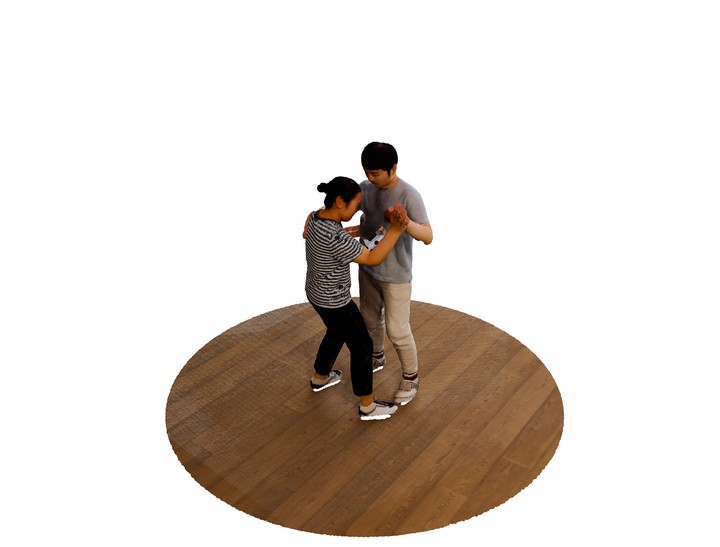}} & \raisebox{-0.5\height}{\includegraphics[width=0.155\textwidth]{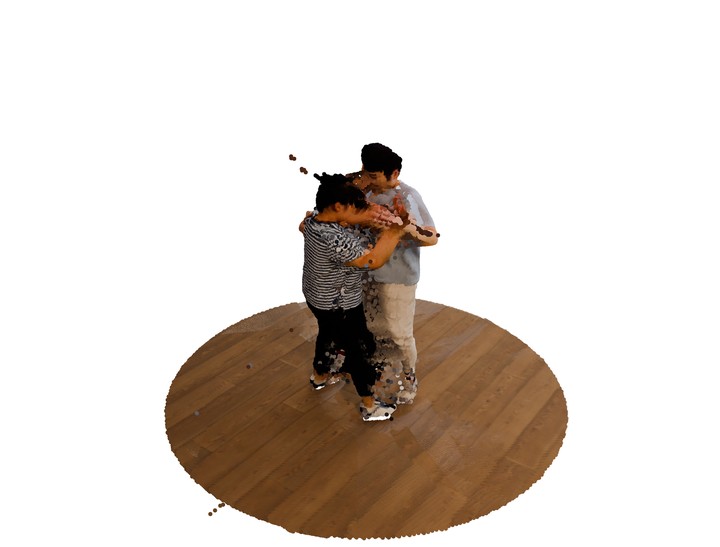}} & \raisebox{-0.5\height}{\includegraphics[width=0.155\textwidth]{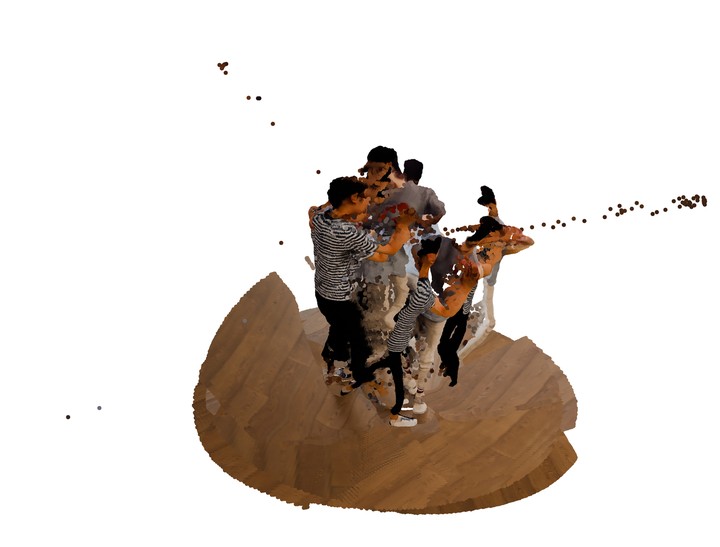}} & \raisebox{-0.5\height}{\includegraphics[width=0.155\textwidth]{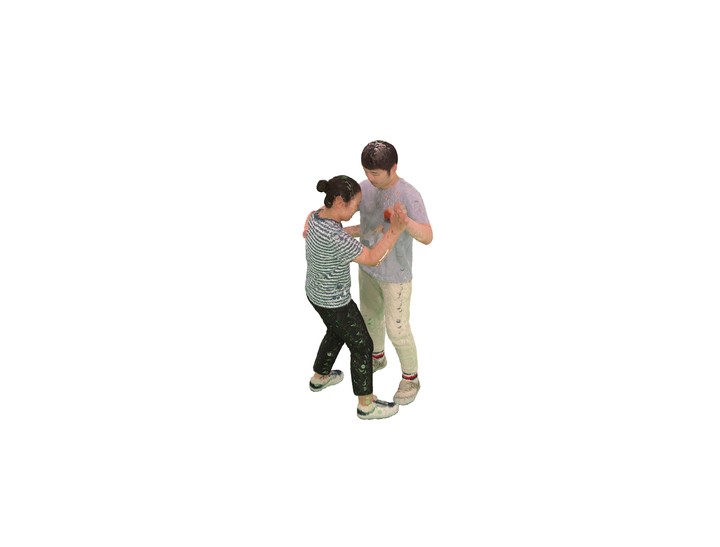}} & \raisebox{-0.5\height}{\includegraphics[width=0.155\textwidth]{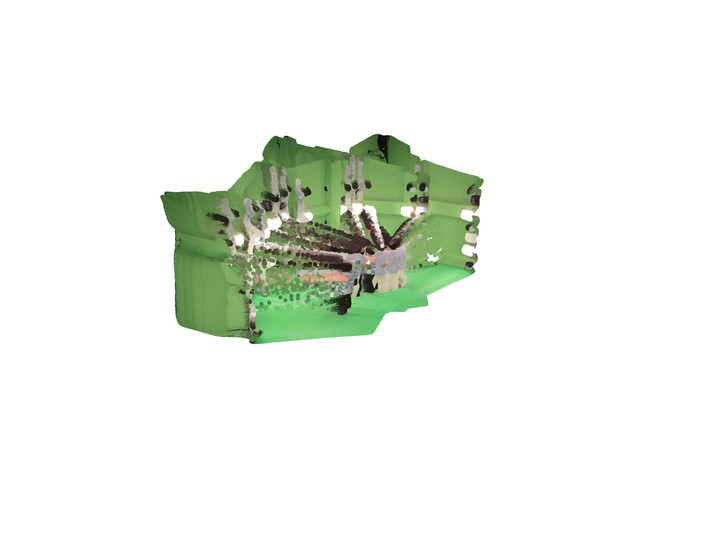}} & \raisebox{-0.5\height}{\includegraphics[width=0.155\textwidth]{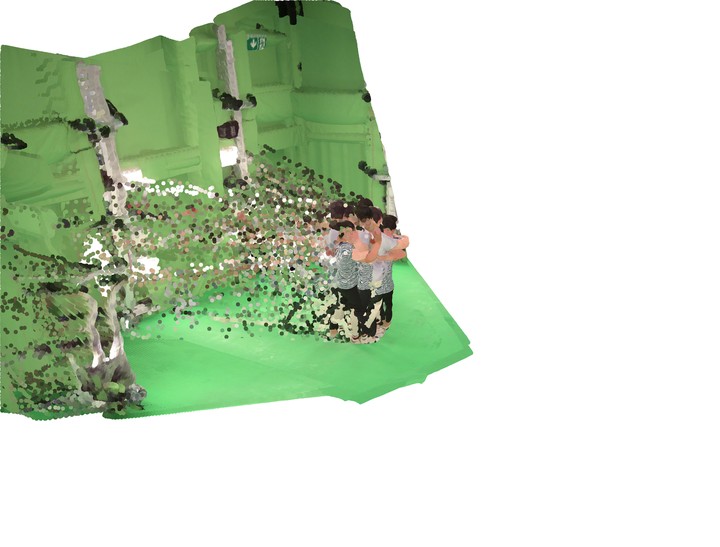}}\\
    \raisebox{-0.5\height}{\rotatebox{90}{\footnotesize scene, top}} & \raisebox{-0.5\height}{\includegraphics[width=0.155\textwidth]{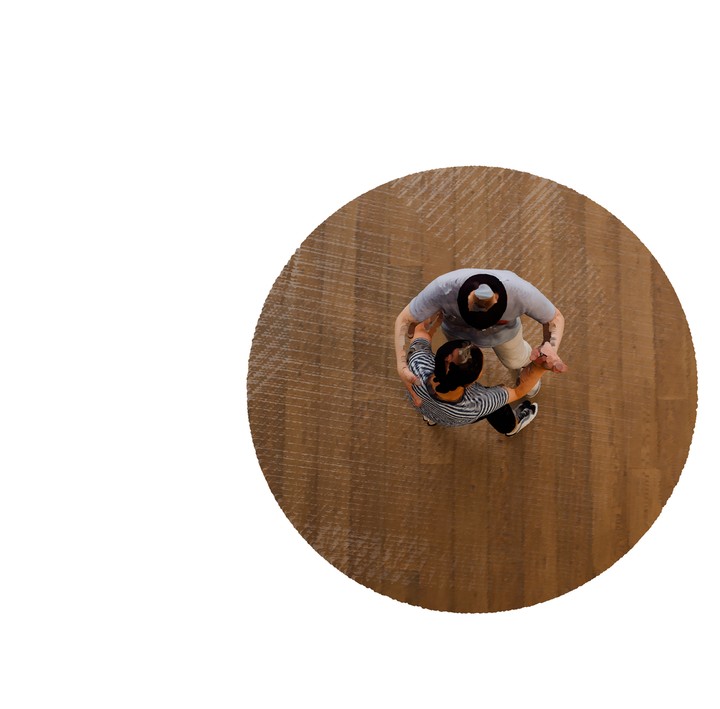}} & \raisebox{-0.5\height}{\includegraphics[width=0.155\textwidth]{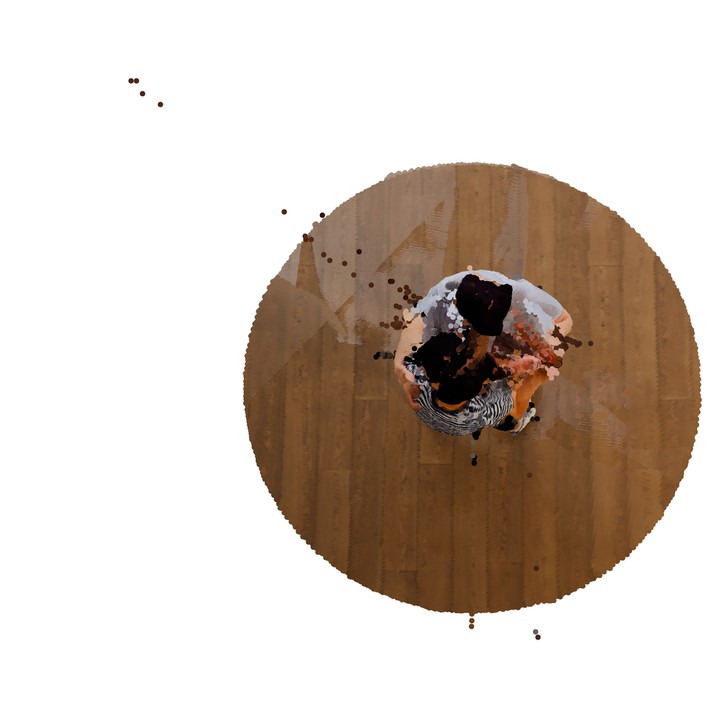}} & \raisebox{-0.5\height}{\includegraphics[width=0.155\textwidth]{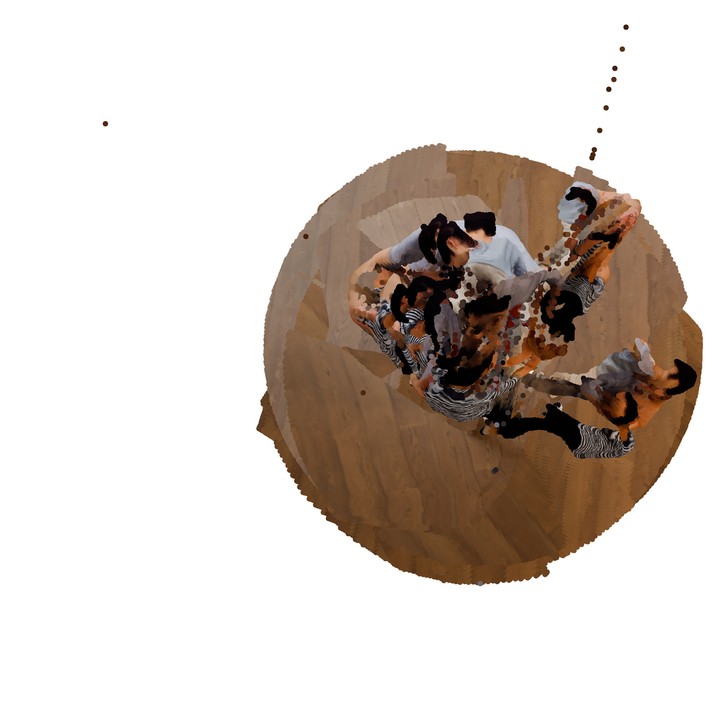}} & \raisebox{-0.5\height}{\includegraphics[width=0.155\textwidth]{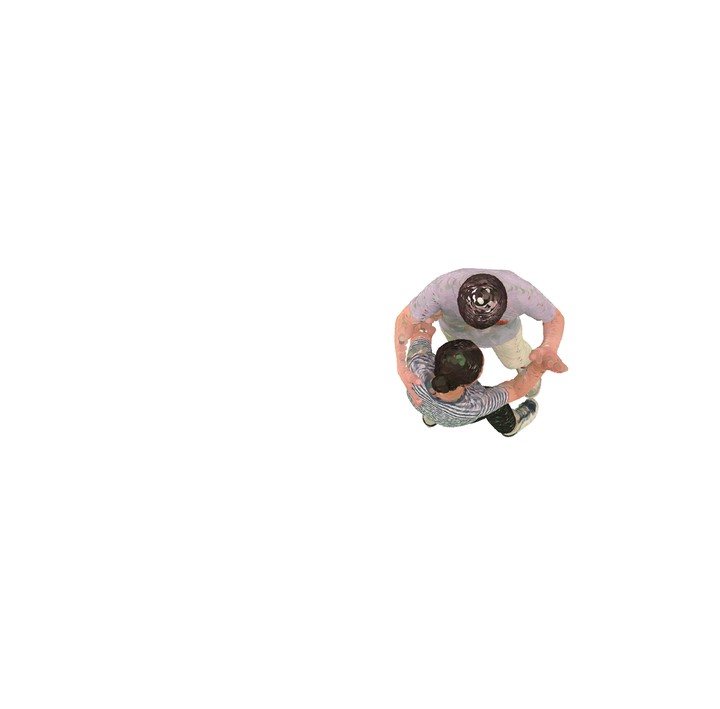}} & \raisebox{-0.5\height}{\includegraphics[width=0.155\textwidth]{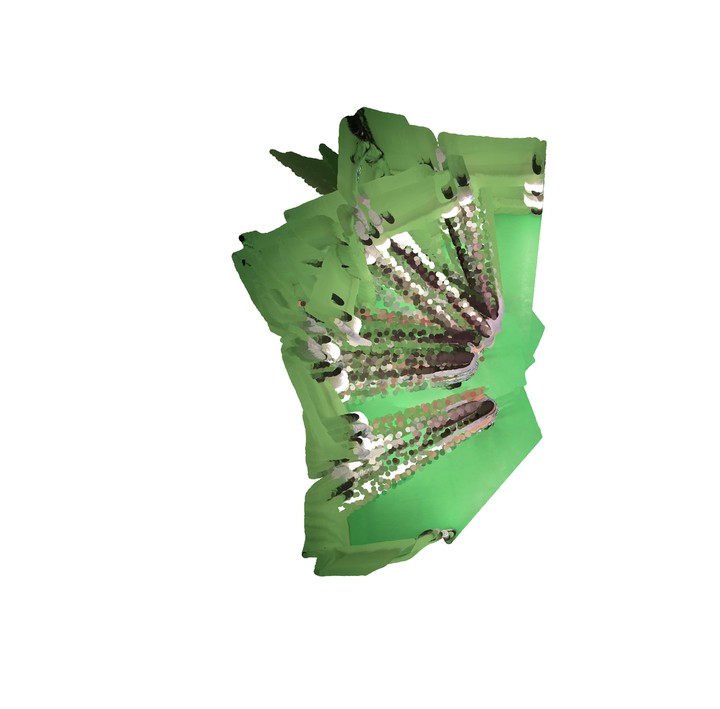}} & \raisebox{-0.5\height}{\includegraphics[width=0.155\textwidth]{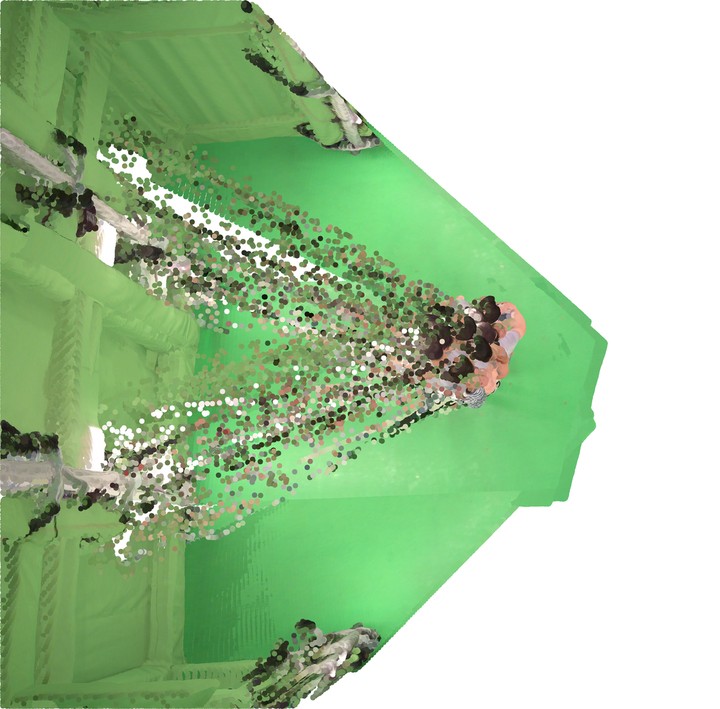}}\\
    \raisebox{-0.5\height}{\rotatebox{90}{\footnotesize actors, orbit}} & \raisebox{-0.5\height}{\includegraphics[width=0.155\textwidth]{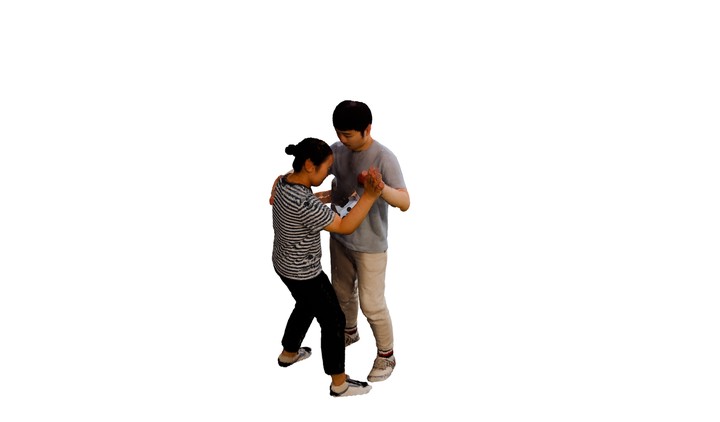}} & \raisebox{-0.5\height}{\includegraphics[width=0.155\textwidth]{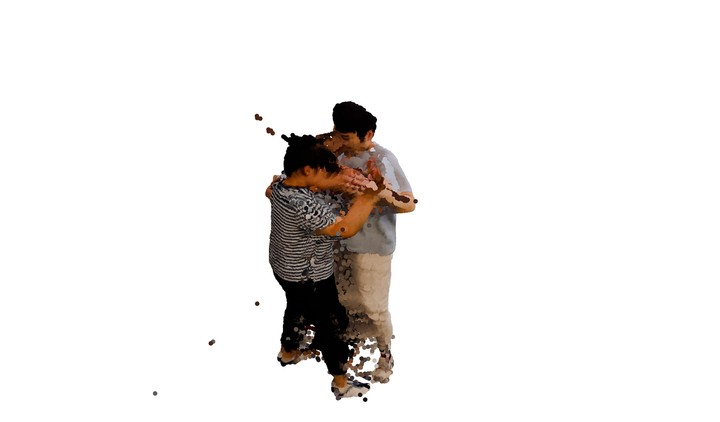}} & \raisebox{-0.5\height}{\includegraphics[width=0.155\textwidth]{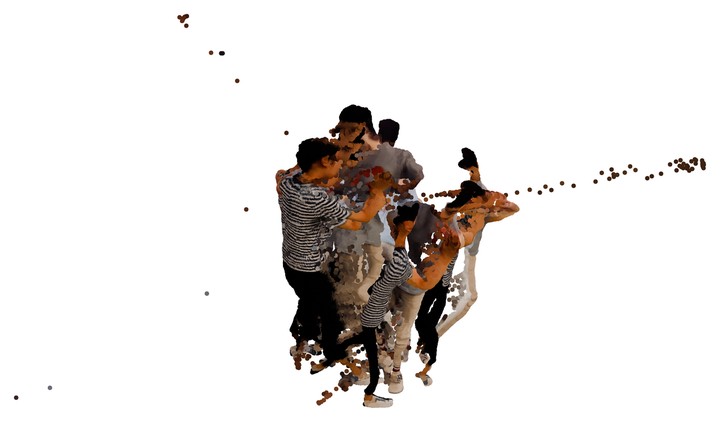}} & \raisebox{-0.5\height}{\includegraphics[width=0.155\textwidth]{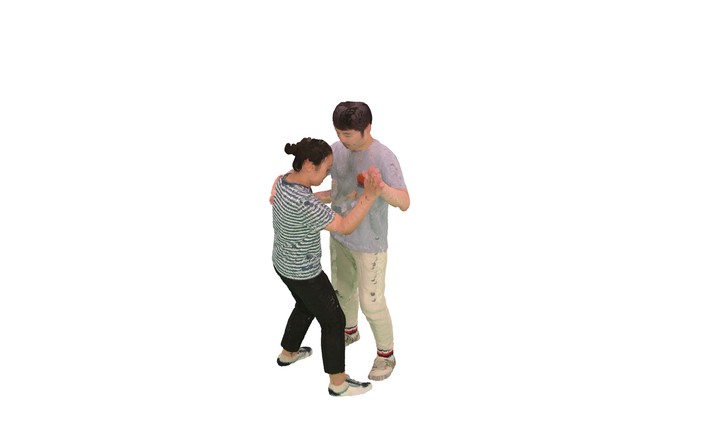}} & \raisebox{-0.5\height}{\includegraphics[width=0.155\textwidth]{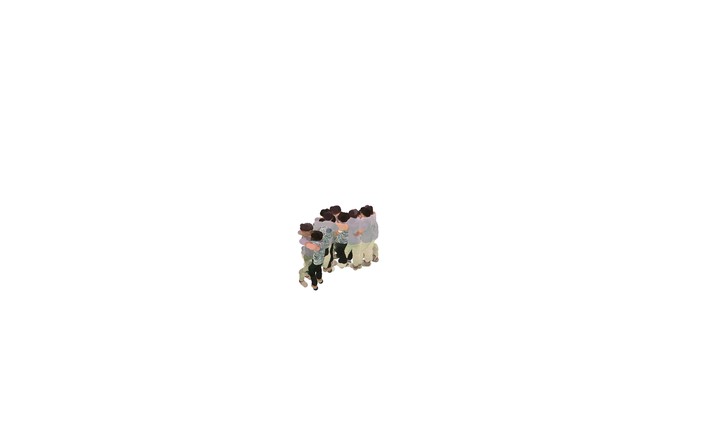}} & \raisebox{-0.5\height}{\includegraphics[width=0.155\textwidth]{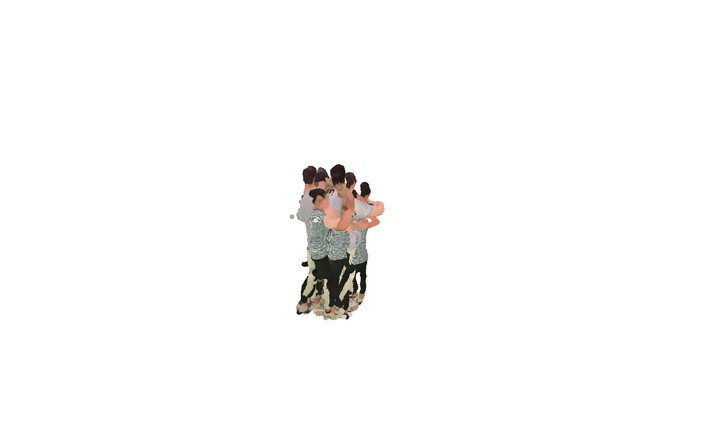}}\\
    \raisebox{-0.5\height}{\rotatebox{90}{\footnotesize actors, top}} & \raisebox{-0.5\height}{\includegraphics[width=0.155\textwidth]{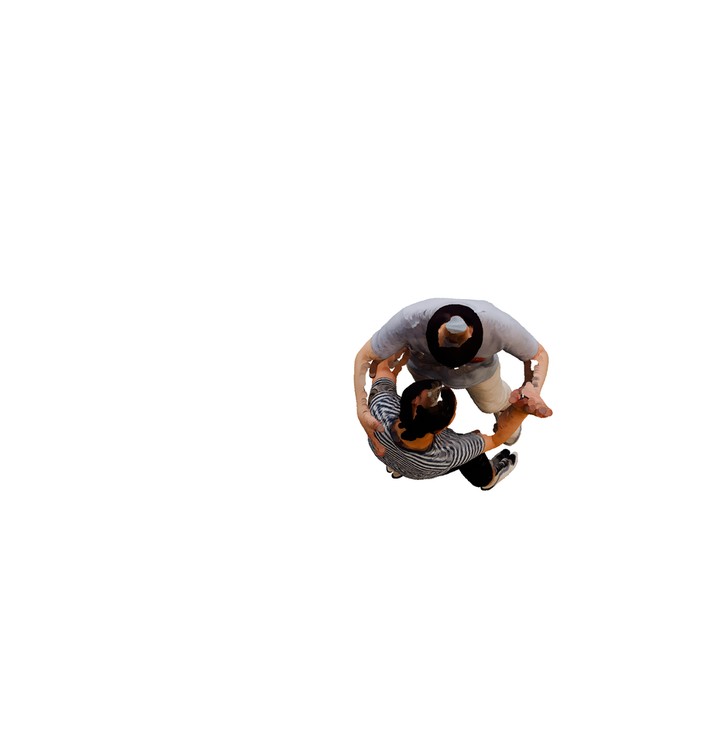}} & \raisebox{-0.5\height}{\includegraphics[width=0.155\textwidth]{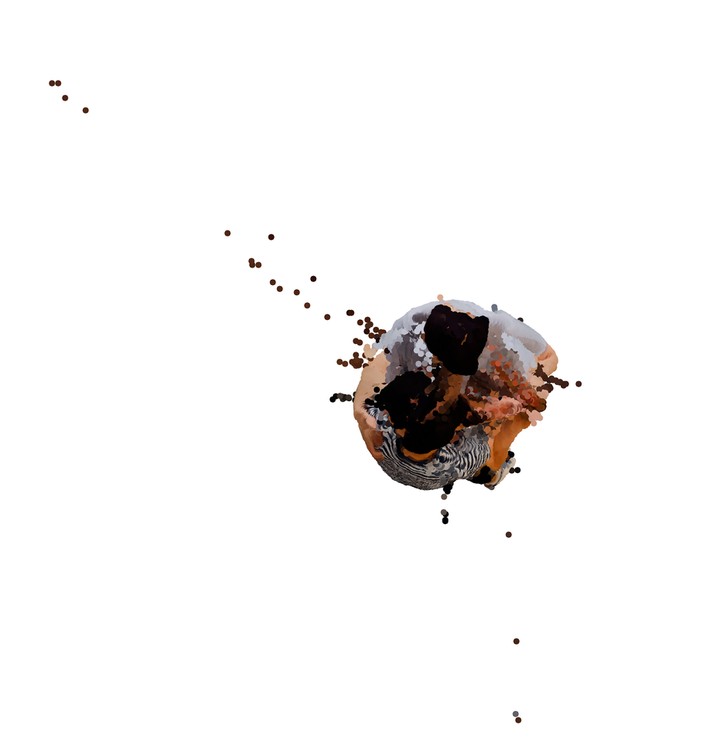}} & \raisebox{-0.5\height}{\includegraphics[width=0.155\textwidth]{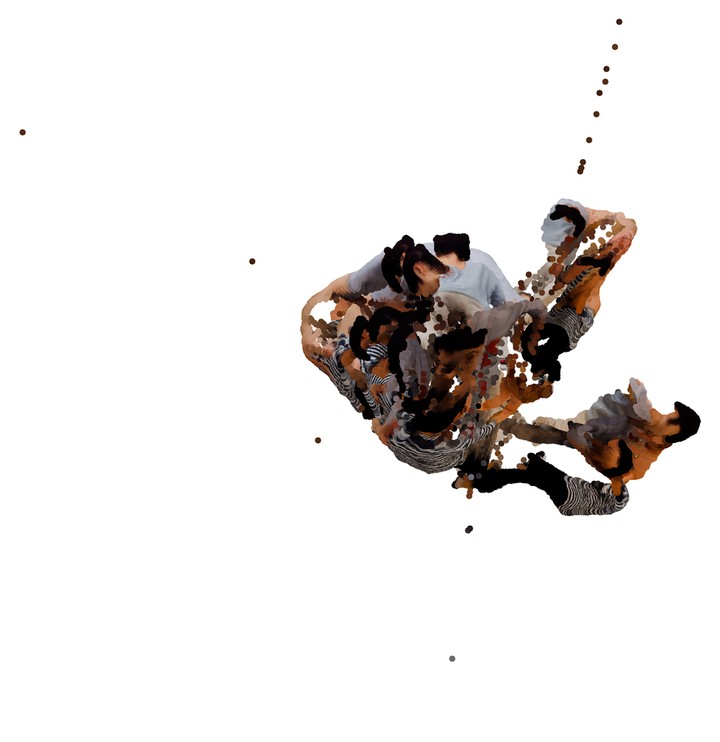}} & \raisebox{-0.5\height}{\includegraphics[width=0.155\textwidth]{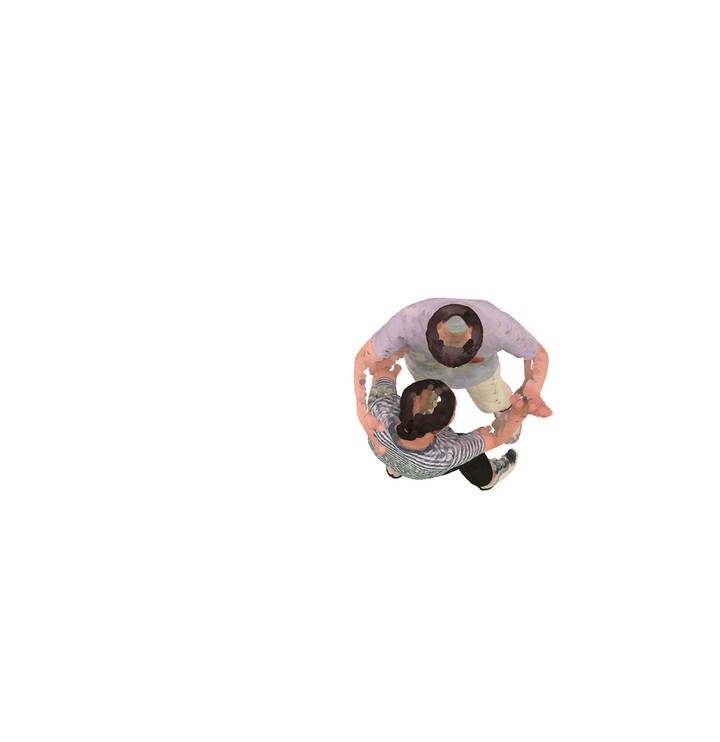}} & \raisebox{-0.5\height}{\includegraphics[width=0.155\textwidth]{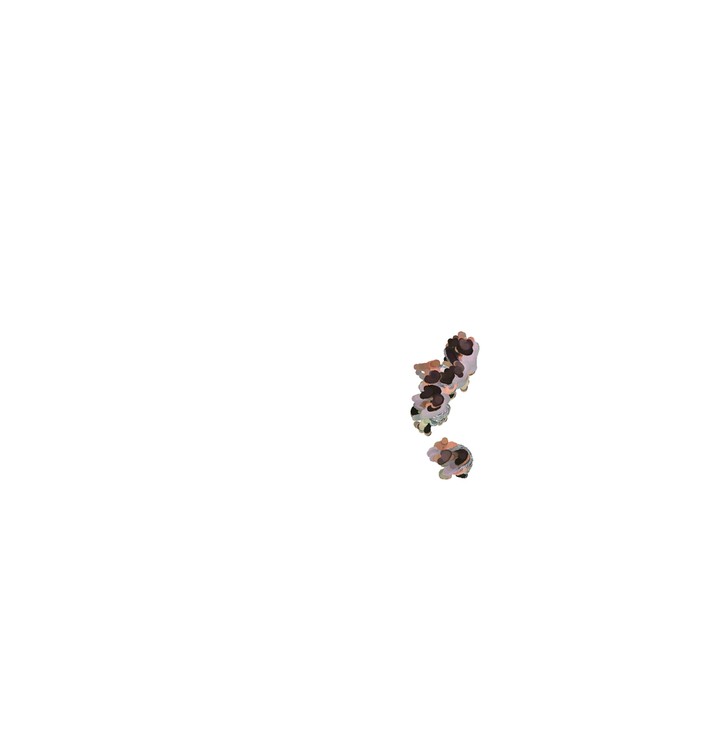}} & \raisebox{-0.5\height}{\includegraphics[width=0.155\textwidth]{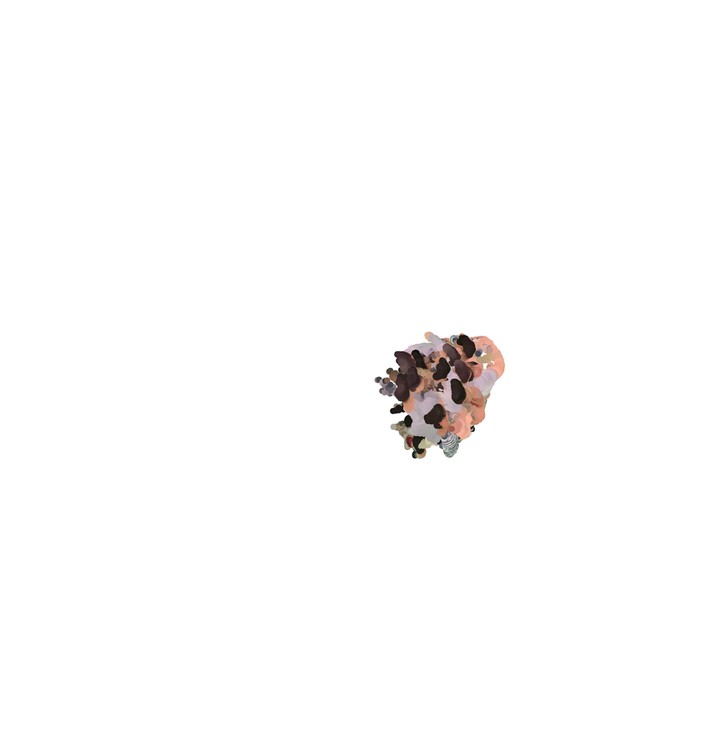}}\\
  \end{tabular}
  \vspace{-0.2cm}
  \caption{\textbf{Multi-view reconstruction fails on the static-only split.} The VGGT-$\Omega$ and OmniX reconstructions of \texttt{pair00\_dance00} at $t{=}70$ on both Hi4D splits, aligned to the ground truth with the pointmap Sim(3) of \cref{tab:reconstruction_quality} and rendered as RGB point clouds from two cameras shared by all columns. The \emph{scene} rows show the reconstruction itself: on the rendered split the pixels whose ground-truth point lies within 1.6\,m of the actors, so that the walls do not hide them, and on the static split every pixel, where the ground-truth scan depth covers only the actors. The \emph{actors} rows keep only the pixels on the actors, given by the surface labels of the rendered split and by the valid scan depth of the static split. All masks drop a 16-pixel band around the actor silhouettes, where a depth map mixes the actor and the surface behind it. On the rendered split VGGT-$\Omega$ recovers the floor and the actors as one consistent body from the eight views, while OmniX duplicates them. On the original captures neither registers the eight static cameras: the green-screen studio is nearly symmetric, and its repeated light panels and tripods are matched across the wrong views, so VGGT-$\Omega$ collapses the whole studio into a small scrambled cluster around the actors and OmniX folds it into a distorted room. This is the failure behind the near-zero E\textsubscript{W} and D\textsubscript{W} of the split and the collapse of \dmetric{W} and \dmetric{N} in \cref{tab:hi4d_static_results}, while the query-view metric, scored in each camera's own frame, is unaffected. The monocular methods reconstruct every view on its own and are not shown.}
  \label{fig:hi4d_static_recon}
\end{figure*}
\begin{table*}[!t]
\centering
\setlength{\tabcolsep}{3.2pt}
\caption{\textbf{Reconstruction quality on the two Hi4D splits.} Metrics follow \cref{tab:reconstruction_quality}. VGGT-$\Omega$ is the shared reconstruction supplied to every point tracker in \cref{tab:hi4d_static_results}; the remaining rows evaluate the native geometry of joint methods. Since every camera in Hi4D-static is fixed, E\textsubscript{C} is undefined. Best and second-best entries per split and metric are bold and underlined. Dashes mark undefined or unavailable values.}
\label{tab:hi4d_static_reconstruction}
\vspace{-0.27cm}
\begin{tabular}{lccccc@{\hspace{8pt}}ccccc}
\toprule
 & \multicolumn{5}{c}{\textbf{Hi4D}} & \multicolumn{5}{c}{\textbf{Hi4D-static}}\\
\cmidrule(lr){2-6}\cmidrule(lr){7-11}
Reconstruction & K\,\(\downarrow\) & E\textsubscript{C}\,\(\uparrow\) & E\textsubscript{W}\,\(\uparrow\) & D\textsubscript{C}\,\(\uparrow\) & D\textsubscript{W}\,\(\uparrow\) & K\,\(\downarrow\) & E\textsubscript{C}\,\(\uparrow\) & E\textsubscript{W}\,\(\uparrow\) & D\textsubscript{C}\,\(\uparrow\) & D\textsubscript{W}\,\(\uparrow\)\\
\midrule
VGGT-$\Omega$~\cite{wang2026vggtomega} (as served) & \textbf{0.95} & \underline{92.6} & \textbf{95.6} & \textbf{45.0} & \textbf{40.7} & \textbf{1.07} & -- & \textbf{14.5} & 7.8 & \textbf{0.2}\\
\midrule
L4P~\cite{badki2026l4p} & 5.83 & 74.1 & -- & 1.2 & -- & 7.56 & -- & -- & 5.8 & --\\
St4RTrack~\cite{feng2025st4rtrack} & 6.52 & 1.0 & -- & 0.0 & -- & 7.78 & -- & -- & 0.4 & --\\
Any4D~\cite{karhade2025any4d} & 3.89 & 82.8 & -- & 6.2 & -- & 3.74 & -- & -- & 11.9 & --\\
SpatialTrackerV2~\cite{xiao2025spatialtrackerv2} & 3.41 & 53.7 & -- & 3.1 & -- & -- & -- & -- & -- & --\\
V-DPM~\cite{sucar2026vdpm} & 3.75 & 82.1 & -- & 0.9 & -- & 3.72 & -- & -- & \textbf{21.6} & --\\
4RC~\cite{luo2026fourrc} & 2.24 & \textbf{96.1} & -- & 23.0 & -- & 1.90 & -- & -- & \underline{13.3} & --\\
OmniX~\cite{jiang2026omnix} & \underline{1.50} & 62.5 & \underline{21.0} & \underline{24.8} & \underline{0.5} & \underline{1.49} & -- & \underline{11.2} & 9.3 & \textbf{0.2}\\
\bottomrule
\end{tabular}%
\end{table*}

\begin{table*}[!t]
\centering
\setlength{\tabcolsep}{2.5pt}
\caption{\textbf{Hi4D: rendered split versus the original static captures.} Every method of \cref{tab:main_results} on the two Hi4D splits, which share the human 3D tracks and cameras 0--3 (\cref{fig:hi4d_static_split}); the rendered split adds an equal number of static tracks on its environment. The rendered split feeds renders of the textured scans inside a synthetic environment from four static and four moving cameras; the static-only split feeds the eight original RGB videos. Point trackers receive the VGGT-$\Omega$ reconstruction of the respective videos. Cells report \dmetric{W}, \dmetric{N}, and \dmetric{Q} as in \cref{tab:main_results}, best and second-best per column are bold and underlined, and gray MV-TAP rows use oracle queries and are unranked. The shared VGGT-$\Omega$ reconstruction registers the original captures far worse than the renders (\cref{tab:hi4d_static_reconstruction}), so the world-space and non-query-view metrics, which depend on cross-view registration, collapse on the static split while query-view accuracy holds. -- marks a method that was not run on the static split (SpatialTrackerV2).}
\label{tab:hi4d_static_results}
\vspace{-0.27cm}
\begin{tabular}{lcccccccc}
\toprule
 &  & \multicolumn{3}{c}{\textbf{Hi4D}} &  & \multicolumn{3}{c}{\textbf{Hi4D-static}}\\
\cmidrule(lr){3-5}\cmidrule(lr){7-9}
Method &  & \dmetric{W} & \dmetric{N} & \dmetric{Q} &  & \dmetric{W} & \dmetric{N} & \dmetric{Q}\\
\midrule
\addlinespace[0.15em]
\multicolumn{9}{l}{1.a) \textbf{Monocular 2D point trackers} \emph{(lifted with VGGT-$\Omega$~\cite{wang2026vggtomega})}}\\
\midrule
PIPs~\cite{harley2022particle} &  & 13.5 & 10.3 & 15.4 &  & 0.9 & 1.0 & 9.5\\
PIPs++~\cite{zheng2023pointodyssey} &  & 13.8 & 10.2 & 16.0 &  & 0.9 & 1.0 & 9.3\\
TAPTRv3~\cite{li2024taptrv3} &  & 15.1 & 11.2 & 17.3 &  & 1.0 & 1.1 & 10.4\\
CoTracker2~\cite{karaev2024cotracker} &  & 15.4 & 11.4 & 17.8 &  & 1.0 & 1.1 & 10.2\\
TAPIR~\cite{doersch2023tapir} &  & 15.7 & 11.7 & 18.7 &  & 1.0 & 1.1 & 10.9\\
CoTracker1~\cite{karaev2024cotracker} &  & 15.7 & 11.7 & 18.0 &  & 1.0 & 1.1 & 10.4\\
BootsTAPIR~\cite{doersch2024bootstap} &  & 15.6 & 11.5 & 18.3 &  & 1.0 & 1.1 & 10.4\\
DOT~\cite{lemoing2024dense} &  & 16.0 & 12.0 & 18.9 &  & 1.0 & \underline{1.2} & 10.8\\
LocoTrack~\cite{cho2024locotrack} &  & 15.7 & 11.5 & 18.6 &  & 1.0 & 1.1 & 10.6\\
CoTracker3~\cite{karaev2024cotracker3} &  & 16.3 & 12.2 & 18.7 &  & 1.0 & 1.1 & 10.8\\
MegaFlow~\cite{zhang2026megaflow} &  & 16.0 & 11.8 & 18.8 &  & 1.0 & \underline{1.2} & 11.0\\
TAPNext++~\cite{jung2026tapnextpp} &  & 16.8 & 12.6 & \textbf{19.6} &  & \underline{1.1} & 1.1 & 11.0\\
Track-On-R~\cite{aydemir2025trackon} &  & 16.1 & 11.8 & 18.9 &  & 1.0 & 1.1 & 10.7\\
AllTracker~\cite{harley2025alltracker} &  & 16.7 & 12.5 & \underline{19.3} &  & 1.0 & 1.1 & 11.0\\
CoWTracker~\cite{lai2026cowtracker} &  & 16.7 & 12.4 & \textbf{19.6} &  & 1.0 & 1.1 & 11.0\\
\midrule
\addlinespace[0.15em]
\multicolumn{9}{l}{1.b) \textbf{Monocular 3D point trackers} \emph{(with VGGT-$\Omega$~\cite{wang2026vggtomega} depth input)}}\\
\midrule
SpatialTracker~\cite{xiao2024spatialtracker} &  & 7.8 & 5.5 & 10.6 &  & 0.6 & 0.8 & 8.3\\
SceneTracker~\cite{wang2024scenetracker} &  & 11.1 & 8.3 & 13.9 &  & 0.7 & 0.8 & 7.6\\
DELTA~\cite{ngo2025delta} &  & 13.6 & 10.6 & 16.8 &  & 0.7 & 0.9 & 9.6\\
SpatialTrackerV2~\cite{xiao2025spatialtrackerv2} &  & 16.1 & 12.3 & 17.8 &  & 0.9 & 1.0 & 9.9\\
TAPIP3D~\cite{zhang2025tapip3d} &  & \underline{20.5} & \underline{15.8} & \textbf{19.6} &  & 0.9 & 1.1 & 10.5\\
\midrule
\addlinespace[0.15em]
\multicolumn{9}{l}{1.c) \textbf{Multi-view 2D point trackers} \emph{(lifted with VGGT-$\Omega$~\cite{wang2026vggtomega})}}\\
\midrule
MV-TAP~\cite{koo2025mvtap} &  & 17.0 & 13.8 & 17.0 &  & 0.7 & 1.0 & 8.3\\
\textcolor{gray}{MV-TAP (MV)$^\ddagger$~\cite{koo2025mvtap}} &  & \textcolor{gray}{22.9} & \textcolor{gray}{21.7} & \textcolor{gray}{23.0} &  & \textcolor{gray}{1.1} & \textcolor{gray}{2.5} & \textcolor{gray}{3.1}\\
\textcolor{gray}{MV-TAP (MT)$^\dagger$~\cite{koo2025mvtap}} &  & \textcolor{gray}{24.0} & \textcolor{gray}{22.9} & \textcolor{gray}{22.9} &  & \textcolor{gray}{1.0} & \textcolor{gray}{2.4} & \textcolor{gray}{2.6}\\
\midrule
\addlinespace[0.15em]
\multicolumn{9}{l}{1.d) \textbf{Multi-view 3D point trackers} \emph{(with VGGT-$\Omega$~\cite{wang2026vggtomega} depth input)}}\\
\midrule
LAPA~\cite{galoaa2025lookaround} &  & 2.3 & 2.5 & 3.4 &  & 0.3 & 0.4 & 1.9\\
MVTracker~\cite{rajic2025mvtracker} &  & 19.1 & 15.1 & 17.8 &  & 0.7 & 1.0 & 5.5\\
MVTracker (finetuned)~\cite{rajic2025mvtracker} &  & \textbf{20.8} & \textbf{16.4} & 18.8 &  & 0.7 & 1.0 & 5.6\\
\midrule
\addlinespace[0.15em]
\multicolumn{9}{l}{2.a) \textbf{Monocular 4D reconstruction and point tracking}}\\
\midrule
L4P~\cite{badki2026l4p} &  & -- & -- & 1.1 &  & -- & -- & 0.0\\
St4RTrack~\cite{feng2025st4rtrack} &  & -- & -- & 0.3 &  & -- & -- & 0.7\\
Any4D~\cite{karhade2025any4d} &  & -- & -- & 3.3 &  & -- & -- & 9.1\\
SpatialTrackerV2~\cite{xiao2025spatialtrackerv2} &  & -- & -- & 4.9 &  & -- & -- & --\\
OmniX~\cite{jiang2026omnix} (monocular) &  & -- & -- & 7.1 &  & -- & -- & 13.4\\
V-DPM~\cite{sucar2026vdpm} &  & -- & -- & 5.6 &  & -- & -- & \underline{14.8}\\
4RC~\cite{luo2026fourrc} &  & -- & -- & 11.2 &  & -- & -- & \textbf{18.6}\\
\midrule
\addlinespace[0.15em]
\multicolumn{9}{l}{2.b) \textbf{Multi-view 4D reconstruction and point tracking}}\\
\midrule
OmniX~\cite{jiang2026omnix} &  & 1.2 & 1.1 & 5.7 &  & \textbf{1.6} & \textbf{1.6} & 9.6\\
\bottomrule
\end{tabular}%
\end{table*}

\section{Raw File Specification}
\label{sec:file_spec}
We release each multi-view TAP-3D sequence in the following directory layout, where \texttt{<view\_id>/} is repeated once per camera view:

\begin{verbatim}
<dataset_split_name>/<sequence_name>/
|-- tracks_xyz.npy
|-- queries_xytv.npy
|-- track_is_static.npy     (Hi4D only)
`-- <view_id>/
    |-- images_jpeg_bytes.npy
    |-- intrinsics.npy
    |-- extrinsics_w2c.npy
    |-- visibility.npy
    |-- depth.npy[.gz|z]     (optional)
    |-- surface_semantic_ids.npy.gz  (Hi4D only)
    `-- foreground_mask.npy  (optional)
\end{verbatim}

\paragraph{Per-view files.}
\begin{itemize}
\item \texttt{images\_jpeg\_bytes.npy}: $(F,)$ \texttt{object}, where $F$ is the number of frames. Each element is a 1-D \texttt{uint8} array of JPEG-encoded bytes that decode to an $(H, W, 3)$ RGB image.
\item \texttt{visibility.npy}: $(F, P)$ \texttt{bool}, where $P$ is the number of point tracks. Entry $[t, p]$ is \texttt{True} when track $p$ is visible from this view at frame $t$.
\item \texttt{intrinsics.npy}: $(4,)$ \texttt{float32}. Pinhole parameters $(f_x, f_y, c_x, c_y)$ in pixels, constant across frames.
\item \texttt{extrinsics\_w2c.npy}: $(F, 4, 4)$ \texttt{float32}. Per-frame world-to-camera rigid transforms in standard CV convention (X-right, Y-down, Z-forward). Applying $E[t]$ to a world-space point yields camera-space coordinates for projection with the pinhole intrinsics.
\item \texttt{depth.npy} (optional; stored as \texttt{depth.npy}, gzip-compressed \texttt{depth.npy.gz}, or \texttt{depth.npz} depending on the split, and read by the same loader): $(F, H, W)$ \texttt{float32}. Metric depth in meters; positive finite values are valid and zero denotes missing/invalid depth. Used for pseudo/ground-truth depth experiments and visualization.
\item \texttt{foreground\_mask.npy} (optional): $(F, H, W)$ \texttt{bool}. Foreground/support mask, used for visualization.
\item \texttt{surface\_semantic\_ids.npy.gz} (Hi4D only): $(F, H, W)$ \texttt{uint8}. Class of the first surface hit at each pixel: 0 no modeled geometry, 1 opaque environment, 2 glass, mirror, or water, 3 captured human scan. Depth is valid only for classes 1 and 3.
\end{itemize}

\paragraph{Shared (sequence-level) files.}
\begin{itemize}
\item \texttt{tracks\_xyz.npy}: $(F, P, 3)$ \texttt{float32}. 3-D world-space positions of each point track at each frame, shared across all views.
\item \texttt{track\_is\_static.npy} (Hi4D only): $(P,)$ \texttt{bool}. \texttt{False} for the human tracks, which come first, and \texttt{True} for the static environment tracks.
\item \texttt{queries\_xytv.npy}: $(P, 4)$ \texttt{float32}. One query per track, stored as $(x, y, t, v)$: pixel coordinates $(x, y)$ in the queried view, 0-indexed frame $t$, and 0-indexed view $v$ (corresponding to a \texttt{<view\_id>} subdirectory). Each query is sampled uniformly at random (seed 42; seed 72 for Hi4D) from the $(\text{frame}, \text{view})$ pairs in which the track is visible.
\end{itemize}

\paragraph{Pixel-center conventions.}
Pixel coordinates use the integer-center convention: pixel $(0, 0)$ is the center of the top-left pixel, so valid coordinates satisfy $-0.5 \le x < W - 0.5$ and $-0.5 \le y < H - 0.5$. Projection uses $x = f_x X / Z + c_x$ and $y = f_y Y / Z + c_y$, and the nearest image-array index of a valid coordinate is $p_x = \lfloor x + 0.5 \rfloor$, $p_y = \lfloor y + 0.5 \rfloor$. Under this convention, pixel center $0$ maps to pixel center $0$ and pixel center $W{-}1$ maps to $W'{-}1$ when resizing; coordinate and intrinsic scaling should therefore use $(W'{-}1)/(W{-}1)$ and $(H'{-}1)/(H{-}1)$, not $W'/W$ and $H'/H$.

\section{Limitations}
\label{sec:limitations}

\paragraph{Annotation precision and low-threshold metrics.}
Annotation precision varies across the benchmark. In particular, DROID, Hi4D, Ego-Exo4D, and Harmony4D rely on estimated depth, calibration, fitted meshes, SLAM/SfM geometry, or model-assisted track construction. Although we manually inspect the projected trajectories and remove visible failures, we cannot measure their absolute label error, such as mean, median, or maximum 3D drift, because independent metric trajectories are unavailable for these subsets. Performance at the strictest thresholds may therefore partly reflect annotation noise rather than model error. We nevertheless expect method comparisons to remain informative because all methods are scored against the same labels and we are not aware of systematic annotation errors that favor a particular method. Strict-threshold results on these subsets should be interpreted together with the larger-threshold and aggregate metrics.

\paragraph{Highly deformable objects.}
The benchmark contains articulated humans and moving rigid objects, but lacks precise annotations for highly deformable objects such as cloth and other soft materials. Producing persistent metric surface correspondences for such objects is substantially harder because neither rigid poses nor standard articulated meshes provide adequate supervision. Extending the benchmark to these objects would test an important class of interactions that is currently under-represented.

\paragraph{Training-data utility.}
We release large-scale training resources derived from DROID and Perpetua. Fine-tuning MVTracker for 5,000 steps on a training mixture containing Perpetua produced initial performance gains. However, we have not run controlled comparisons that isolate the contribution of Perpetua, nor have we trained on the released DROID trajectories. We therefore do not establish how much either resource improves multi-view tracking, reconstruction, or transfer to held-out evaluation subsets.

\paragraph{Scaling protocol.}
Like TAPVid-3D~\cite{koppula2024tapvid3d}, we rescale predictions before evaluation to account for global scale ambiguity. Our headline 3D metrics specifically use query-median scaling. We have not systematically tested how alternative scaling protocols affect absolute scores or method rankings. The reported comparisons should therefore be interpreted under this protocol.

\section{Compute}
\label{sec:compute}
All reported numbers were computed on NVIDIA GH200 GPUs (96\,GB VRAM) of an HPC cluster, with a handful of runs on an NVIDIA H200 (140\,GB VRAM). Summed over the final runs behind \cref{tab:main_results,tab:gt_depth_results}, the predictions took about 2{,}500 GPU-hours of GPU time. The view-scaling study (\cref{fig:views_scaling}) adds about 2{,}500 GPU-hours and the depth-source study (\cref{tab:depth_source_results}) about 800. The benchmark holds 284 sequences with 53{,}000 timesteps and 207{,}000 frames, and every row of \cref{tab:main_results,tab:gt_depth_results} processes all of them forward and time-reversed, so even at one second per multi-view frame a row costs about 30 GPU-hours. These figures are a lower bound. Counting also the evaluations on earlier iterations of the labels and other reruns, evaluating the baselines took about 10{,}000 GPU-hours in total. Of this total, 58\% went to the runs behind the tables, 22\% to the view-scaling study (\cref{fig:views_scaling}), 10\% to the depth-source study (\cref{tab:depth_source_results}), and 10\% to development runs. By method, the 4D reconstruction and tracking methods of groups 2.a and 2.b take 43\% (V-DPM 22\%, OmniX 14\%, St4RTrack 4\%, Any4D 2\%), the 2D trackers that must be rerun once per distinct query timestep 20\% (MegaFlow 9\%, CoWTracker 6\%, DOT 3\%, AllTracker 1\%), MV-TAP 8\%, TAPIP3D 5\%, the feed-forward reconstruction sources 4\%, and the remaining methods 11\% together. This excludes the costs related to building the datasets. Fine-tuning MVTracker took 24 GPUh on NVIDIA H200 GPUs (140\,GB VRAM) and a total of about 150 GPU-hours for iteration and data preprocessing.

\end{document}